%% file: main_1_tmlr.tex
\documentclass[10pt]{article}
\usepackage[english]{babel}

\usepackage[accepted]{tmlr}
\input{math_commands.tex}

\input{preamble_tmlr}

\def\month{04}   % Camera-ready: insert correct month
\def\year{2026}  % Camera-ready: insert correct year
\def\openreview{\url{https://openreview.net/forum?id=jtnMIbJIso}}

\title{Variational Visual Question Answering for Uncertainty-Aware Selective Prediction}

\author{\name Tobias Jan Wieczorek \email tobias.wieczorek@tu-darmstadt.de\\
\addr TU Darmstadt \& hessian.AI, Germany
\AND
\name Nathalie Daun \\
\addr TU Darmstadt \& hessian.AI, Germany
\AND
\name Mohammad Emtiyaz Khan \\
\addr RIKEN Center for Advanced Intelligence Project, Tokyo, Japan
\AND
\name Marcus Rohrbach \\
\addr TU Darmstadt \& hessian.AI, Germany
}

\begin{document}

\maketitle

\input{sec_tmlr/0_abstract}
\input{sec_tmlr/1_intro}
\input{sec_tmlr/2_related_work}
\input{sec_tmlr/4_varvqa}

\input{sec_tmlr/5_1_experiments_setup}
\input{sec_tmlr/5_2_experiments}

\input{sec_tmlr/6_conclusion}
\input{sec_tmlr/7_acknowledgements}

\FloatBarrier
{
    \small
    \bibliographystyle{ieeenat_fullname}
    \bibliography{main}
}

% ============================================================
% APPENDIX
% ============================================================
\appendix
\startsupplement

\input{sec_tmlr/A_hyperparam_report}

\input{sec_tmlr/B_time}

\input{sec_tmlr/C_calibration_dropoutbpa}

\input{sec_tmlr/D_full_results}
\input{sec_tmlr/E_qualitative}

\input{sec_tmlr/F_largefigures}
\input{sec_tmlr/G_ensembles}
\input{sec_tmlr/H_threshold_generalization}
\input{sec_tmlr/I_selector_comparison}

\end{document}

%% file: math_commands.tex
\usepackage{amsmath,amsfonts,bm}

\def\eqref#1{equation~\ref{#1}}
\def\1{\bm{1}}

\DeclareMathAlphabet{\mathsfit}{\encodingdefault}{\sfdefault}{m}{sl}
\SetMathAlphabet{\mathsfit}{bold}{\encodingdefault}{\sfdefault}{bx}{n}

\DeclareMathOperator*{\argmax}{arg\,max}

%% file: preamble_tmlr.tex
\usepackage[dvipsnames]{xcolor}
\usepackage{arydshln, hyperref, url, amsmath, amssymb, mathtools, amsthm, bbm, csquotes, multirow, subcaption, comment, balance, array, booktabs, xspace, placeins}

\usepackage[nameinlink]{cleveref}
\crefname{figure}{Fig.}{Figs.}
\Crefname{figure}{Figure}{Figures}
\crefname{table}{Tab.}{Tabs.}
\Crefname{table}{Table}{Tables}
\crefname{section}{Sec.}{Secs.}
\Crefname{section}{Section}{Sections}
\crefname{equation}{Eq.}{Eqs.}
\Crefname{equation}{Equation}{Equations}
\crefname{appendix}{App.}{App.}
\Crefname{appendix}{Appendix}{Appendix}

\definecolor{TolMutedBlue}{HTML}{332288}
\definecolor{TolMutedGreen}{HTML}{117733}
\definecolor{TolMutedPurple}{HTML}{AA4499}

\hypersetup{
    colorlinks=true,
    citecolor=TolMutedBlue,
    linkcolor=TolMutedGreen,
    urlcolor=TolMutedPurple,
    filecolor=TolMutedGreen,
    menucolor=TolMutedGreen,
    runcolor=TolMutedGreen,
}

\newcommand{\myparagraph}[1]{\vspace{0.75em}\noindent\textbf{#1.}\hspace{0.75em}}

\newcommand{\cf}{\textit{cf.}\ }
\newcommand{\eg}{\textit{e.g.}\ }
\newcommand{\ie}{\textit{i.e.}\ }

\newcommand{\gray}{\textcolor{gray}}

\newcommand{\startsupplement}{
    \setcounter{section}{0}
    \renewcommand{\thesection}{\Alph{section}}
    \crefname{section}{Section}{Sections}
    \renewcommand{\theHsection}{\Alph{section}}
}

\newcommand{\approach}{Variational VQA\xspace}
\newcommand{\approachShort}{VarVQA\xspace}

\newcommand{\gmpmu}{$g^{\mu}_{\textrm{MP}}$}
\newcommand{\gsigma}{$g_{\textrm{BPA}}$}

%% file: sec_tmlr/0_abstract.tex
\begin{abstract}

Despite remarkable progress in recent years, Vision Language Models (VLMs) remain prone to overconfidence and hallucinations on tasks such as Visual Question Answering (VQA) and Visual Reasoning. Bayesian methods can potentially improve reliability by helping models \emph{predict selectively}, that is, models respond only when they are sufficiently confident. Unfortunately, such approaches can be costly and ineffective for large models, and there exists little evidence to show otherwise for multimodal applications. Here, we show for the first time the effectiveness and competitive edge of variational Bayes for selective prediction in VQA. We build on recent advances in variational methods for deep learning and propose an extension called ``Variational VQA''. This method improves calibration and yields significant gains for selective prediction on VQA and Visual Reasoning, particularly when the error tolerance is low ($\leq 1\%$). Often, just one posterior sample yields more reliable answers than those given by models trained with AdamW. In addition, we propose a new risk-averse selector that outperforms standard sample averaging by considering the variance of predictions. Overall, we present compelling evidence that variational learning is a viable option to make large VLMs safer and more trustworthy.

\end{abstract}

%% file: sec_tmlr/1_intro.tex
\section{Introduction}

\label{sec:intro}

\begin{figure}[!thbp]
    \centering
    \includegraphics[width=0.95\textwidth]{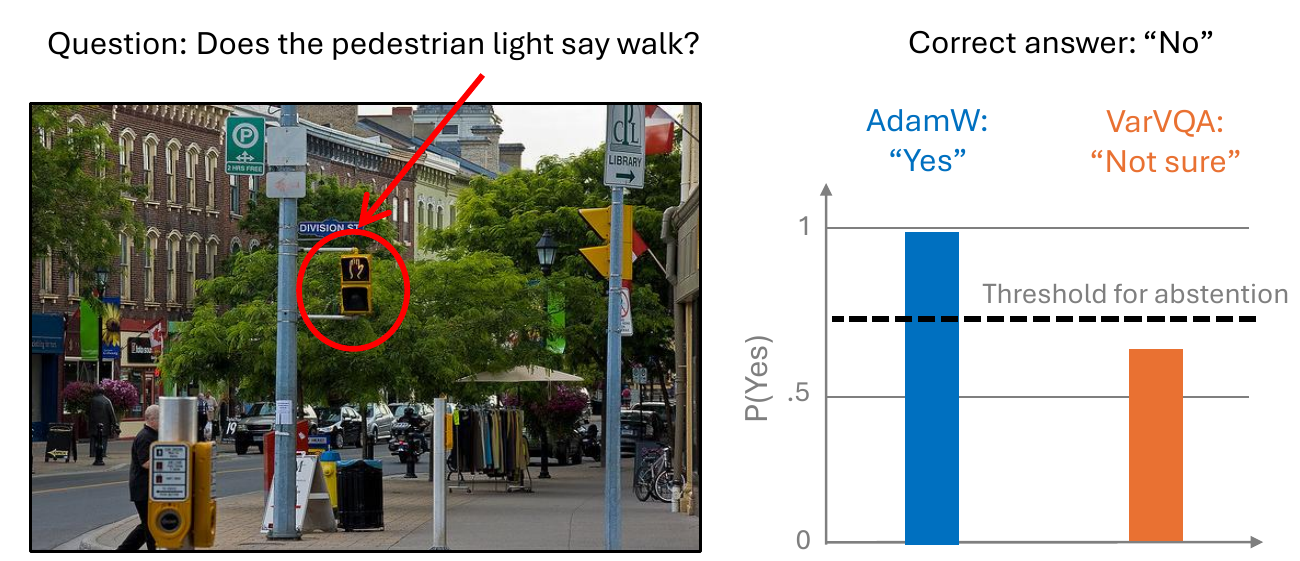}
    \caption{Despite recent performance gains, VLMs trained with popular optimizers like AdamW do not know when they are wrong. Our \approachShort approach uses posterior variances to help the model decide when to abstain. The result shown on the right is for BEiT-3 \citep{beit3}, which achieves near-human accuracy on VQAv2 \citep{vqav2}.}
    \label{fig:teaser}
\end{figure}

Advances in VLMs \citep{beit3, qwen2, llavanext} have led to substantial gains on classical VQA benchmarks \citep{vqa, vqav2}, with performance now approaching or surpassing human levels. However, even strong VQA models are miscalibrated, prone to hallucinations, and often confidently guess answers instead of expressing uncertainty (\cf \cref{fig:teaser}). In short, these models do not have a good notion of confidence about their own knowledge. This shortcoming hinders their deployment in safety-critical domains such as medical diagnosis or assistance for the visually impaired. When a model is confronted with adversarial \citep{advqa} or unanswerable \citep{vizwiz} inputs, which are common in the real world, these issues become even more pronounced.

One way to improve reliability is to allow a model to abstain when it is unsure of its response. The selective prediction framework formalizes such abstentions \citep{chow1957optimum,el-yaniv2010foundations}, where the main challenge is to find a good confidence estimator that separates correct answers from incorrect ones. Given such an estimator, the incorrect answers can be replaced with an output that indicates that the model does not know the answer such `Not sure' or `I do not know', which replaces potentially costly errors with abstentions. Although recent work has connected selective prediction to hallucinations, \citep{openaiWhyHallucinate}, the literature on multimodal models remains relatively sparse. Previous approaches in the multimodal domain have proposed to incorporate additional model components to improve confidence estimates: \citet{reliable_vqa} train a lightweight head on top of the frozen VLM backbone, while ~\citet{selselpred} use external vision tools and an additional language model to quantify uncertainty. Both works do not attempt to improve the reliability of the underlying model. Instead, their solutions introduce additional overhead to the prediction pipeline while also adding new failure points for uncertainty estimation.

Variational Bayesian (VB) methods \citep{graves2011practical} can potentially address the unreliability of VLMs without requiring additional components or tools. In particular, the uncertainty in the learned posterior distribution over model parameters can be used to help the model make a prediction only when it is sufficiently confident. This theory remains untested though, as for a long time, VB approaches have been ineffective for large transformer-based architectures. However, the recently developed Improved Variational Online Newton (IVON) optimizer \citep{ivon} has enabled effective variational training of models such as GPT-2 \citep{gpt2} with no significant overhead compared to AdamW \citep{adamw}. So far, IVON has been limited to unimodal domains, and in this work, we are the first to test its effectiveness for multimodal applications, particularly with regards to selective prediction. Our contributions are as follows:

\begin{enumerate}
    \item We demonstrate that variational training is effective for large multimodal transformer-based architectures and introduce the Variational VQA (VarVQA) framework for selective VQA abstentions.
    
    %\item \textbf{We provide a comprehensive evaluation methodology for multimodal reliability} encompassing Calibration and selective prediction, with emphasis on high-stakes scenarios where error tolerance is low.
    
    \item We demonstrate improved uncertainty estimation across multiple dimensions: better calibration, and enhanced selective prediction with particularly large gains at low error tolerances, as well as increased robustness under distribution shift.
    
    \item We establish superior sample efficiency compared to Monte-Carlo (MC) Dropout, showing that Variational VQA provides better reliability given an equal compute budget.
    
    \item We propose a new risk-averse selector function that leverages output variance, yielding consistent improvements in \textit{high-stakes} selective prediction where errors are particularly costly.
\end{enumerate}

%% file: sec_tmlr/2_related_work.tex
\section{Background and Related Work}\label{sec:related_work}

\subsection{Uncertainty and Reliability in Visual Question Answering} 

Visual Question Answering (VQA) is a popular multimodal task that requires a model to understand two modalities and their interaction to predict answers. While multimodal models \citep{blip-2, beit3, qwen2} have recently achieved human-level performance on standard benchmarks like VQAv2 \citep{vqav2}, they still perform poorly at selective prediction on the same benchmarks \citep{lyp}. Our work here is the first to address the selective prediction task using variational Bayesian methods.

%and the community has moved to newer VQA benchmarks that test more diverse capabilities, like MMBench \citep{mmbench} and MME \citep{mme}. However, even models that reach near-human accuracy on VQAv2 still perform poorly when evaluated in terms of selective prediction \citep{lyp}. 

%\subsection{Reliability in VQA: Selective Prediction and Calibration} 

\myparagraph{Selective Prediction} In the selective prediction framework \citep{chow1957optimum, el-yaniv2010foundations}, a ``selector'' assigns a confidence score to the answer given by a model and subsequently decides whether the prediction is accepted or the model is forced to abstain instead (that is, it says ``I do not know''). This decision is made by comparing the assigned confidence against a given abstention threshold. In VQA in particular, the model learns a function $f:\mathcal{I}\times\mathcal{Q}\rightarrow\mathcal{A}$ to predict an answer $a\in\mathcal{A}$, given a multimodal input $x=(i,a)$ consisting of an image $i\in\mathcal{I}$ and a question $q\in\mathcal{Q}$. In selective prediction notation, the model output space is augmented by an \emph{abstain} output $\emptyset$. This transforms the predictive model $f$ into a selective model $m$, incorporating both $f$ and a selector $g$. The answer $f(x)$ is accepted if $g(x)$ is above the abstention threshold $\gamma\in\mathbb{R}$, and rejected otherwise. We follow the notation of \citet{reliable_vqa}:

\begin{equation}\label{eq:selective_model}
    m(x) = (f,g)(x) = \begin{cases}
    f(x) & \text{if } g(x) \geq \gamma, \\
    \emptyset & \text{if } g(x) < \gamma.
    \end{cases}
\end{equation}  

A high threshold $\gamma$ corresponds to a conservative case, in which the model answers only the questions on which it is most confident. Lowering $\gamma$ reduces abstentions (higher coverage), but increases the error rate among accepted answers (higher risk). The tradeoff between risk and coverage is unavoidable, but a better confidence estimator yields a lower risk at any given coverage level. In practice, the cost of error or the risk level is specified in advance, and $\gamma$ is set accordingly, see \Cref{sec:experiments_metrics}. Typically, the answer likelihood \citep{geifman_selpred} or the predictive entropy are used as selection functions. 

Most of the prior work on selective prediction can be classified into either external approaches or integrated approaches. In external setups, a selector is built on top of the frozen predictive model, for example in the form of a trainable model head \citep{reliable_vqa, mielke2022reducing, harmony}, LoRA parameters \citep{aspire} or vision tools \citep{selselpred}. In integrated setups, predictor and selector have at least one combined training phase. Integrated selectors take different forms as well, such as a model head \citep{selectivenet} or a dedicated abstention class \citep{deep_gamblers}. However, if model and selector are trained together, instabilities often ensue, which require special treatment \citep{selectivenet}. Bayesian approaches have not been considered for selective prediction so far, with the exception of concurrent work by \citep{nico_selpred}, which has explored IVON for generative language modeling, although they do not consider multimodal tasks. In contrast to prior work on selective prediction in VQA, our objective is to \emph{directly} improve the reliability of model confidence estimates without additional parameters, training phases, or tools. In other words, we train VLMs where reliability is ``baked-in'' by design, not added as an afterthought. 

\myparagraph{Calibration} Calibration represents a different angle on uncertainty estimation, namely the alignment of a model's predictive confidence with its accuracy: When the model expresses $x\%$ confidence in an answer, it should be correct $x\%$ of the time. The difference between calibration and selective prediction becomes clear when considering a model that is correct on $y\%$ of examples and is also always exactly $y\%$ confident (for a fixed $y$ satisfying $0\leq y\leq100$). Although this model is perfectly calibrated, it cannot distinguish its correct and incorrect outputs and thus fails at the task of deciding when to abstain. Prior work has found that large neural networks often exhibit overconfidence, particularly in Out-Of-Domain (OOD) settings \citep{ovadiaCanYouTrust2019}. In unimodal classification tasks, temperature \citep{guo2017calibration} and Platt (vector) \cite{platt1999probabilistic} scaling are effective at improving calibration. Ensembling  \citep{deep_ensembles} typically yields even better results, but requires prohibitive resources to train $N$ models. New ideas, such as prompting the model to express a verbalized confidence have been mostly ineffective for VLMs \citep{xuan2025seeing}. We show that Variational VQA yields well-calibrated VLMs, achieving a lower Expected Calibration Error (ECE) than vector scaling, while matching other sampling methods like Monte-Carlo Dropout \citep{gal2016dropout}. In general, we argue that for a VLM to be reliable, it should be calibrated and also know when to abstain. Both these are improved with Variational VQA in comparison to models trained with the AdamW optimizer.

\subsection{Variational Bayes for Deep Learning}\label{sec:rltwork_varlearning}
Variational Bayesian Learning provides a principled approach to estimate uncertainty by learning probability distributions (often Gaussians) over the weights of a neural network. While conventional deep learning methods estimate network weights $\theta$ by minimizing \emph{empirical risk} $\bar{\ell}(\theta)=\frac{1}{N}\sum_{i=1}^{N}\ell_i(\theta)$, \emph{variational} methods estimate a distribution $q(\theta)$ over network weights by minimizing the variational objective

\begin{equation}\label{eq:variational_loss}
    \mathcal{L}(q(\theta)) = \lambda \mathbb{E}_{q(\theta)} \left[\mathcal{\bar{\ell}}(\theta)\right] + \mathbb{D}_{\text{KL}}(q(\theta) \parallel p(\theta)).
\end{equation}

Here, $N$ is the size of the training set, $\ell_i(\theta)$ the loss for example $i$, $\mathbb{D}_{\text{KL}}(q||p)$ denotes the Kullback-Leibler divergence, $\lambda$ is a scaling parameter, and $p(\theta)$ the prior weights distribution. To keep computational costs manageable, $q(\theta)$ is often chosen to be a diagonal-covariance Gaussian, that is, $q(\theta) = \mathcal{N}(\theta \mid m, \text{diag}(v))$, where $m$ and $v$ are the parameter mean and variance vectors. The objective $\mathcal{L}(q(\theta))$ can be reparametrized in terms of $m$ and $v$, and $\mathcal{L}(m,v)$ is typically approximated through MC sampling of the weights. 

\myparagraph{IVON} In the early 2010s, variational methods that directly optimize parameter means and variances through standard deep learning techniques such as SGD achieved promising results \citep{graves2011practical, blundellWeightUncertaintyNeural2015}. However, in subsequent years, these approaches could not keep up with the growth in scale of network architectures \citep{trippe2018overpruning, foong2020expressiveness, coker2022wide}. Recently, natural gradient methods that build Hessian estimates through an Adam-like update \citep{khan2018fast, osawa2019practical} have addressed this issue. The IVON optimizer \citep{ivon} further develops those and obtains comparable accuracy and better uncertainty estimates than AdamW at nearly identical training cost. In particular, IVON uses an Adam-like \citep{adam} update for the parameter means $m$ and variances $v$. Similarly to adaptive scaling in Adam, IVON updates $m$ by using gradients scaled with $h$ - an online estimate of the diagonal Hessian. In an update step, we first sample $\theta\sim q$, then compute a minibatch gradient $\hat{g}$. The following four lines are used to compute the minibatch Hessian estimate $\hat{h}$, update the moving average of the Hessian $h$, and then update the means $m$ and variances $v$:

\begin{align}
    \hat{h} &\gets \frac{\hat{g}(\theta - m)}{v},\label{eq:ivon_h}\\
    h &\gets \beta_2 h + (1-\beta_2)\hat{h} + \frac{(1-\beta_2)^2(h-\hat{h})^2}{2(h+\delta)}\\
    m &\gets m - \alpha\frac{\bar{g}+\delta m}{h+\delta},\label{eq:ivon_m}\\
    v &\gets \frac{1}{\lambda(h+\delta)}.\label{eq:ivon_v}
\end{align}

Here, $\alpha$ is the learning rate, $\delta$ the weight decay and $\bar{g}$ a moving average of the gradient. IVON also uses Adam-like momentum for the gradients and the Hessian. A notable difference to Adam, however, is the absence of the square root over the scaling vector $h+\delta$ in line \cref{eq:ivon_m}. Notably, the initialization of the Hessian estimate $h_0$ is a crucial hyperparameter to set carefully. For more details, we refer to the original paper by \citet{ivon}.

We use IVON because it offers several advantages compared to other Bayesian baselines. Unlike the Laplace approximation \citep{laplace_approx, laplace_practical}, it does not require an additional pass through the data to compute the Hessian. Neither does it require additional training like Stochastic Weight Averaging (SWA) \citep{swa}. Compared to MC Dropout \citep{gal2016dropout}, the advantage is the availability of a fixed posterior form that can be more easily used for downstream tasks. For instance, the method is easily amenable to ensembling \citep{deep_ensembles}, which can further improve performance \citep{nico_selpred}. We offer new insights compared to previous IVON works \citep{ivon, ivon_lora, nico_selpred}, by showing its effectiveness in training multimodal models and for selective prediction. We further propose a new selection function that uses the output variance, which was never utilized in prior work. 

%% file: sec_tmlr/4_varvqa.tex
\section{Variational VQA}\label{sec:varvqa}

Our Variational VQA approach uses the IVON optimizer to train large VLMs and evaluates the reliability of its output confidences in comparison to baselines like AdamW and MC Droput. In \Cref{sec:var_inference}, we describe how model confidences are obtained, in \Cref{sec:baseline_selectors} we describe the baseline selectors, and in \Cref{sec:bpa} we present our new risk-averse selector.

\subsection{Inference and Model Confidence}\label{sec:var_inference}

At inference, variational methods typically use the learned posterior through MC sampling. However, if computing efficiency is imperative, one can ignore the variances (set $v=0$) and use the mean parameters $m$ for inference \citep{ivon}, which requires only one forward pass. We refer to this approach as `VarVQA mean'. For an input $x$, denote the model's output likelihood vector by $f(x;\theta)$, where the k-th entry $f_k(x;\theta)$ contains the likelihood of class $k$ (out of $K\in\mathbb{N}$ classes). VarVQA, on the other hand, performs sampling, that is, we draw model parameters $\theta^{(s)}\sim q$ to get $S\in\mathbb{N}$ likelihood vectors $f(x;\theta^{(s)})$. These are aggregated to obtain a mean likelihood vector $\bar{\mu}$ and a mean likelihood variance vector $\bar{\sigma}^2$:

\begin{align}
    \bar{\mu}(x) = \frac{1}{S}\sum^S_{s=1} f(x;\theta^{(s)}), \qquad
    \bar{\sigma}^2(x) = \frac{1}{S-1}\sum^S_{s=1}\left[f(x;\theta^{(s)}) - \bar{\mu}(x)\right]^2. \label{eq:mu_sigma}
\end{align}

\subsection{Baseline selector functions}\label{sec:baseline_selectors}

We start with the baseline selector for deterministic methods (AdamW, VarVQA mean). We employ the widely used MaxProb selector \citep{geifman_selpred}, which uses the answer likelihood. In a classification task, the MaxProb selector is defined as $g_{\textrm{MP}}(x) = \max_k f_k(x;\theta)$. We use MaxProb, because we find it to consistently outperform predictive entropy and related functions. In the case of multiple samples, a reasonable baseline is predictive averaging \citep{gal2016dropout}. Here the selector is $g^{\mu}_{\textrm{MP}}(x)= \max_k \bar{\mu}_k(x)$, where $\bar{\mu}_k(x)$ is the $k$-th entry of the vector $\bar{\mu}(x)$, \cf \cref{eq:mu_sigma}. 

\subsection{A new risk-averse selector}\label{sec:bpa}

In this work, particularly for the context of selective prediction, we propose to go Beyond Predictive Averaging (BPA) by also employing the empirical output variances (\cf \cref{eq:mu_sigma}). This is done in a risk-averse \citep{pratt1978risk} manner, by penalizing high-variance predictions. While \citet{pratt1978risk} subtracts the variance (with a prefactor), we found the standard deviation to work best:

\begin{equation}
    g_{\textrm{BPA}}(x) = \bar{\mu}_\kappa(x) - \bar{\sigma}_\kappa(x) \label{eq:g_custom_var}
\end{equation}

\begin{figure}[!tbhp]
    \centering
    \includegraphics[width=0.8\textwidth]{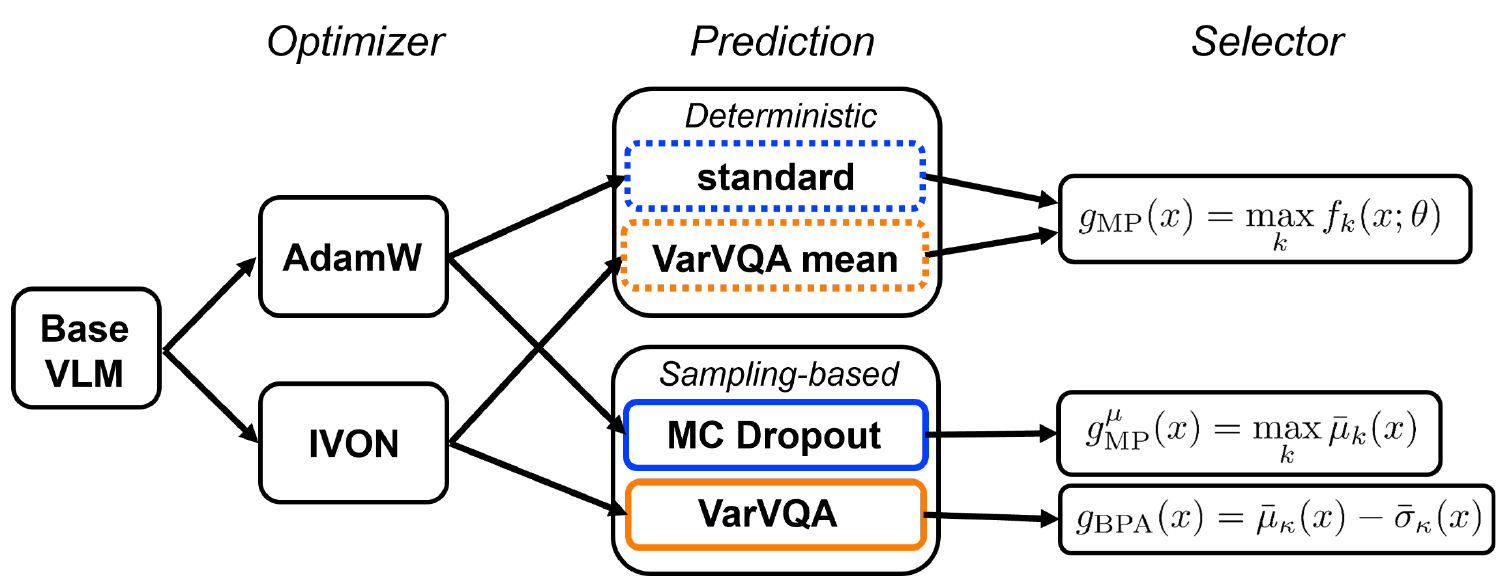}
    \caption{Overview of the methods we experiment with and their selectors. Variational VQA employs ~\gmpmu ~for calibration and \gsigma ~for selective prediction.}
    \label{fig:overview_selfunc}
\end{figure}

Here, $\kappa=\argmax_k \bar{\mu_k}(x)$. Thus, the risk-averse selector does not change the prediction, only the confidence. 

Another perspective motivating $g_{\text{BPA}}$ comes from Bayesian credible intervals. If we approximate the distribution of the highest-likelihood class as Gaussian — that is, $\mathcal{N}(\bar{\mu}_\kappa, \bar{\sigma}_\kappa^2)$ — then the term $\bar{\mu}_\kappa - \bar{\sigma}_\kappa$ corresponds to approximately the 16th percentile of this distribution. In other words, there is roughly an 84\% posterior probability that the true class probability exceeds $\bar{\mu}_\kappa - \bar{\sigma}_\kappa$. This provides a conservative lower confidence bound that is particularly valuable in high-stakes selective prediction where overconfident predictions carry significant costs. We find that the $1\sigma$ choice empirically balances conservatism against practical utility.

All our selective prediction results with VarVQA use \gsigma. In \Cref{sec:experiments_selfunc}, we provide an ablation against predictive averaging. When it comes to calibration, VarVQA uses predictive averaging, as the subtraction of $\sigma$ leads to systematic underconfidence\footnote{In selective prediction, only relative confidences matter, so there is no negative impact.}. When using MC Dropout with AdamW, we found no systematic benefits of \gsigma. We speculate that this is because the posterior was not actively learned. Thus, we use only \gmpmu ~for Dropout. The selectors used for each method are visually summarized in \Cref{fig:overview_selfunc}.

%% file: sec_tmlr/5_1_experiments_setup.tex
\section{Experiments}\label{sec:experiments}

We describe our experimental setup, models and datasets in \Cref{sec:experiments_setup} and the evaluation metrics in \Cref{sec:experiments_metrics}. Our results show that \approach is effective for multimodal models, more sample-efficient than MC Dropout (\cref{sec:experiments_id}), and more robust to OOD data than AdamW-trained models (\cref{sec:experiments_ood}).
Moreover, our novel selector $g_{\textrm{BPA}}$ outperforms posterior predictive averaging on high-stakes selective prediction (\cf \cref{sec:experiments_metrics}) across multiple models and tasks (\cref{sec:experiments_selfunc}).
    %\item sets new selective prediction SOTAs on VQAv2 (\cref{sec:experiments_prior}).

\subsection{Experimental Setup}\label{sec:experiments_setup}
We explore the effectiveness of \approach on two large VLMs: ViLT \citep{vilt} and BEiT-3 \citep{beit3}. BEiT-3 is near-SOTA\footnote{As of 10/2025, see the VQAv2 Challenge on EvalAI} on VQAv2, but still small enough for full fine-tuning. Both ViLT and BEiT-3 treat VQA as a classification task to 3129 answers, which is standard practice \citep{anderson2018bottom}. In terms of multimodal tasks, we explore VQA (fine-tuning on VQAv2 \citep{vqav2}, evaluation on VQAv2 and AdVQA \citep{advqa}) and Visual Reasoning (fine-tuning and evaluation on NLVR2 \citep{nlvr2}). The publicly available VQAv2 test splits do not include labels, which are required to evaluate calibration and selective prediction (\cf \cref{sec:experiments_metrics}). Therefore, we follow previous work \citep{reliable_vqa} and divide the VQAv2 validation set into dev/val/test. All results are averaged over three training runs with different seeds. Error bars and shaded regions indicate standard error.

\myparagraph{Hyperparameters}
We use the optimal hyperparameters reported in \citep{vilt, beit3} for AdamW. For IVON, most defaults, \cf \citet{ivon}, can be used, but the learning rate and Hessian initialization need to be adjusted. However, we find that due to a strong correlation between the two, the dimensionality of the search space is effectively one. A full account is provided in \Cref{sec:supp_hyperparam_report}.

\myparagraph{Number of Samples}
By default, \approach uses $S=64$ MC samples. We did not find significant improvements beyond this number. For early stopping, we use eight MC samples to save compute.

\myparagraph{Temperature and Vector Scaling}
Previous work \citep{reliable_vqa} has shown that calibrating models with widespread methods like Temperature Scaling \citep{guo2017calibration} and Vector Scaling \citep{platt1999probabilistic} has only a small effect on their selective prediction performance. We confirm these findings and show that the effect is consistently positive, and can be applied on top of any method (\eg AdamW or VarVQA) to receive small additional gains. Full results are in \Cref{sec:supp_calibration}.

\subsection{Evaluation Metrics}\label{sec:experiments_metrics}

\myparagraph{Accuracy}  We work with the standard VQA accuracy \citep{vqa}, which can also take non-integer values (0.3, 0.6, 0.9), besides 0 and 1, if fewer than 4 out of 10 annotators agree. NLVR2 accuracy is binary.

\myparagraph{Calibration}
We evaluate calibration using the Expected Calibration Error (ECE) \citep{naeini2015obtaining, guo2017calibration}, as is standard practice. The ECE is computed by partitioning the model's answer confidences on a dataset $\mathcal{D}$ into $m$ bins $\mathcal{D}_m$, and then summing the bin-wise deviations of confidence from accuracy. We use $m=15$ in our experiments.

\begin{equation}\label{eq:ece}
    \textrm{ECE} = \sum_{m=1}^M \frac{|D_m|}{|D|} \left| \textrm{Acc}(D_m) - g(D_m) \right|.
\end{equation}

\myparagraph{Coverage at Risk}
For the selective prediction metrics, we follow prior work \citep{geifman_selpred, reliable_vqa, lyp}. The standard selective prediction metric is \emph{Coverage at Risk} ($C@R$), which measures the percentage of questions the model is able to answer (\ie where it does not abstain), while keeping the error tolerance $r$ below a given risk level $R$:

\begin{align}\label{eq:cov_at_risk}
    C@R &= \max_{\gamma} C(\gamma) \quad \text{s.t.} \quad r(\gamma) \leq R, \quad \textrm{where we define} \\
    C(\gamma) &= \frac{1}{|D|}\sum_{x\in D}\mathbbm{1}(g(x)\geq \gamma), \quad \textrm{and}\\
    r(\gamma) &= \frac{\frac{1}{|D|}\sum_{x\in D}(1-\mathrm{Acc}(f(x))\mathbbm{1}(g(x)\geq \gamma)}{C(\gamma)}. \\
\end{align}

A larger $C@R$ is better, as a model that abstains on (almost) all inputs is not useful. We also compute the area under the Risk-Coverage curve (AUC) \citep{kamath2020selective}. A weakness of $C@R$ is that the threshold $\gamma$ is determined using the test set. This is necessary as otherwise, a comparison of results would be challenging: For a given risk $R$, one would have to judge both \emph{threshold generalization} (\ie whether the test risk matches the bound $R$), and the achieved test coverage.

\myparagraph{Effective Reliability}
\citet{reliable_vqa} suggested \textit{Effective Reliability} $\Phi_c$ that avoids test set threshold selection. It differs from accuracy by a negative cost $c$ assigned to wrong answers:

\begin{equation}\label{eq:effective_reliability}
    \phi_c(x) = \begin{cases}
                    \textrm{Acc}(x) & \text{if } g(x) \geq \gamma \text{ and } \textrm{Acc}(x) > 0, \\
                    -c & \text{if } g(x) \geq \gamma \text{ and } \textrm{Acc}(x) = 0, \\
                    0 & \text{if } g(x) < \gamma.
                \end{cases}
\end{equation}

The total effective reliability is $\Phi_c=\frac{1}{|\mathcal{D}|}\sum_{x\in\mathcal{D}}\phi_c(x)$, and the abstention threshold $\gamma$ is determined by optimizing $\Phi_c$ on validation data. We report accuracy (Acc), ~$C@R$ and $\Phi_c$ in percent, while keeping the ECE in $[0,1]$ to be consistent with \citet{reliable_vqa}.

\myparagraph{High-Stakes metrics}
Both selective prediction metrics ($C@R$ and $\Phi_c$) feature a parameter that controls the severity of mistakes. Our findings match previous work (\cf Tabs. 1,2 in \citep{reliable_vqa}): Models disproportionately struggle with settings in which errors are very costly (low-$R$, high-$c$)\footnote{The achieved $C@R$ and $\Phi_c$ in these settings are much further below the theoretical optimum than for high $R$/low $c$.}. We collectively refer to these metrics as \emph{high-stakes}. For practical applications, it is arguably more important that models perform well in high-stakes metrics than in low-stakes metrics, since large amounts of errors (even as low as $5\%$) are not acceptable in many real-world scenarios. Moreover, for ID experiments we observe saturation\footnote{For example, BEiT-3 large on VQAv2 achieves $C@10\%>81\%$ and $C@20\%>98\%$.} in low-stakes metrics and thus focus our reported results on high-stakes.

It should be noted that, if stakes are set too high (\ie cost $c$ too high or risk $R$ too small), results can become noisy, as the impact of individual overconfident samples rises. This issue increases with smaller and less well-curated datasets (label noise can have an impact). In our experiments, we observe that the results were stable only up to $c\approx 100$ and down to $R\approx\frac{1}{2}\%$, which is why we stop reporting there.

%% file: sec_tmlr/5_2_experiments.tex
\subsection{In-Distribution Experiments}\label{sec:experiments_id}

\begin{figure}[!tbhp]
  \centering
  \begin{subfigure}{0.24\textwidth}
    \includegraphics[width=\textwidth]{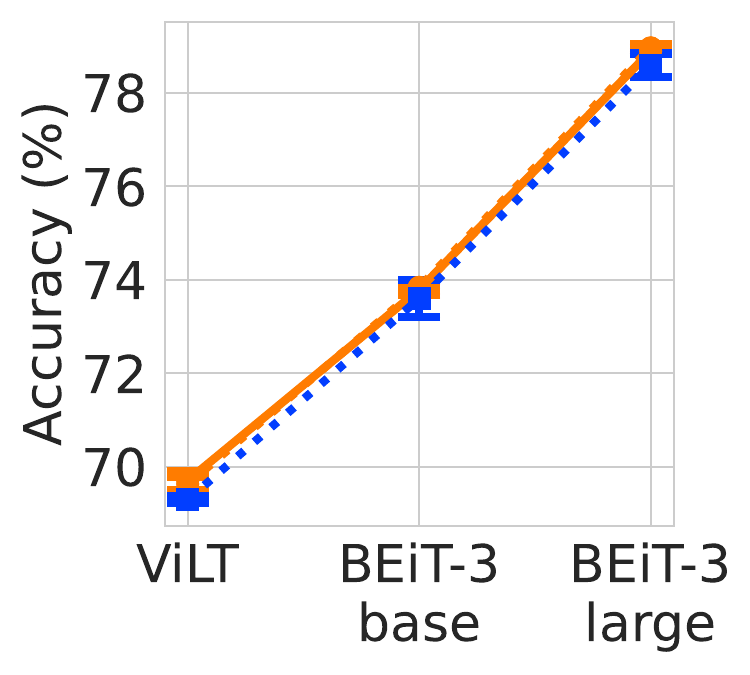}
    \caption{\textbf{Accuracy}}
    \label{fig:all_accuracy_id}
  \end{subfigure}
  \begin{subfigure}{0.24\textwidth}
    \includegraphics[width=\textwidth]{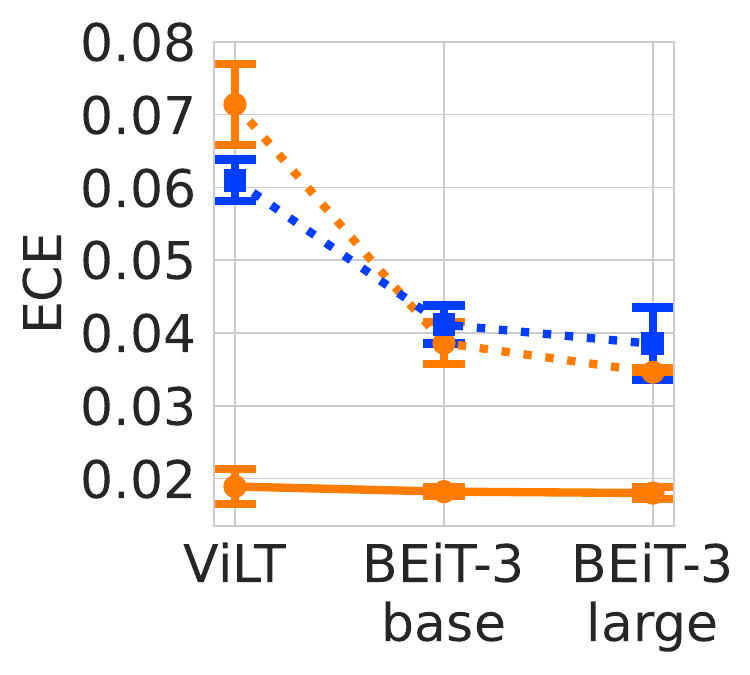}
    \caption{\textbf{Calibration}~($\downarrow$)}
    \label{fig:all_ece_id}
  \end{subfigure}
  \begin{subfigure}{0.24\textwidth}
    \includegraphics[width=\textwidth]{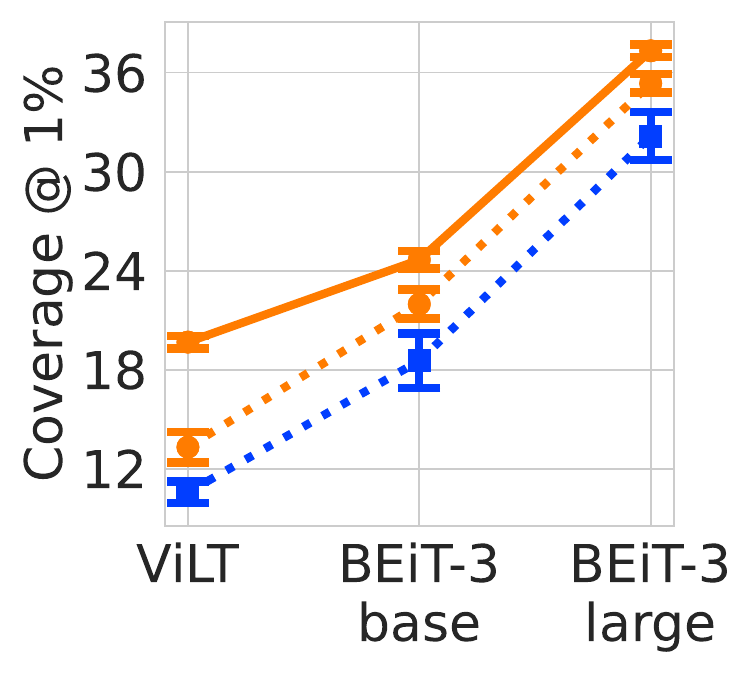}
    \caption{\textbf{Selective Prediction}}
    \label{fig:all_cov1_id}
  \end{subfigure}
  \begin{subfigure}{0.24\textwidth}
    \includegraphics[width=\textwidth]{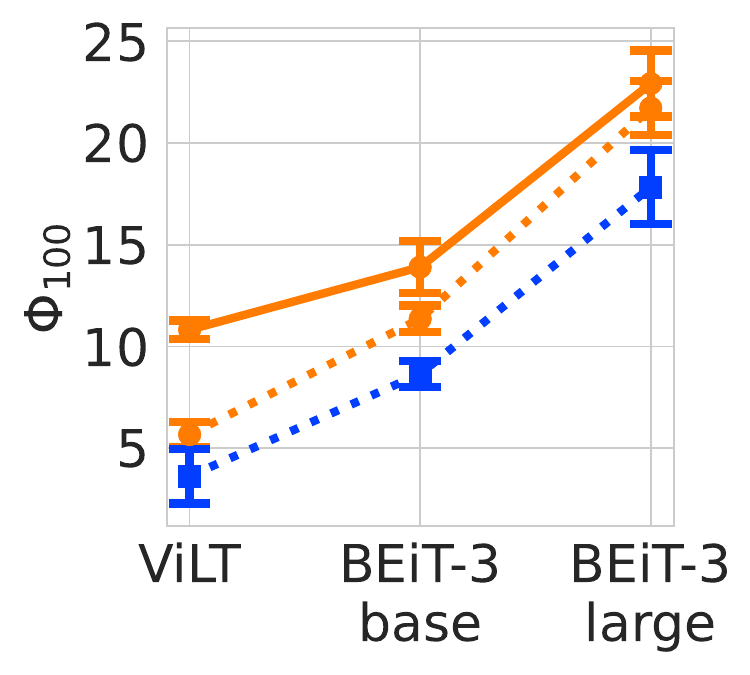}
    \caption{\textbf{Selective Prediction}}
    \label{fig:all_phi100_id}
  \end{subfigure}
  \begin{subfigure}{0.5\textwidth}
    \includegraphics[width=\textwidth]{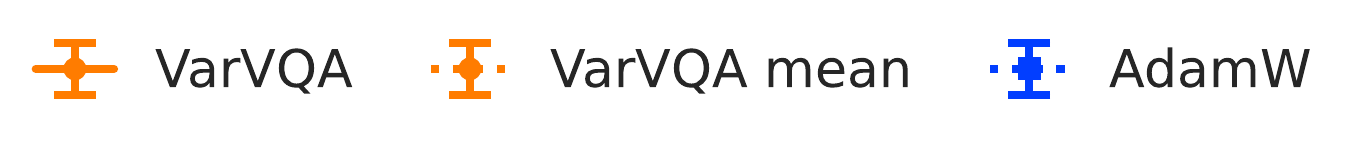}
  \end{subfigure}
  \caption{Accuracy, calibration and selective prediction results for different models after fine-tuning on VQAv2. Error bars indicate standard error across three seeds.}
  \label{fig:all_comparison_id}
\end{figure}

We show ID results after fine-tuning on VQAv2 in \Cref{tab:metrics_id_vqav2} and on NLVR2 (Visual Reasoning) in \Cref{tab:metrics_id_nlvr2}. \Cref{fig:all_comparison_id} visualizes the VQAv2 results. \approach matches the accuracy of the conventional AdamW optimizer (\cref{fig:all_accuracy_id}), while achieving better calibration in terms of lower ECE (\cref{fig:all_ece_id}), and better (high-stakes) selective prediction in terms of higher $C@1\%$ (\cref{fig:all_cov1_id}) and $\Phi_{100}$ (\cref{fig:all_phi100_id}). Additionally, `VarVQA mean' (\cf \cref{sec:var_inference}), is frequently more reliable than AdamW (lower ECE, higher $C@R$, $\Phi_c$), while needing the same inference compute. Finally, the VarVQA sampling strategy consistently outperforms MC Dropout, which uses the same amount of samples at inference, in terms of selective prediction, while achieving a low ECE of $\lesssim 0.03$ throughout and $<0.02$ on VQAv2 with all three tested models. Regarding selective prediction, the benefits of VarVQA over AdamW are largest for the high-stakes metrics. When only one mistake per 200 samples is allowed ($C@\frac{1}{2}\%$), VarVQA on different VLMs improves $7\%-9\%$ on VQAv2 and $9\%-14\%$ on NLVR2 vs. AdamW. When Deep Ensembles \citep{deep_ensembles} are applied on top of an existing method, the reliability improves (\cf \Cref{tab:metrics_ensemble_vqav2}). However, on VQAv2, even vanilla VarVQA is often better than the AdamW Ensemble and the VarVQA Ensemble stays consistently ahead. More in-depth results are in Appendix \Cref{sec:supp_ensembling}. We note that Deep Ensembles cause significant overhead through the necessity of training $N$ models instead of one. (We use $N=3$ due to computational constraints.)

\begin{table}[tbhp]
    \small
    \renewcommand{\arraystretch}{1.2}
    \centering
    \caption{Reliability evaluation on VQAv2 for fine-tuned models. The variable $N$ denotes the number of forward passes. Best results per model are \textbf{bold}.}
    \begin{tabular}{ll|c|c|c|cccc|cc} \toprule
        \multirow{3}{*}{Model} & \multirow{3}{*}{Method} & \multirow{3}{*}{$N$} &  \multirow{3}{*}{\gray{Acc.}} & \multirow{2}{*}{Calibration} & \multicolumn{4}{c|}{Selective Prediction} & \multicolumn{2}{c}{\gray{Sel. Prediction}} \\
        & & & & & \multicolumn{4}{c|}{\textit{high-stakes}} & \multicolumn{2}{c}{\textit{\gray{low-stakes}}} \\
        & & & & ECE~($\downarrow$) & $C@\frac{1}{2}\%$ & $C@1\%$ & $\Phi_{50}$ & $\Phi_{100}$ & \gray{$C@5\%$} & \gray{$\Phi_{10}$}\\ \midrule
        
        \multirow{4}{*}{\centering \;\,ViLT} & AdamW & 1 & \gray{69.30} & 0.061 & 5.03 & 10.58 & 8.41 & 2.89 & \gray{36.24} & \gray{24.05}\\
         & VarVQA mean & 1 & \gray{69.63} & 0.071 & 6.77 & 13.32 & 9.74  & 5.45 & \gray{37.93} & \gray{25.08} \\ \cdashline{2-11}
        & AdamW Dropout & 64 & \gray{69.66} & \textbf{0.019} & 10.44 & 16.63 & 12.51 & 8.44 & \gray{38.49} & \gray{26.18} \\
        & VarVQA & 64 & \gray{69.71} & \textbf{0.019} & \textbf{13.81} & \textbf{19.68} & \textbf{12.93} & \textbf{10.88} & \textbf{\gray{39.53}} & \textbf{\gray{27.15}} \\ \midrule
        
        \multirow{4}{1.1cm}{\centering BEiT-3 base} & AdamW & 1 & \gray{73.60} & 0.041 & 10.35 & 18.55  & 15.59 & 8.65 & \gray{47.93} & \gray{33.40} \\
        & VarVQA mean & 1 & \gray{73.84} & 0.039 & 14.08 & 21.98  & 16.72 & 11.36 & \gray{49.57} & \gray{34.80} \\ \cdashline{2-11}
        & AdamW Dropout & 64 & \gray{73.46} & 0.019 & 13.07 & 20.11  & 16.61 & 9.44 & \gray{47.49} & \gray{33.36} \\
        & VarVQA & 64 & \gray{73.79} & \textbf{0.018} & \textbf{18.10} & \textbf{24.66} &\textbf{19.26} & \textbf{13.90} & \textbf{\gray{49.76}} & \textbf{\gray{35.22}} \\ \midrule
        
        \multirow{4}{1.1cm}{\centering BEiT-3 large} & AdamW & 1 & \gray{78.59} & 0.039 & 21.63 & 32.15 & 26.31 & 17.80 & \gray{63.19} & \gray{45.83} \\
        & VarVQA mean & 1 & \gray{78.96} & 0.035 & 25.32 & 35.35 & 28.31 & 21.25 & \gray{\textbf{64.83}} & \gray{47.43} \\ \cdashline{2-11}
        & AdamW Dropout & 64 & \gray{78.41} & \textbf{0.018} & 25.28 & 34.52  & 27.99 & 20.65 & \gray{63.00} & \gray{46.23} \\
        & VarVQA & 64 & \gray{78.89} & \textbf{0.018} & \textbf{28.13} & \textbf{37.05} & \textbf{29.56} & \textbf{23.21} & \gray{64.68} & \textbf{\gray{48.06}} \\ \bottomrule
        
    \end{tabular}
    \label{tab:metrics_id_vqav2}
\end{table}

\newpage

\begin{table}[tbhp]
    \small
    \renewcommand{\arraystretch}{1.2}
    \centering
    \caption{Reliability evaluation on NLVR2 for fine-tuned models.The variable $N$ denotes the number of forward passes. Best results per model are \textbf{bold}.}
    \begin{tabular}{ll|c|c|c|cccc|cc} \toprule
        \multirow{3}{*}{Model} & \multirow{3}{*}{Method} & \multirow{3}{*}{$N$} & \multirow{3}{*}{\gray{Acc.}} & \multirow{2}{*}{Calibration} & \multicolumn{4}{c|}{Selective Prediction} & \multicolumn{2}{c}{\gray{Sel. Prediction}} \\
        & & & & & \multicolumn{4}{c|}{\textit{high-stakes}} & \multicolumn{2}{c}{\textit{\gray{low-stakes}}} \\
        & & & & ECE~($\downarrow$) & $C@\frac{1}{2}\%$ & $C@1\%$ & $\Phi_{50}$ & $\Phi_{100}$ & \gray{$C@5\%$} & \gray{$\Phi_{10}$}\\ \midrule
        
        \multirow{4}{1.1cm}{\centering BEiT-3 base} & AdamW & 1 & \gray{83.45} & 0.059 & 6.42 & 11.61 & 4.58 & 2.24  & \gray{54.79} & \gray{26.18} \\
        & VarVQA mean & 1 & \gray{83.28} & 0.058 & 5.15 & 15.58 & 6.44 & 1.41 & \gray{55.66} & \gray{27.30} \\ \cdashline{2-11}
        & AdamW Dropout & 64 & \gray{83.18} & \textbf{0.016} & 9.98 & 15.99 & 6.95 & 2.95 & \gray{55.43} & \gray{27.63} \\
        & VarVQA & 64 & \gray{83.11} & 0.031 & \textbf{15.42} & \textbf{23.36} & \textbf{11.20} & \textbf{5.00} & \gray{\textbf{57.16}} & \gray{\textbf{29.23}} \\ \midrule
        
        \multirow{4}{1.1cm}{\centering BEiT-3 large} & AdamW & 1 & \gray{88.34} & 0.041 & 16.53 & 41.14 & 18.08 & 9.45 & \gray{78.53} & \gray{45.64} \\
        & VarVQA mean & 1 & \gray{88.83} & 0.062 & 17.15 & 31.07 & 15.27 & 3.57 & \gray{80.17} & \gray{45.02} \\ \cdashline{2-11}
        & AdamW Dropout & 64 & \gray{88.11} & \textbf{0.017} & \textbf{33.21} & 44.69 & 23.43 & 14.71 & \gray{76.99} & \gray{46.55} \\
        & VarVQA & 64 & \gray{89.26} & 0.029 & 32.89 & \textbf{49.24} & \textbf{25.56} & \textbf{14.85}  & \gray{\textbf{82.11}} & \gray{\textbf{49.51}} \\ \bottomrule
        
    \end{tabular}
    \label{tab:metrics_id_nlvr2}
\end{table}

\begin{table}[tbhp]
    \small
    \renewcommand{\arraystretch}{1.2}
    \centering
    \caption{Comparison of VarVQA to Deep Ensembles \citep{deep_ensembles}, applied to AdamW-trained models and on top of VarVQA; both on VQAv2. The ensembles use three models each.}
    \begin{tabular}{ll|c|c|cccc|cc} \toprule
        \multirow{3}{*}{Model} & \multirow{3}{*}{Method} &  \multirow{3}{*}{\gray{Acc.}} & \multirow{2}{*}{Calibration} & \multicolumn{4}{c|}{Selective Prediction} & \multicolumn{2}{c}{\gray{Sel. Prediction}} \\
        & & & & \multicolumn{4}{c|}{\textit{high-stakes}} & \multicolumn{2}{c}{\textit{\gray{low-stakes}}} \\
        & & & ECE~($\downarrow$) & $C@\frac{1}{2}\%$ & $C@1\%$ & $\Phi_{50}$ & $\Phi_{100}$ & \gray{$C@5\%$} & \gray{$\Phi_{10}$}\\ \midrule
        
        \multirow{3}{*}{\centering \;\,ViLT} 
        % & AdamW & \gray{69.30} & 0.061 & 5.03 & 10.58 & 8.41 & 2.89 & \gray{36.24} & \gray{24.05}\\
        & AdamW Ensemble & \gray{69.69} & 0.049 & 6.97 & 12.30 & 9.97 & 3.67 & \gray{37.86} & \gray{25.14} \\
        & VarVQA & \gray{69.71} & 0.019 & 13.81 & 19.68 & 12.93 & 10.88 & \gray{39.53} & \gray{27.15} \\
        & VarVQA Ensemble & \gray{70.08} & 0.018 & 14.39 & 20.09 & 14.88 & 10.65 & \gray{40.51} & \gray{27.57} \\ \midrule
        
        \multirow{3}{1.1cm}{\centering BEiT-3 base} 
        % & AdamW & \gray{73.60} & 0.041 & 10.35 & 18.55  & 15.59 & 8.65 & \gray{47.93} & \gray{33.40} \\
        & AdamW Ensemble  & \gray{74.70} & 0.018 & 15.84 & 23.66 & 19.19 & 11.82 & \gray{51.33} & \gray{36.65} \\
        & VarVQA & \gray{73.79} & 0.018 & 18.10 & 24.66 &19.26 & 13.90 & \gray{49.76} & \gray{35.22} \\
        & VarVQA Ensemble & \gray{74.18} & 0.015 & 18.34 & 25.70 & 19.27 & 10.98 & \gray{51.16} & \gray{36.01} \\ \midrule
        
        \multirow{3}{1.1cm}{\centering BEiT-3 large} 
        % & AdamW & \gray{78.59} & 0.039 & 21.63 & 32.15 & 26.31 & 17.80 & \gray{63.19} & \gray{45.83} \\
        & AdamW Ensemble & \gray{79.45} & 0.020 & 26.50 & 36.95 & 30.51 & 19.78 & \gray{66.25} & \gray{48.84} \\
        & VarVQA & \gray{78.89} & 0.018 & 28.13 & 37.05 & 29.56 & 23.21 & \gray{64.68} & \gray{48.06} \\
        & VarVQA Ensemble & \gray{79.14} & 0.015 & 28.68 & 37.97 & 30.40 & 23.91 & \gray{65.56} & \gray{48.52} \\ \bottomrule
        
    \end{tabular}
    \label{tab:metrics_ensemble_vqav2}
\end{table}

\begin{figure}[tbhp]
    \centering

  \begin{subfigure}{0.48\textwidth}
    \includegraphics[width=0.48\linewidth]{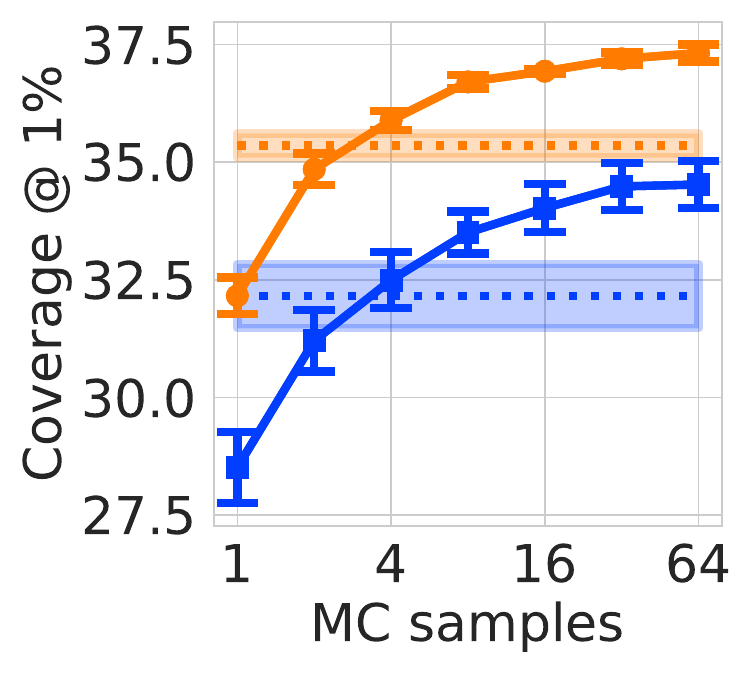}
    \includegraphics[width=0.48\linewidth]{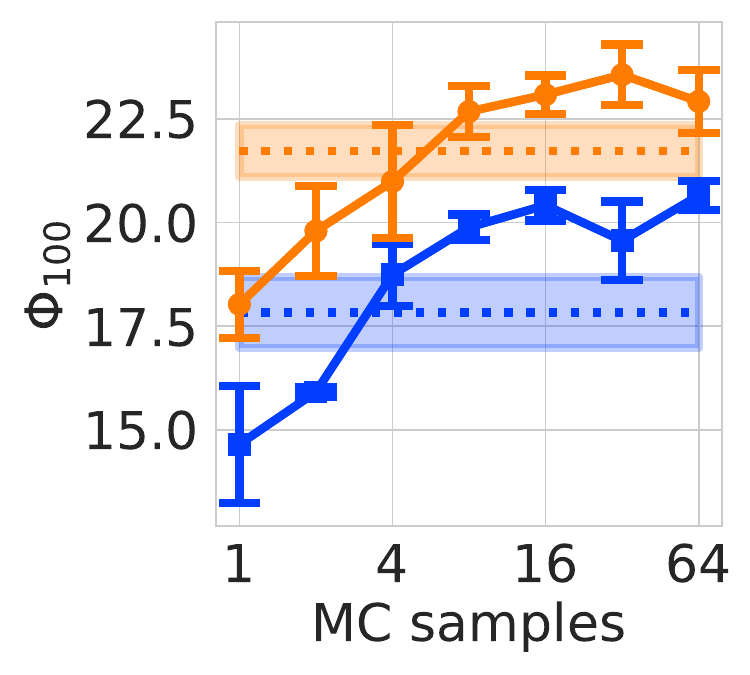}
    \caption{BEiT-3 large, VQAv2}
    \label{fig:mcdropout_b3l_vqa}
  \end{subfigure}
  \begin{subfigure}{0.48\textwidth}
    \includegraphics[width=0.48\linewidth]{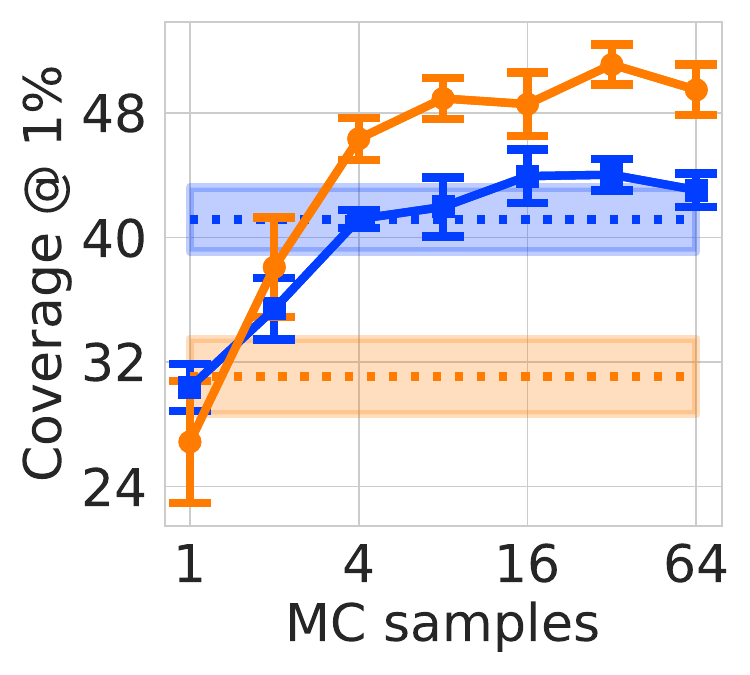}
    \includegraphics[width=0.48\linewidth]{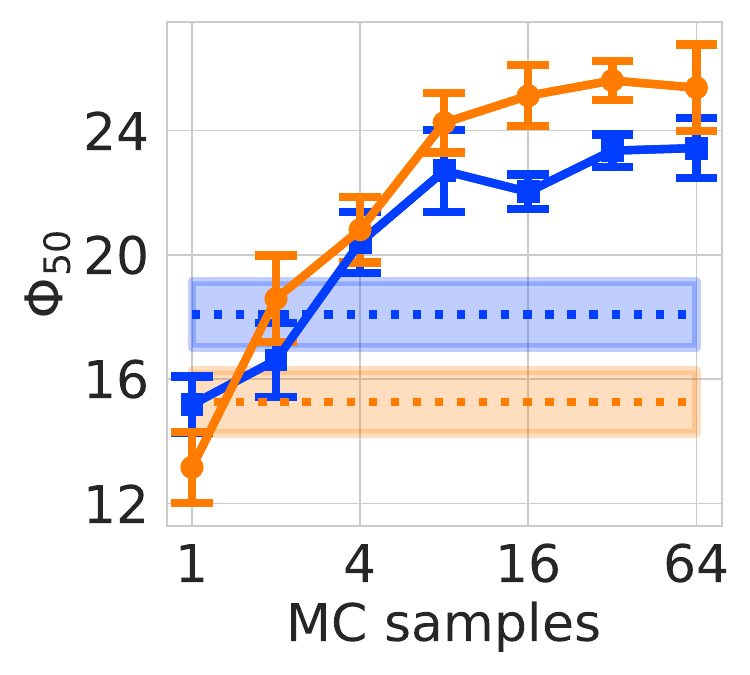}
    \caption{BEiT-3 large, NLVR2}
    \label{fig:mcdropout_b3l_nlvr}
  \end{subfigure}

  \begin{subfigure}{0.48\textwidth}
    \includegraphics[width=0.48\linewidth]{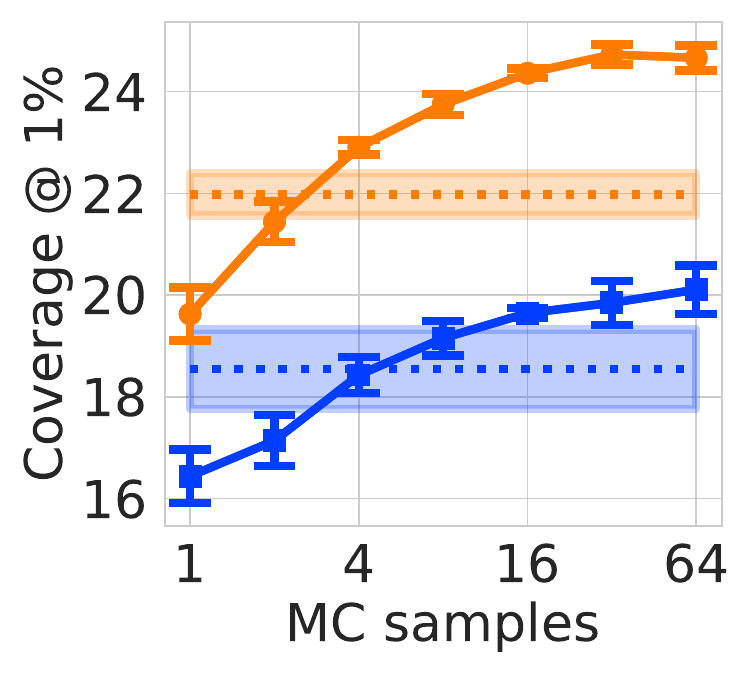}
    \includegraphics[width=0.48\linewidth]{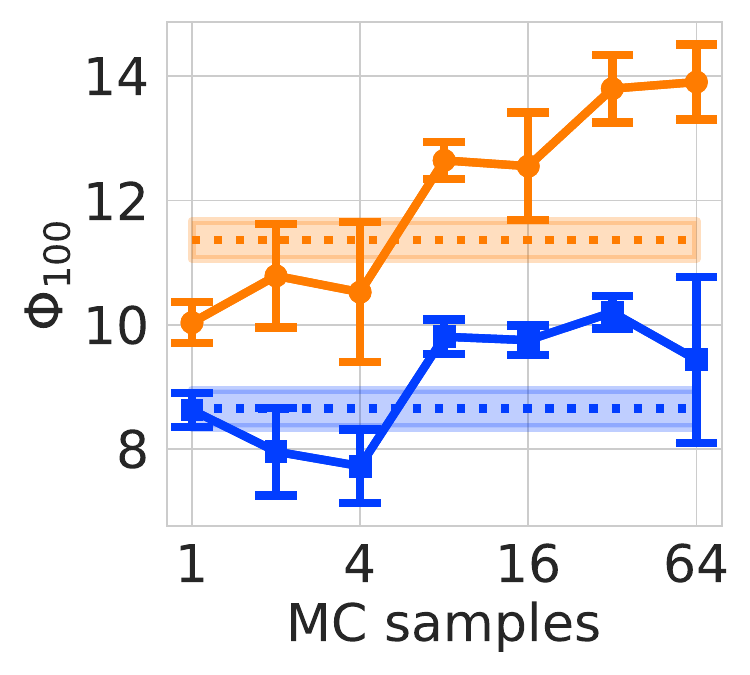}
    \caption{BEiT-3 base, VQAv2}
    \label{fig:mcdropout_b3b_vqa}
  \end{subfigure}
  \begin{subfigure}{0.48\textwidth}
    \includegraphics[width=0.48\linewidth]{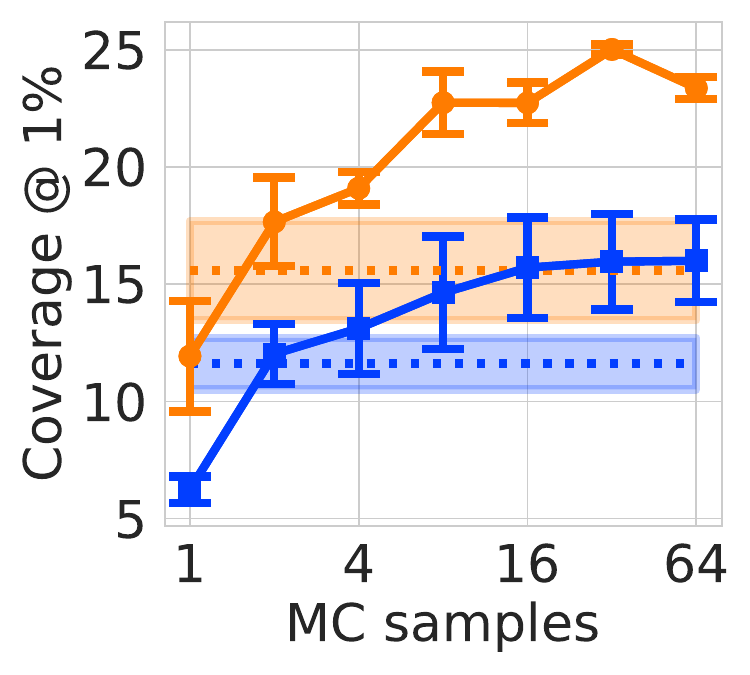}
    \includegraphics[width=0.48\linewidth]{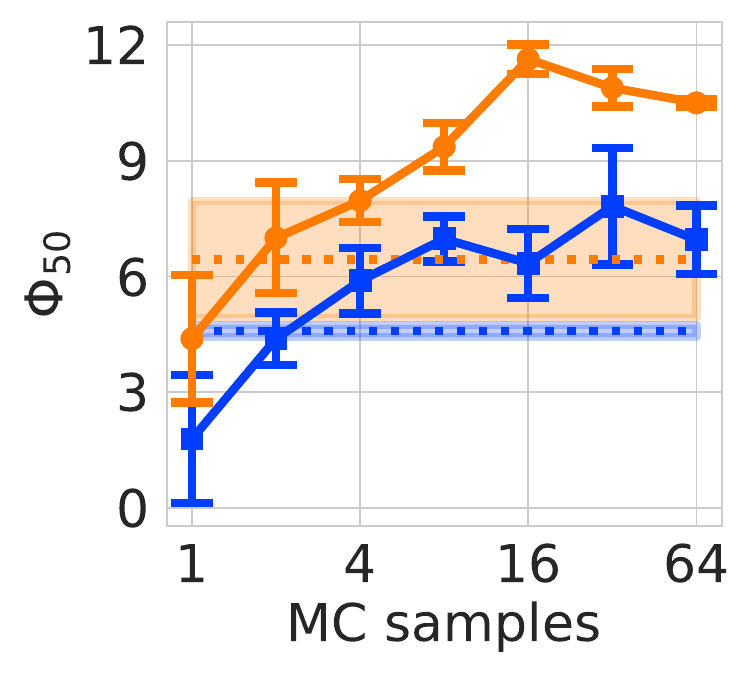}
    \caption{BEiT-3 base, NLVR2}
    \label{fig:mcdropout_b3b_nlvr}
  \end{subfigure}
  \begin{subfigure}{0.7\textwidth}
    \includegraphics[width=\textwidth]{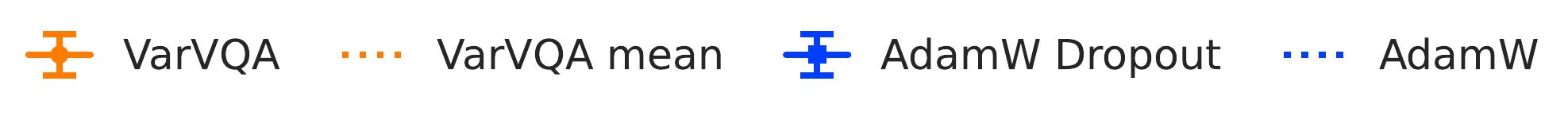}
  \end{subfigure}
  
  \caption{Comparison of \approach to MC Dropout, which uses the same inference compute, on high-stakes selective prediction. Error bars and shaded regions indicate standard error across three seeds.}
  
  \label{fig:mcdropout}
\end{figure}

\subsection{Mixed ID/OOD Experiments}\label{sec:experiments_ood}

\begin{figure}[tbhp]
  \centering
  \begin{subfigure}{0.23\textwidth}
    \includegraphics[width=\textwidth]{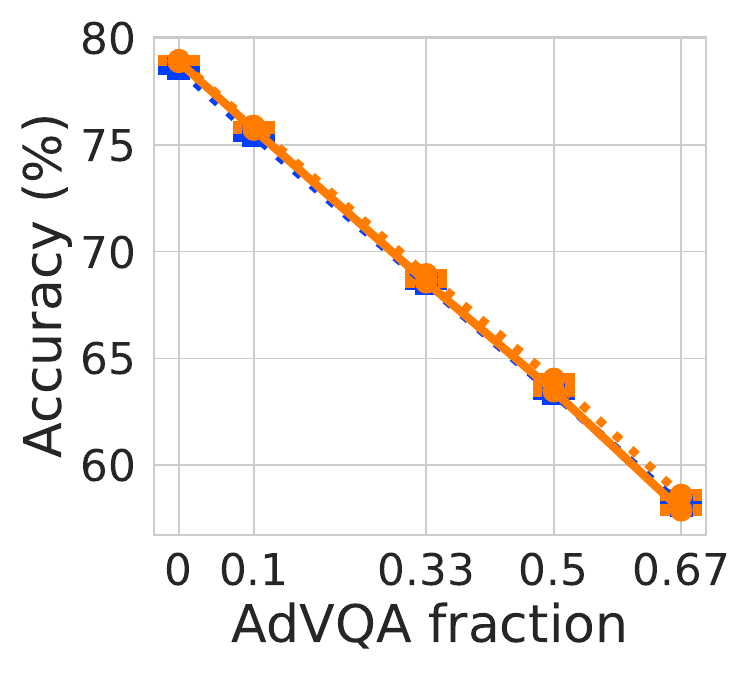}
    \caption{\textbf{Accuracy}}
    \label{fig:accuracy_ood}
  \end{subfigure}
  \begin{subfigure}{0.23\textwidth}
    \includegraphics[width=\textwidth]{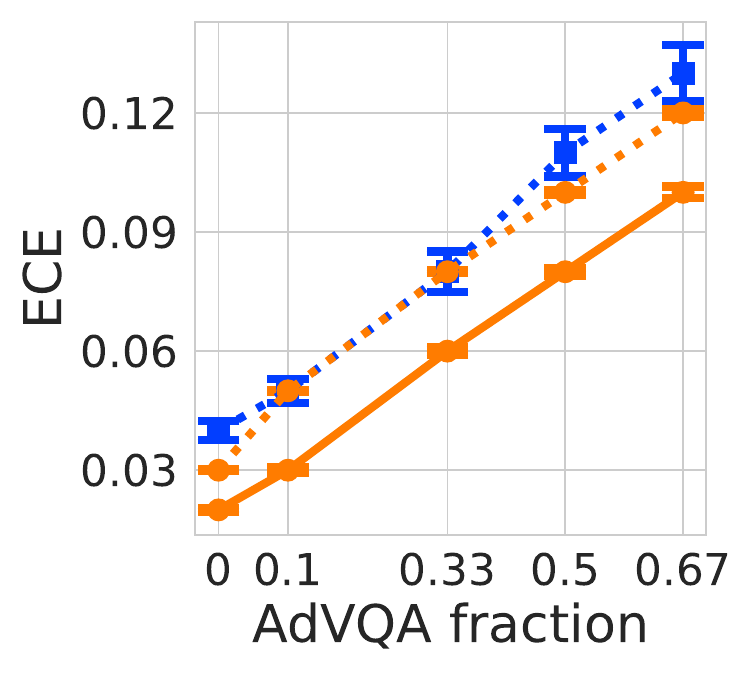}
    \caption{\textbf{Calibration}~($\downarrow$)}
    \label{fig:ece_ood}
  \end{subfigure}
  \begin{subfigure}{0.23\textwidth}
    \includegraphics[width=\textwidth]{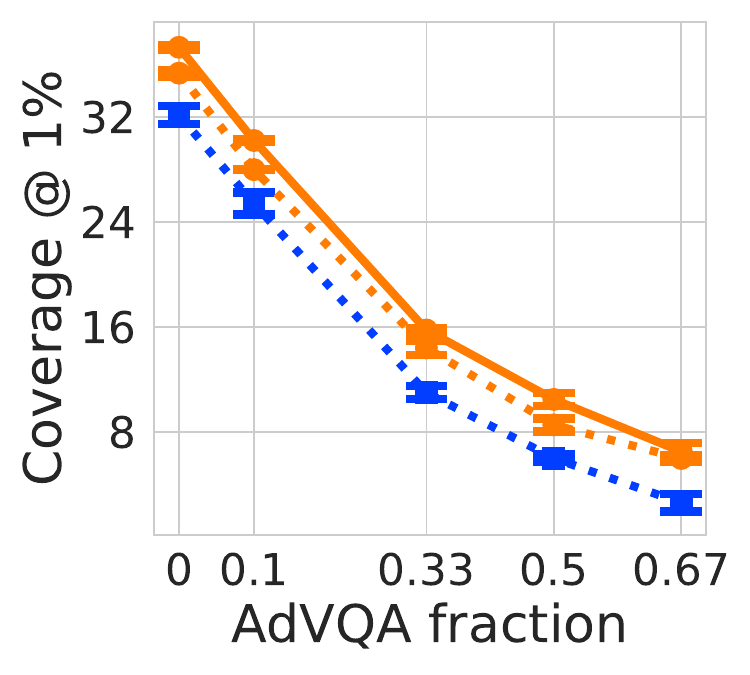}
    \caption{\textbf{Selective Prediction}}
    \label{fig:cov001_ood}
  \end{subfigure}
  \begin{subfigure}{0.23\textwidth}
    \includegraphics[width=\textwidth]{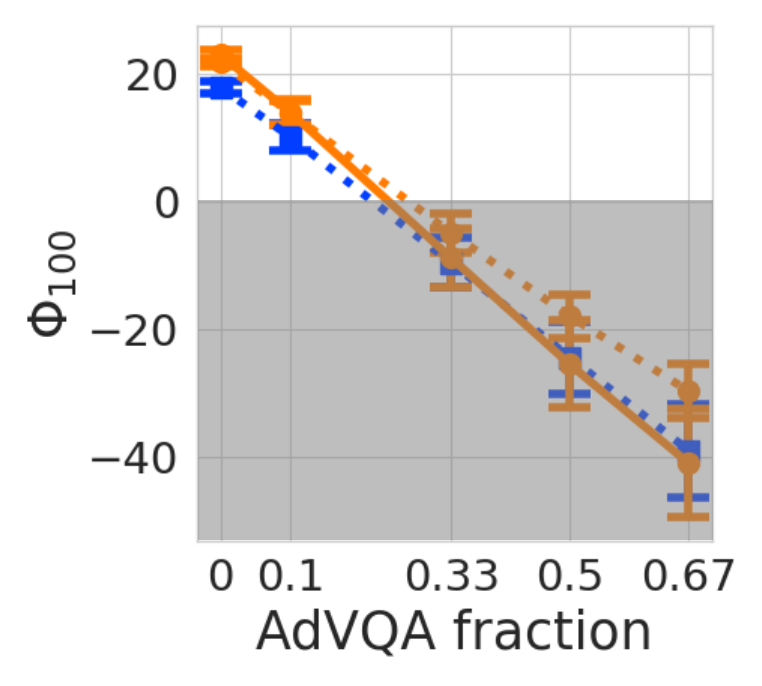}
    \caption{\textbf{Selective Prediction}}
    \label{fig:phi100_ood}
  \end{subfigure}
  \begin{subfigure}{0.5\textwidth}
    \includegraphics[width=\textwidth]{figures/legends/3_horizontal.pdf}
  \end{subfigure}
  \caption{Accuracy, calibration and selective prediction results for different VQAv2/AdVQA mixtures for BEiT-3 large. Error bars indicate standard error across three seeds. In \textbf{(d)}, every model in the gray area is worse than a model that abstains on every input.}
  \label{fig:ood}
\end{figure}

Following \citep{lyp}, we use VQAv2 \citep{vqav2} and AdVQA \citep{advqa} as ID and OOD datasets, respectively. Both datasets use COCO images \citep{coco}, but AdVQA has a different multimodal distribution (more challenging questions). We use the splits from \citep{lyp}, which draw testing data from $P_{\textrm{mix}}$, where 

\begin{equation}\label{eq:ood_mix}
    P_{\textrm{mix}} = \alpha P_{\textrm{OOD}} + (1 - \alpha) P_{\textrm{ID}},
\end{equation}

using $P_{\textrm{ID}}=\textrm{VQAv2}$ and $P_{\textrm{OOD}}=\textrm{AdVQA}$. Different mixtures are obtained by varying $\alpha\in [0,1]$. \Cref{fig:ood} shows the results for BEiT-3 large. Although the accuracy drops equally fast for all methods, \approach remains better calibrated (\cref{fig:ece_ood}). The decline in $C@1\%$ is equal in absolute numbers (\cref{fig:cov001_ood}), but this implies that the relative performance of VarVQA vs. AdamW is increasing at higher OOD fractions. Thus, there is reason to believe that \approach may be fundamentally more robust to OOD data than AdamW-trained models. The results for the other models and metrics are in \Cref{sec:supp_full_results}.

\subsection{Beyond Predictive Averaging}\label{sec:experiments_selfunc}

We compare the performance of our novel selector \gsigma ~(\cf \cref{sec:bpa}) to the baseline $g^{\mu}_{\textrm{MP}}$ ~(\cf \cref{sec:baseline_selectors}). The full results are shown in \Cref{tab:conffunc_comparison_vqav2,tab:conffunc_comparison_nlvr2}. For the high-stakes selective prediction metrics, \gsigma ~consistently outperforms the sample averaging of \gmpmu, achieving \eg $5\%$ higher $C@\frac{1}{2}\%$ on NLVR2 for BEiT-3 base. For the mostly saturated low-stakes selective prediction metrics (grayed), there is no clear winner. When using MC Dropout, we did not find any systematic improvement of \gsigma ~over \gmpmu.

\begin{table}[!tbhp]
    \renewcommand{\arraystretch}{1.2}
    \centering
    \caption{Comparison of our risk-averse selection function \gsigma ~(\cref{eq:g_custom_var}) against \gmpmu ~on VQAv2 with VarVQA ($N=64$ samples as always). Best results per model are \textbf{bold}.} 
    \begin{tabular}{@{\ }l|ll|cccc|cc@{\ }}
    \toprule
    \multirow{2}{*}{Dataset} & \multirow{2}{*}{Model} & \multirow{2}{*}{Selector} & \multicolumn{4}{c|}{\textit{high-stakes}} & \multicolumn{2}{c}{\gray{\textit{low-stakes}}} \\
    & & & $C@ \frac{1}{2}\%$ & $C@ 1\%$ & $\Phi_{50}$ & $\Phi_{100}$& \gray{$C@ 5\%$} & \gray{$\Phi_{10}$} \\ 
    \midrule
    \multirow{6}{*}{VQAv2} & \multirow{2}{*}{ViLT} & \gmpmu & 13.35 & 19.24 & \textbf{13.04} & 10.05 & \gray{39.52} & \gray{26.64} \\
    & & \gsigma & \textbf{13.81} & \textbf{19.68} & 12.93 & \textbf{10.88} & \gray{\textbf{39.53}} & \gray{\textbf{27.15}} \\ 
    \cmidrule{2-9}
    & \multirow{2}{*}{BEiT-3 base} & \gmpmu & 17.15 & 23.87 & 18.64 &  12.23 & \gray{\textbf{49.91}} & \gray{35.17} \\
    & & \gsigma & \textbf{18.10} & \textbf{24.66} & \textbf{19.26} & \textbf{13.90} & \gray{49.76} & \gray{\textbf{35.22}} \\ 
    \cmidrule{2-9}
    & \multirow{2}{*}{BEiT-3 large} & \gmpmu & 27.09 & 36.00 & 28.82 & 22.14 & \gray{\textbf{64.82}} & \gray{47.58} \\
     & & \gsigma & \textbf{28.13} & \textbf{37.05} & \textbf{29.56} & \textbf{23.21} & \gray{64.68} & \gray{\textbf{48.06}} \\
    \bottomrule
    \end{tabular}
    
    \label{tab:conffunc_comparison_vqav2}
\end{table}

\begin{table}[tbhp]
    \renewcommand{\arraystretch}{1.2}
    \centering
    \caption{Comparison of our risk-averse selection function \gsigma ~(\cref{eq:g_custom_var}) against \gmpmu ~on NLVR2 with VarVQA ($N=64$ samples as always). Best results per model are \textbf{bold}.} 
    \begin{tabular}{@{\ }l|ll|cccc|cc@{\ }}
    \toprule
    \multirow{2}{*}{Dataset} & \multirow{2}{*}{Model} & \multirow{2}{*}{Selector} & \multicolumn{4}{c|}{\textit{high-stakes}} & \multicolumn{2}{c}{\gray{\textit{low-stakes}}} \\
     & & & $C@ \frac{1}{2}\%$ & $C@ 1\%$ & $\Phi_{50}$ & $\Phi_{100}$& \gray{$C@ 5\%$} & \gray{$\Phi_{10}$} \\ 
     \midrule
    \multirow{4}{*}{NLVR2} & \multirow{2}{*}{BEiT-3 base} & \gmpmu & 10.64 & 22.20 & 9.75 & 3.95 & \gray{\textbf{57.18}} & \gray{\textbf{29.28}} \\
    & & \gsigma & \textbf{15.42} & \textbf{23.36} & \textbf{11.20} & \textbf{5.00} & \gray{57.16} & \gray{29.23} \\ 
    \cmidrule{2-9}
    & \multirow{2}{*}{BEiT-3 large} & \gmpmu & 27.61 & 48.16 & 24.26 & 13.59 & \gray{\textbf{82.16}} & \gray{\textbf{49.51}} \\
    & & \gsigma & \textbf{32.89} & \textbf{49.24} & \textbf{25.56} & \textbf{14.85} &  \gray{82.11} & \gray{\textbf{49.51}} \\ 
    \bottomrule
    \end{tabular}
    
    \label{tab:conffunc_comparison_nlvr2}
\end{table}

 %~(\cf \Cref{sec:supp_gbpa_dropout}), possibly because the output variances originate from an ad-hoc posterior. In contrast, when the posterior distribution over parameters is learned, \eg with IVON, the output variances benefit and carry meaningful information that can improve abstention decisions.

\subsection{Qualitative Results}\label{sec:experiments_qual}

\begin{figure}[tbhp]
    \includegraphics[width=\textwidth]{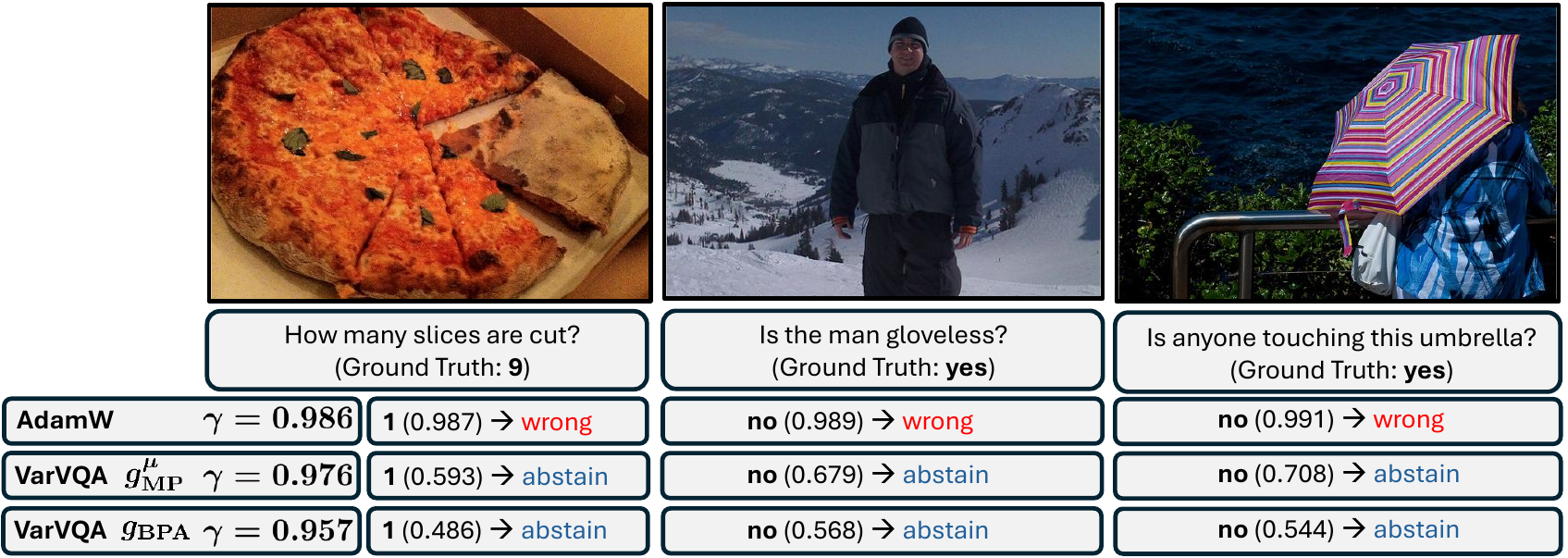}
    \caption{Qualitative examples on VQAv2 with BEiT-3 large where AdamW is wrong while VarVQA abstains. The abstention thresholds $\gamma$ were determined by optimizing $\Phi_{100}$ on VQAv2 validation data. Model answers are displayed in \textbf{bold}, the corresponding answer confidences are provided in brackets.}
    \label{fig:qual_adam_wrong_ivon_abstain_vqa}
\end{figure}

\begin{figure}[tbhp]
    \centering
    \includegraphics[width=\textwidth]{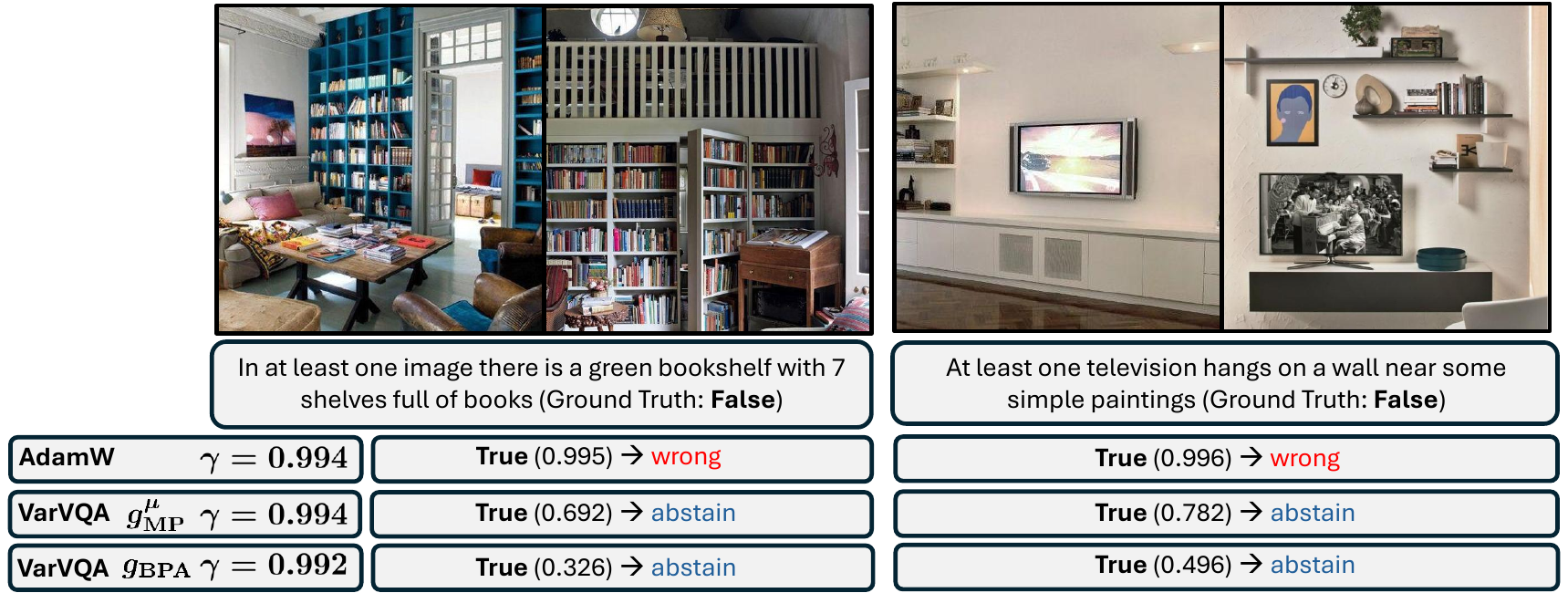}
    \caption{Qualitative examples on NLVR2 with BEiT-3 large where AdamW is wrong while VarVQA abstains. The abstention thresholds $\gamma$ were determined by optimizing $\Phi_{100}$ on NLVR2 validation data. Model answers are displayed in \textbf{bold}, the corresponding answer confidences are provided in brackets.}
    \label{fig:qual_adam_wrong_ivon_abstain_nlvr}
\end{figure}

We show qualitative examples that highlight the difference in uncertainty estimates between AdamW and \approach in \Cref{fig:qual_adam_wrong_ivon_abstain_vqa,fig:qual_adam_wrong_ivon_abstain_nlvr}. Further qualitative examples for VQAv2, AdVQA and NLVR2, including failure cases, can be found in \Cref{sec:supp_qualitative}. As the accuracy of the AdamW- and IVON-trained models is similar, we focus on cases where they predict the same answer, as this reflects the typical behavior. The key improvement of VarVQA lies not in better accuracy, but rather in improved uncertainty estimates. A further study that investigates the behavior on the different question categories of VQAv2 and AdVQA (\emph{Binary}, \emph{Number}, and \emph{Other}), can also be found in \Cref{sec:supp_qualitative}.

%% file: sec_tmlr/6_conclusion.tex
\section{Discussion}
\label{sec:conclusion}
In this work, we explore \approach, \ie the application of Variational Learning for multimodal tasks. Our implementation replaces the standard AdamW optimizer with the IVON method and uses multiple samples from the learned posterior at inference. In addition, our new selector goes beyond the standard predictive averaging by incorporating the output's variance into the abstention decision. Our findings demonstrate that \approach has two possible applications: when inference costs should be minimal, parameter means can be used at inference to at least match the accuracy of AdamW and decently increase reliability. When higher inference costs are acceptable, multiple MC samples from the posterior can be used. Better reliability is demonstrated by better calibration as well as better selective prediction, both in distribution for multiple tasks, and in the challenging mixed ID/OOD setting. The novel selector further improves selective prediction in high-stakes settings with almost no computational overhead.

Variational VQA also has some limitations, particularly involving hyperparameter tuning with IVON. While we observe correlations between the critical hyperparameters (discussed in the Appendix), which can be exploited to reduce the search space, tuning still remains more involved than with AdamW. Additionally, while VarVQA makes large gains in high-stakes selective prediction vs. AdamW, overconfidence still remains an issue, and Coverages remain well below the theoretical optimum ($\approx Acc.$ for low risks). Thus, more work is needed to make models truly `know what they do not know'.

An exciting avenue for future work is to avoid the computational burden of sampling for VarVQA by variance propagation in one forward pass. Recently, \citet{li2024streamlining} proposed a new method in this domain that has shown promising results for unimodal tasks with IVON. Such `streamlining' is only possible if learned parameter variances are available, which is not the case for \eg MC Dropout. While \approach intrinsically improves reliability, the incorporation of previous methods through \eg  training a (variational) selector on top of the (variational) model, could also further enhance reliability. Improvements could also be obtained by using a more expressive posterior that uses full covariance, however at the time of writing, there is no practical alternative to IVON for large models that uses full covariance.

%% file: sec_tmlr/7_acknowledgements.tex
\myparagraph{Acknowledgements} This research was partially funded by an Alexander von Humboldt Professorship in Multimodal Reliable AI, sponsored by Germany’s Federal Ministry for Research, Technology and Space and by a LOEWE-Spitzen-Professur (LOEWE/4a//519/05.00.002(0010)/93). Mohammad Emtiyaz Khan was supported by the Bayes duality project, JST CREST Grant Number JPMJCR2112. The work has benefited from the Excellence Cluster “Reasonable AI” by the Deutsche Forschungsgemeinschaft (DFG, German Research Foundation) under Germany's Excellence Strategy – EXC-3057. For compute, we gratefully acknowledge support from the hessian.AI Service Center (funded by the Federal Ministry of Research, Technology and Space, BMFTR, grant no. 16IS22091) and the hessian.AI Innovation Lab (funded by the Hessian Ministry for Digital Strategy and Innovation, grant no. S-DIW04/0013/003).

%% file: sec_tmlr/A_hyperparam_report.tex
\clearpage
\maketitle
\renewcommand{\theenumi}{\Alph{enumi}} 

\section*{\LARGE Supplement}

\begin{itemize}

    \setlength{\itemsep}{1em} 

    \item \Cref{sec:supp_hyperparam_report}: Hyperparameters for training and inference.

    \item \Cref{sec:supp_trainingtime}: Training and inference time differences between AdamW and VarVQA.

    \item \Cref{sec:supp_calibration}: The impact of common calibration methods on the baseline and on VarVQA.

    %\item \Cref{sec:supp_gbpa_dropout}: Evaluating the new risk-averse selector \gsigma ~on MC Dropout.

    %\item In \Cref{sec:supp_confidencefunctions}, we further analyze the confidence functions $g_{\mu^*}$ and $g_{\mu^*-\sigma^*}$ and present the proof for the claim in \Cref{sec:experiments_selfunc} of the main paper.

    \item \Cref{sec:supp_full_results}: Extended results from the main paper.

    %\item In \Cref{sec:supp_priorwork} we show an extended comparison to prior work (\cf \cref{sec:experiments_prior} in the main paper).

    \item \Cref{sec:supp_qualitative}: More qualitative examples, including failure cases.

    \item \Cref{sec:supp_ensembling}: More experiments on applying Deep Ensembles \citep{deep_ensembles} to our VarVQA method and the baselines.

    \item \Cref{sec:thresh_gen}: Measuring \textit{threshold generalization} (\cref{sec:experiments_metrics}, see \textit{Coverage at Risk}), \ie how close the test risk is to the target, given a validation-selected abstention threshold.

    \item \Cref{sec:selec_comp}: Comparing VarVQA to using a task-specific selector head \citep{reliable_vqa}, which requires an additional training phase.

\end{itemize}

\section{Experimental Details and Hyperparameters}\label{sec:supp_hyperparam_report}

All models were trained on a single server with 8 NVIDIA A100-80GB GPUs. For BEiT-3 \citep{beit3}, we use the official implementation on GitHub, whereas for ViLT \citep{vilt} we use the huggingface implementation. For early stopping, we consistently use $C@(1-5)\% = \frac{1}{5}\sum_{i=1}^{5} C@i$, which focuses on small risks (high-stakes). This is because we find that early stopping for accuracy or validation loss often selects an epoch that is already starting to lose performance in the high-stakes selective prediction metrics, both for AdamW and IVON. In general, we consistently observe that these high-stakes metrics suffer from overfitting first, followed by low-stakes selective prediction later, and accuracy declining last (=latest) in training. Thus, \textit{early stopping during fine-tuning is crucial for optimal reliability}.

\paragraph{AdamW Training.} We only make small changes compared to the default hyperparameters; the details are listed in \Cref{tab:supp_hyperparameters_adam}.
\begin{itemize}
    \item As the default ViLT implementation has dropout = 0, we performed a hyperparameter search to find the optimal lr-dropout combination, which resulted in a slightly lower learning rate than the default ($3\cdot10^{-5}$ vs. $10^{-4}$).
    \item BEiT-3 large is trained in mixed precision (bf16).
    \item Modest gradient clipping is added for all models.
\end{itemize}

\paragraph{IVON Training.} We generally follow \citet{ivon} for the initial selection of all IVON-specific hyperparameters. The specific hyperparameter settings for IVON are listed in \Cref{tab:supp_hyperparameters_ivon}. Our high-level findings and guidelines are as follows.
\begin{itemize}
    \item IVON needs gradient clapping for stability, as with no clipping, the Hessian estimate will frequently diverge.
    \item The gradient clipping for IVON needs to be slightly higher than that of AdamW, as AdamW typically produces smaller gradients.
    \item All IVON hyperparameters except the learning rate (lr) and $h_0$ can be left at default values.
    \item There exists a correlation between lr and $h_0$, \ie a smaller lr requires a larger $h_0$ for optimal results and vice versa. This correlation is approximately linear for our three models and VQAv2 training: $\mathrm{lr}\cdot h_0 = 0.01$ was almost always optimal.
\end{itemize}

To find the optimal IVON hyperparameters, we performed a (Bayesian) hyperparameter search. 

\begin{table*}
    \caption{Hyperparameters for AdamW finetuning on VQAv2 (\emph{bsz}: batch size, \emph{clip}: gradient clipping norm, \emph{lr}: learning rate, \emph{$\delta$}: weight decay). Warmup epochs are in brackets. *For BEiT, drop path is used (dropout=0).}
    \small
    \renewcommand{\arraystretch}{1.1}
    \centering
    \begin{tabular}{l|ccccl|cccc} 
    \toprule
     \multirow{2}{*}{\textbf{Model}} & \multicolumn{5}{c|}{\textbf{General hyperparam.}} & \multicolumn{4}{c}{\textbf{Optimizer-specific hyperparam.}} \\ \cmidrule{2-10}
     & precision & bsz & epochs & clip & dropout & lr & $\delta$ & $\beta_1$ & $\beta_2$\\ 
    \midrule
    ViLT  & fp32 & 256 & 10 (1) & 10 & 0.10 & $3\cdot10^{-5}$ & 0.01 & 0.9 & 0.999\\
    BEiT-3 base & fp32& 128 & 10 (1) & 10 & 0.10* & $3\cdot10^{-5}$ & 0.01 & 0.9 & 0.98\\
    BEiT-3 large & amp (bf16) & 128 & 10 (1) & 20 & 0.15* & $2\cdot10^{-5}$ & 0.01 & 0.9 & 0.98\\
    \bottomrule
    \end{tabular}
    
    \label{tab:supp_hyperparameters_adam}
\end{table*}

\begin{table*}
    \caption{Hyperparameters for IVON finetuning on VQAv2. (\emph{bsz}: batch size, \emph{clip}: norm for gradient clipping, \emph{lr}: learning rate, \emph{$\delta$}: weight decay, \emph{$h_0$}: Hessian initialization, \emph{$\lambda$}: size of training set, \emph{$R_{\textrm{clip}}$}: radius for gradient clipping). Warmup epochs are in brackets. *For BEiT, drop path is used (dropout=0).}
    \small
    \renewcommand{\arraystretch}{1.15}
    \centering
    \setlength{\tabcolsep}{5pt}
    \begin{tabular}{l|ccccl|ccccccc} 
    \toprule
    \multirow{2}{*}{\textbf{Model}} & \multicolumn{5}{c|}{\textbf{General hyperparam.}} & \multicolumn{7}{c}{\textbf{Optimizer-specific hyperparam.}} \\ \cmidrule{2-13}
    & precision & bsz & epochs & clip & dropout & lr & $\delta$ & $\beta_1$ & $\beta_2$ & $h_0$ & $\lambda$ & $R_{\textrm{clip}}$ \\ 
    \midrule
    ViLT  & fp32 & 256 & 10 (1) & 25 & 0.05 & 0.2 & $5\cdot10^{-5}$ & 0.9 & 0.99995 & 0.05 & $5\cdot10^5$ & 0.001 \\
    BEiT-3 base & fp32 & 128 & 10 (1) & 25 & 0.10* & 0.02 & $5\cdot10^{-5}$ & 0.9 & 0.99995 & 0.5 & $5\cdot10^5$ & 0.001      \\
    BEiT-3 large & amp (bf16) & 128 & 10 (1) & 50 & 0.15* & 0.02 & $5\cdot10^{-5}$ & 0.9 & 0.99995 & 0.5 & $5\cdot10^5$ & 0.001         \\
    \bottomrule
    \end{tabular}
    
    \label{tab:supp_hyperparameters_ivon}
\end{table*}

\paragraph{MC Dropout.} For our comparison to MC Dropout (\cf \cref{sec:experiments_id}), we tune the dropout rate for ViLT, both for AdamW and for IVON, where we discovered that combining MC Dropout with sampling at inference can provide modest benefits. Thus, all ViLT results for \approach were obtained using MC Dropout together with MC Sampling from the learned posterior at inference. As BEiT-3 already provides a default dropout rate, we use it for AdamW and IVON training, and also for AdamW inference. Unlike ViLT, BEiT-3 with IVON did not improve when using MC Dropout at inference on top of sampling. We leave it to future work to further investigate the exact relationship of variational inference and MC Dropout with IVON.

%% file: sec_tmlr/B_time.tex
\section{Training and Inference Time}\label{sec:supp_trainingtime}
\begin{figure}[tbhp]
    \centering
  \begin{subfigure}[t]{0.49\textwidth}
    \centering
    \includegraphics[width=\linewidth]{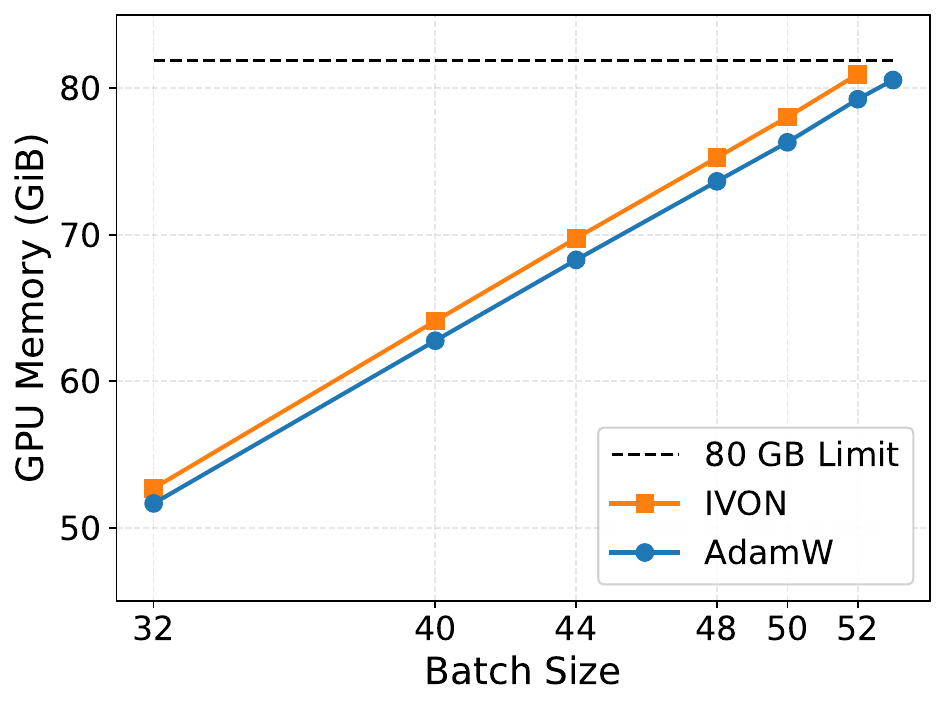}
    \caption{Peak GPU memory for training BEiT-3 base.}
    \label{fig:gpu_mem_base}
  \end{subfigure}
  \hfill
  \begin{subfigure}[t]{0.49\textwidth}
    \centering
    \includegraphics[width=\linewidth]{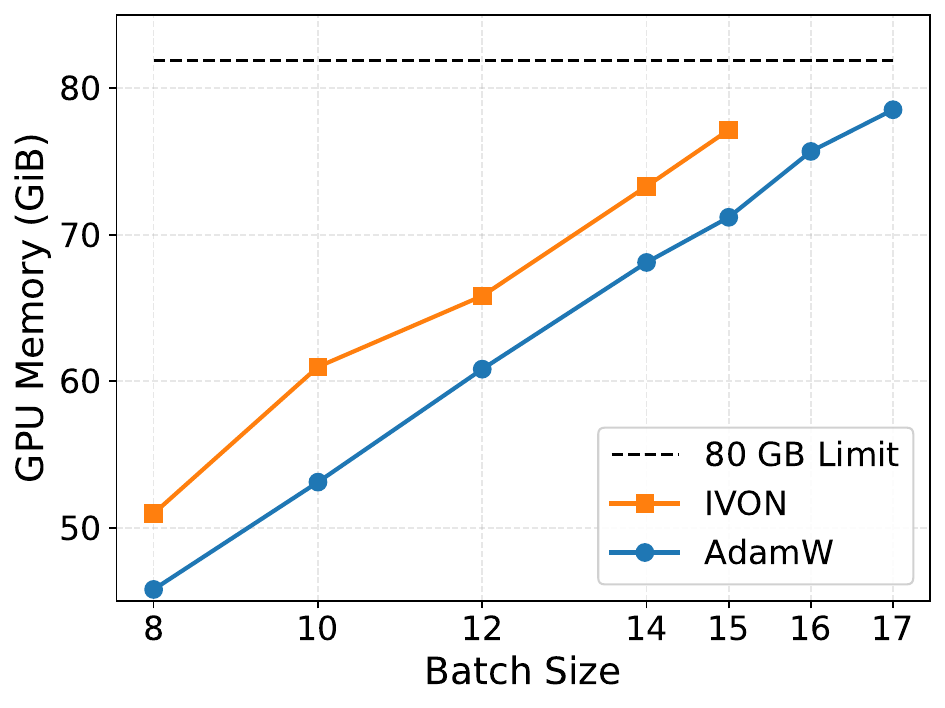}
    \caption{Peak GPU memory for training BEiT-3 large.}
    \label{fig:gpu_mem_large}
  \end{subfigure}
  
  \begin{subfigure}[t]{0.49\textwidth}
    \centering
    \includegraphics[width=\linewidth]{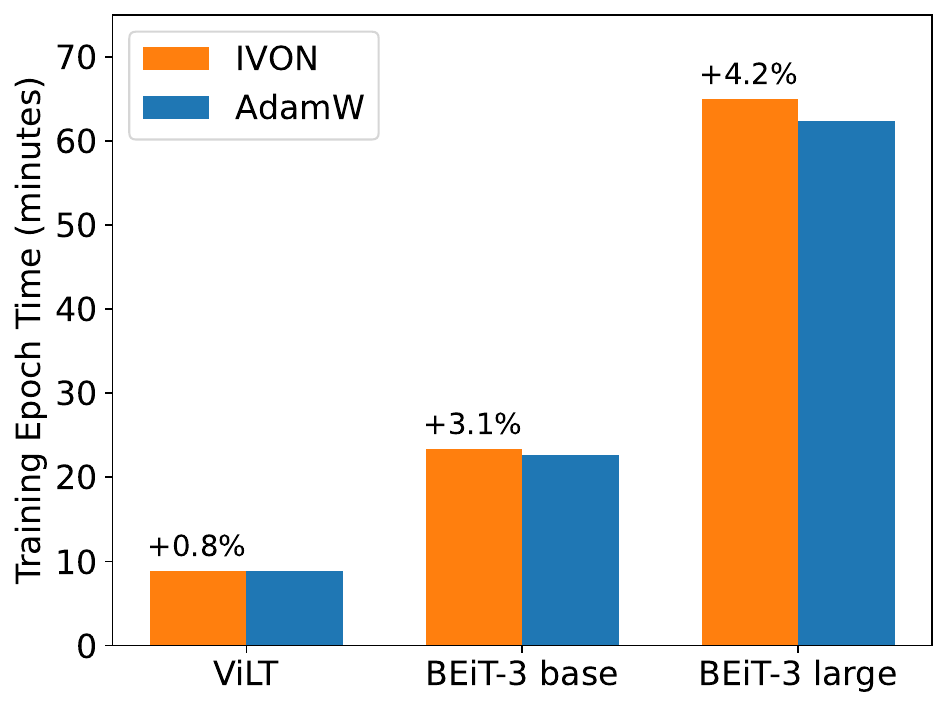}
    \caption{Training time per epoch.}
    \label{fig:wallclock_train}
  \end{subfigure}
  \hfill
  \begin{subfigure}[t]{0.49\textwidth}
    \centering
    \includegraphics[width=\linewidth]{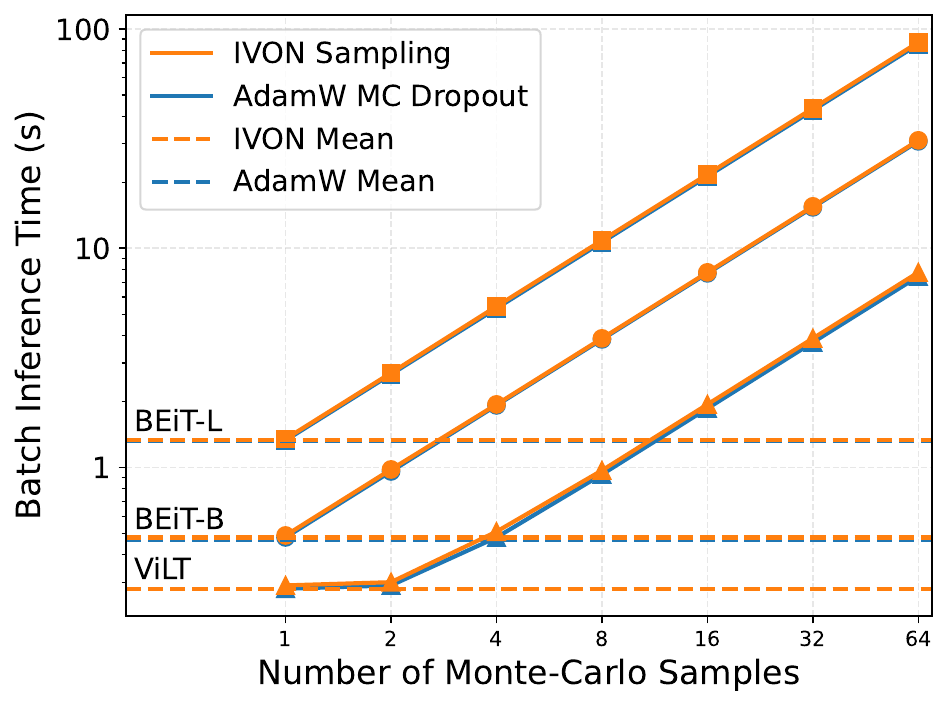}
    \caption{Inference time for different numbers of MC samples.}
    \label{fig:wallclock_eval}
  \end{subfigure}
  
  \caption{Computational overhead comparison (IVON vs AdamW). All runs were executed on NVIDIA A100 GPUs (80GB memory), and all numbers shown here are on VQAv2. (a-b) Peak GPU memory usage during training with varying batch sizes. (c-d) Wall-clock time for training (per epoch) and inference (per batch).}
  \label{fig:resource_overhead}
\end{figure}

We extensively compare the resource consumption of IVON and AdamW in terms of memory and training time in \Cref{fig:resource_overhead}. Peak GPU memory when training on VQAv2 is slightly higher for BEiT-3 base (\cref{fig:gpu_mem_base}) and moderately higher for BEiT-3 large (\cref{fig:gpu_mem_large}), indicating that more work might be needed to improve the efficiency of IVON on large models. The training overhead also increases for larger models, from just $0.8\%$ longer training with ViLT to $4.2\%$ with BEiT-3 large, partially due to the smaller maximum batch size. Finally, at inference, the IVON sampling is only negligbly slower than MC Droput (\cref{fig:wallclock_eval}). Here, the flat part for ViLT for one and two MC samples can be explained by the data loading taking longer than the batch processing through the model. We note again that the deterministic VarVQA mean uses the same inference time for a forward pass as an AdamW-trained model.

\begin{comment}
\begin{table}[!h]
  \caption{Training time comparison (per epoch) for full fine-tuning on VQAv2. All models were trained on a single node with 8 NVIDIA A100s. The epoch times are averaged across several machines.}
  \centering
  \renewcommand{\arraystretch}{1.25}
   \begin{tabular}{l|ccl}
    \toprule
    Model & $t_{\mathrm{epoch, AdamW}}$ & $t_{\mathrm{epoch, IVON}}$ & $\Delta t$ \\ \midrule
    ViLT & 8:40 min & 9:00 min & +4$\%$ \\ 
    BEiT-3 base & 26 min & 30 min & +15$\%$ \\ 
    BEiT-3 large & 1h 17 min & 1h 29 min & +15$\%$ \\ \bottomrule
  \end{tabular}
  \label{tab:supp_time}
\end{table}
\end{comment}

%% file: sec_tmlr/C_calibration_dropoutbpa.tex
\section{The Impact of Calibration}\label{sec:supp_calibration}

We apply common calibration methods \citep{guo2017calibration,platt1999probabilistic} on top of our trained models. For VQA, the models we investigate use sigmoids in the output layer\footnote{Softmax is not used, because VQAv2 is a multi-label task where the sum of all labels can be greater than 1. That, in turn, is due to the way that labels are inferred from the answers of 10 annotators, \cf \cref{sec:experiments_metrics}.}, therefore, temperature scaling cannot change relative confidence rankings (due to the strict monotonicity of the sigmoid). We thus use vector scaling and train a linear layer to learn the parameters, following \cite{reliable_vqa}. For NLVR2, the binary Softmax output is equivalent to a single sigmoid due to $p(x) = \frac{e^x}{e^x+e^y} = \frac{1}{1+e^{y-x}} = \sigma(x-y)$. As NLVR2 is balanced, temperature scaling and vector scaling are equivalent. We therefore use the former.

In \Cref{tab:metrics_calib_id_vqav2}, the results for VQAv2 are shown. Vector scaling consistently lowers ECE and provides slight benefits for the Selective Prediction metrics, while the accuracy remains approximately the same. The results for NLVR2 in \cref{tab:metrics_calib_id_nlvr2}, where temperature scaling instead of vector scaling is applied on top of the fine-tuned models, confirm these findings. Interestingly, while MC Dropout provides a lower ECE than VarVQA on NLVR2 (\cf \cref{tab:metrics_id_nlvr2}), the additional step of temperature scaling reverses the order, \ie VarVQA + temperature scaling achieves a lower ECE than AdamW Dropout + temperature scaling. As temperature scaling does not change the confidence ranking, the selective prediction metrics and accuracy remain unchanged.

\begin{comment}
\section{New selector function evaluated on MC Dropout}\label{sec:supp_gbpa_dropout}

We evaluate the impact of using our new risk-averse selector function \gsigma ~in combination with MC Dropout, repeating the experiments from \Cref{sec:experiments_selfunc}. For VarVQA, the new selector \gsigma ~clearly outperforms \gmpmu ~in high-stakes metrics (better scores in 19/20 cases, \cf \cref{tab:conffunc_comparison_vqav2,tab:conffunc_comparison_nlvr2}), while the performance is roughly equal in low-stakes metrics (\gsigma ~is better 4 times, worse 5 times, and once the score is tied). For MC Dropout however, the picture is different. In \Cref{tab:conffunc_comparison_mcdrop_vqav2,tab:conffunc_comparison_mcdrop_nlvr2}, we show the results when using $N=64$ samples (as always) for VQAv2 and NLVR2, respectively. On the high-stakes metrics, the risk-averse selector \gsigma ~yields a similar performance to \gmpmu ~(better 11 times, worse 9 times), whereas on the low-stakes metrics, \gsigma ~is clearly worse (loses 8 times, wins only 2 times). In conclusion, we find that the output variances produced by MC Dropout are not helpful for selective prediction, possibly because they originate from an ad-hoc posterior. In contrast, when the posterior distribution over parameters is learned, \eg with IVON, the output variances benefit and carry meaningful information that can improve abstention decisions.
\end{comment}

\begin{table}[!tbhp]
    \small
    \renewcommand{\arraystretch}{1.2}
    \centering
    \caption{Reliability evaluation on VQAv2 for fine-tuned models with an additional step of vector scaling. See \cref{tab:metrics_id_vqav2} for the comparison of the uncalibrated models. The variable $N$ denotes the number of forward passes. Best results per model are \textbf{bold}.}
    \begin{tabular}{ll|c|c|c|cccc|cc} \toprule
        \multirow{3}{*}{Model} & \multirow{3}{*}{Method} & \multirow{3}{*}{$N$} & \multirow{3}{*}{\gray{Acc.}} & \multirow{2}{*}{Calibration} & \multicolumn{4}{c|}{Selective Prediction} & \multicolumn{2}{c}{\gray{Sel. Prediction}} \\
        & & & & & \multicolumn{4}{c|}{\textit{high-stakes}} & \multicolumn{2}{c}{\textit{\gray{low-stakes}}} \\
        & & & & ECE~($\downarrow$) & $C@\frac{1}{2}\%$ & $C@1\%$ & $\Phi_{50}$ & $\Phi_{100}$ & \gray{$C@5\%$} & \gray{$\Phi_{10}$}\\ \midrule
        %\multirow{4}{*}{ViLT} & AdamW & 1 & \gray{69.30} & 0.061 & 5.03 & 10.58 & 8.41 & 2.89 & \gray{36.24} & \gray{24.05}\\
        \multirow{4}{*}{\centering \;\,ViLT} & AdamW & 1 & \gray{69.29} & 0.024 & 7.02 & 13.32 & 9.80 & 3.22 & \gray{36.57} & \gray{24.13} \\ 
        %& VarVQA mean & 1 & \gray{69.63} & 0.071 & 6.77 & 13.32 & 9.74 & 5.45 & \gray{37.93} & \gray{25.08} \\
        & VarVQA mean & 1 & \gray{69.62} & 0.030 & 8.85 & 15.17 & 9.82 & 7.02 & \gray{38.19} & \gray{25.13} \\
        \cdashline{2-11}
        %& AdamW Dropout & 64 & \gray{69.66} & 0.019 & 10.44 & 16.63 & 12.51 & 8.44 & \gray{38.49} & \gray{26.18} \\
        & AdamW Dropout & 64 & \gray{69.66} & \textbf{0.007} & 12.06 & 17.34 & 12.76 & 9.75 & \gray{38.45} & \gray{26.42} \\
        %& VarVQA & 64 & \gray{69.71} & 0.019 & 13.81 & 19.68 & 12.93 & 10.88 & \gray{39.53} & \gray{27.15} \\ 
        & VarVQA & 64 & \gray{69.70} & 0.009 & \textbf{14.41} & \textbf{19.88} & \textbf{13.74} & \textbf{11.21} & \gray{\textbf{39.63}} & \gray{\textbf{27.26}} \\ \midrule
        %\multirow{4}{1.1cm}{BEiT-3 base} & AdamW & 1 & \gray{73.60} & 0.041 & 10.35 & 18.55 & 15.59 & 8.65 & \gray{47.93} & \gray{33.40} \\
        \multirow{4}{1.1cm}{\centering BEiT-3 base} & AdamW & 1 & \gray{73.67} & 0.017 & 13.16 & 20.84  & 16.01 & 9.36 & \gray{48.16} & \gray{33.67} \\
        %& VarVQA mean & 1 & \gray{73.84} & 0.039 & 14.08 & 21.98 & 16.72 & 11.36 & \gray{49.57} & \gray{34.80} \\
        & VarVQA mean & 1 & \gray{73.84} & 0.014 & 15.65 & 22.96 & 16.64 & 11.09 & \gray{49.76} & \gray{34.57} \\ \cdashline{2-11}
        %& AdamW Dropout & 64 & \gray{73.46} & 0.019 & 13.07 & 20.11  & 16.61 & 9.44 & \gray{47.49} & \gray{33.36} \\
        & AdamW Dropout & 64 & \gray{73.56} & 0.010 & 13.70 & 21.35 & 16.09 & 11.07 & \gray{47.82} & \gray{33.64} \\
        %& VarVQA & 64 & \gray{73.79} & 0.018 & 18.10 & 24.66 & \textbf{19.26} & 13.90 & \gray{49.76} & \gray{\textbf{35.22}} \\ 
        & VarVQA & 64 & \gray{73.79} & \textbf{0.008} & \textbf{18.66} & \textbf{25.25} & \textbf{19.26} & \textbf{14.41} & \textbf{\gray{50.09}} & \textbf{\gray{35.22}} \\ \midrule
        %\multirow{4}{1.1cm}{BEiT-3 large} & AdamW & 1 & \gray{78.59} & 0.039 & 21.63 & 32.15 & 26.31 & 17.80 & \gray{63.19} & \gray{45.83} \\
        \multirow{4}{1.1cm}{\centering BEiT-3 large} & AdamW & 1 & \gray{78.60} & 0.016 & 25.10 & 34.52 & 27.74 & 18.36 & \gray{63.34} & \gray{46.41} \\ 
        %& VarVQA mean & 1 & \gray{78.96} & 0.035 & 25.32 & 35.35 & 28.31 & 21.25 & \gray{64.83} & \gray{47.43} \\
        & VarVQA mean & 1 & \gray{78.93} & 0.013 & 27.29 & 36.23 & 28.80 & 22.27 & \gray{64.72} & \gray{47.68} \\
        \cdashline{2-11}
        %& AdamW Dropout & 64 & \gray{78.41} & 0.018 & 25.28 & 34.52  & 27.99 & 20.65 & \gray{63.00} & \gray{46.23} \\
        & AdamW Dropout & 64 & \gray{78.49} & 0.008 & 27.77 & 36.07 & 26.48 & 21.86 & \gray{63.25} & \gray{46.65} \\
        %& VarVQA & 64 & \gray{78.89} & 0.018 & 28.13 & 37.05 & 29.56 & \textbf{23.21} & \gray{64.68} & \gray{48.06} \\ 
        & VarVQA & 64 & \gray{78.86} & \textbf{0.007} & \textbf{29.61} & \textbf{37.79} & \textbf{30.80} & \textbf{22.87} & \textbf{\gray{64.84}} & \textbf{\gray{48.29}} \\ \bottomrule
    \end{tabular}
    \label{tab:metrics_calib_id_vqav2}
\end{table}

\begin{table}[!tbhp]
    \small
    \renewcommand{\arraystretch}{1.2}
    \centering
    \caption{Reliability evaluation on NLVR2 for fine-tuned models with an additional step of temperature scaling. See \cref{tab:metrics_id_nlvr2} for the comparison of the uncalibrated models. The variable $N$ denotes the number of forward passes. Best results per model are \textbf{bold}.}
    \begin{tabular}{ll|c|c|c|cccc|cc} \toprule
        \multirow{3}{*}{Model} & \multirow{3}{*}{Method} & \multirow{3}{*}{$N$} & \multirow{3}{*}{\gray{Acc.}} & \multirow{2}{*}{Calibration} & \multicolumn{4}{c|}{Selective Prediction} & \multicolumn{2}{c}{\gray{Sel. Prediction}} \\
        & & & & & \multicolumn{4}{c|}{\textit{high-stakes}} & \multicolumn{2}{c}{\textit{\gray{low-stakes}}} \\
        & & & & ECE~($\downarrow$) & $C@\frac{1}{2}\%$ & $C@1\%$ & $\Phi_{50}$ & $\Phi_{100}$
        & \gray{$C@5\%$} & \gray{$\Phi_{10}$}\\ \midrule
        %\multirow{4}{1.1cm}{BEiT-3 base} & AdamW & 1 & \gray{83.45} & 0.059 & 6.42 & 11.61 & 4.58 & 2.24  & \gray{54.79} & \gray{26.18} \\
        \multirow{4}{1.1cm}{\centering BEiT-3 base} & AdamW & 1 & \gray{83.45} & 0.011 & 6.42 & 11.61 & 4.58 & 2.24  & \gray{54.79} & \gray{26.18} \\
        %& VarVQA mean & 1 & \gray{83.28} & 0.058 & 5.15 & 15.58 & 6.44 & 1.41 & \gray{55.66} & \gray{27.30} \\
        & VarVQA mean & 1 & \gray{83.28} & 0.012 & 5.15 & 15.58 & 6.44 & 1.41 & \gray{55.66} & \gray{27.63} \\
        \cdashline{2-11}
        %& AdamW Dropout & 64 & \gray{83.18} & 0.016 & 9.98 & 15.99 & 6.95 & 2.95 & \gray{55.43} & \gray{27.63} \\
        & AdamW Dropout & 64 & \gray{83.18} & 0.011 & 9.98 & 15.99 & 6.95 & 2.95 & \gray{55.43} & \gray{27.63} \\
        %& VarVQA & 64 & \gray{83.11} & 0.031 & \textbf{15.42} & \textbf{23.36} & \textbf{11.20} & \textbf{5.00} & \gray{\textbf{57.16}} & \gray{\textbf{29.23}} \\
        & VarVQA & 64 & \gray{83.11} & \textbf{0.009} & \textbf{15.42} & \textbf{23.36} & \textbf{11.20} & \textbf{5.00} & \gray{\textbf{57.16}} & \gray{\textbf{29.23}} \\ \midrule
        %\multirow{4}{1.1cm}{BEiT-3 large} & AdamW & 1 & \gray{88.34} & 0.041 & 16.53 & 41.14 & 18.08 & 9.45 & \gray{78.53} & \gray{45.64} \\
        \multirow{4}{1.1cm}{\centering BEiT-3 large} & AdamW & 1 & \gray{88.34} & 0.012 & 16.53 & 41.14 & 18.08 & 9.45 & \gray{78.53} & \gray{45.64} \\ 
        %& VarVQA mean & 1 & \gray{88.83} & 0.062 & 17.15 & 31.07 & 15.27 & 3.57 & \gray{80.17} & \gray{45.02} \\
        & VarVQA mean & 1 & \gray{88.83} & 0.010 & 17.15 & 31.07 & 15.27 & 3.57 & \gray{80.17} & \gray{45.02} \\
        \cdashline{2-11}
        %& AdamW Dropout & 64 & \gray{88.11} & 0.017 & \textbf{33.21} & 44.69 & 23.43 & 14.71 & \gray{76.99} & \gray{46.55} \\
        & AdamW Dropout & 64 & \gray{88.11} & 0.009 & \textbf{33.21} & 44.69 & 23.43 & 14.71 & \gray{76.99} & \gray{46.55} \\
        %& VarVQA & 64 & \gray{89.26} & 0.029 & 32.89 & \textbf{49.24} & \textbf{25.56} & \textbf{14.85}  & \gray{\textbf{82.11}} & \gray{\textbf{49.51}} \\ 
        & VarVQA & 64 & \gray{89.26} & \textbf{0.006} & 32.89 & \textbf{49.24} & \textbf{25.56} & \textbf{14.85} & \gray{\textbf{82.11}} & \gray{\textbf{49.51}} \\ \bottomrule
    \end{tabular}
    \label{tab:metrics_calib_id_nlvr2}
\end{table}

%% file: sec_tmlr/D_full_results.tex
\section{Extended Results}\label{sec:supp_full_results}
We extend the results for different numbers of MC samples and the comparison to MC dropout (\cf \cref{fig:mcdropout}) for both VQAv2 (\Cref{fig:mc_ablation_b3l_vqa,fig:mc_ablation_b3b_vqa,fig:mc_ablation_vilt_vqa}) and NLVR2 (\Cref{fig:mc_ablation_b3l_nlvr,fig:mc_ablation_b3b_nlvr}). Furthermore, we extend the results for different ID/OOD fractions (\cf \cref{fig:ood}) in \Cref{fig:id_ood_full_b3l,fig:id_ood_full_b3b,fig:id_ood_full_vilt}. The findings of the main paper for BEiT-3 large hold true across BEiT-3 base and ViLT, namely:

\begin{itemize}
    \item \approach is as effective as AdamW for training large multimodal models - it matches or sometimes even surpasses the accuracy obtained with AdamW.
    \item \approach reduces miscalibration in terms of the Expected Calibration Error (ECE).
    \item \approach improves selective prediction through more appropriate abstentions. The largest improvements are obtained for the high-stakes metrics.
    \item \approach gives consistently better results in terms of selective prediction and at least equally good results in terms of calibration compared to MC dropout, which has the same inference overhead.
    \item The benefits of \approach translate to the mixed ID/OOD setting, where the benefits are again more apparent for the high-stakes metrics $C@1\%$ and $\Phi_{100}$.
\end{itemize}

Additionally, we analyze the fraction of answered and abstained questions for the different question categories of VQAv2 and AdVQA (\emph{Binary}, \emph{Number}, and \emph{Other}) in \Cref{tab:qual_categories_vqa}. Particularly the `Number' and `Other' categories are challenging to the models,  with Coverages rapidly dropping to single digits when only a small fraction of OOD samples are added. Overall, VarVQA performs best in all categories.

%% file: sec_tmlr/E_qualitative.tex
\section{Qualitative Examples}\label{sec:supp_qualitative}
We present further qualitative results, on VQAv2, AdVQA and NLVR2. In particular, we show cases in which VarVQA is correct while AdamW abstains, for both VQAv2 (\cref{fig:qual_adam_abstain_ivon_correct_vqa}) and NLVR2 (\cref{fig:qual_adam_abstain_ivon_correct_nlvr}). Additionally, we show further cases in which VarVQA abstains while AdamW is wrong, for AdVQA (OOD, \cref{fig:qual_adam_wrong_ivon_abstain_advqa}). Finally, we also show failure cases of our method, \ie AdamW abstains while VarVQA is wrong, both for VQAv2 (\cref{fig:qual_adam_abstain_ivon_wrong_vqa}) and NLVR2 (\cref{fig:qual_adam_abstain_ivon_wrong_nlvr}). For all examples, we set the abstention threshold $\gamma$ by optimizing $\Phi_{100}$ on ID validation data\footnote{For NLVR2, we use $\Phi_{50}$, as $\Phi_{100}$ is very noisy due to the small size of the dataset.}. We always pick examples where the answers of AdamW and VarVQA are identical and where the gap in their confidence is largest. Interestingly, a large number of examples where VarVQA performs better on NLVR2 seem to be related to counting, we leave it to future work to explore this further.

%% file: sec_tmlr/F_largefigures.tex
\begin{figure}[thbp]
  \centering
  \begin{subfigure}{0.24\textwidth}
    \includegraphics[width=\textwidth]{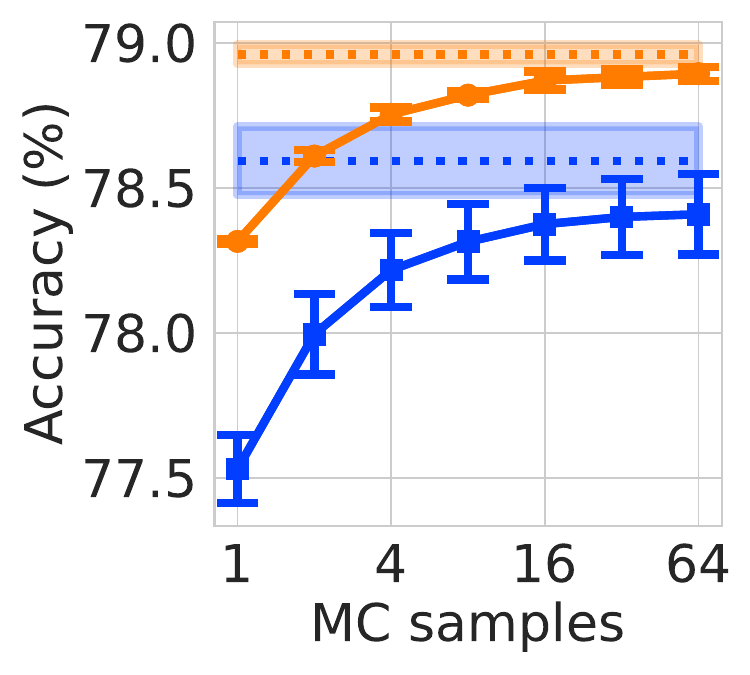}
    \caption{\textbf{Accuracy}}
  \end{subfigure}
  \begin{subfigure}{0.24\textwidth}
    \includegraphics[width=\textwidth]{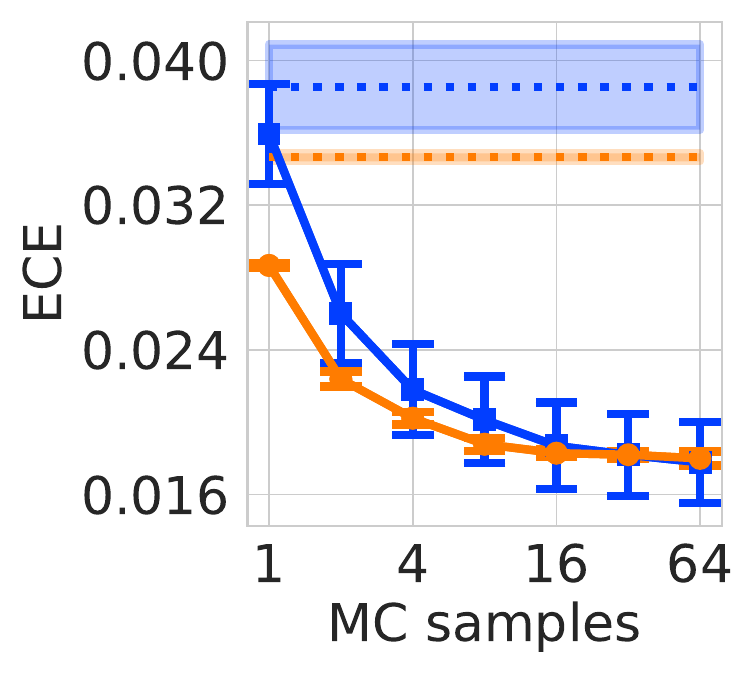}
    \caption{\textbf{Calibration} ($\downarrow$)}
  \end{subfigure}
  \begin{subfigure}{0.24\textwidth}
    \includegraphics[width=\textwidth]{figures/b3l_vqa/cov_0.01_ablation_b3l.pdf}
    \caption{\textbf{Sel. Prediction}}
  \end{subfigure}
  \begin{subfigure}{0.24\textwidth}
    \includegraphics[width=\textwidth]{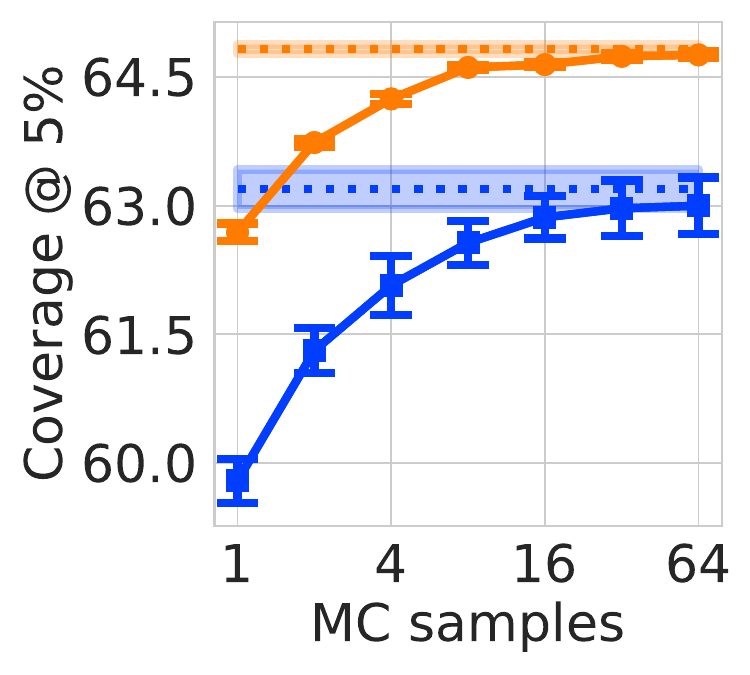}
    \caption{\textbf{Sel. Prediction}}
  \end{subfigure}
  \begin{subfigure}{0.24\textwidth}
    \includegraphics[width=\textwidth]{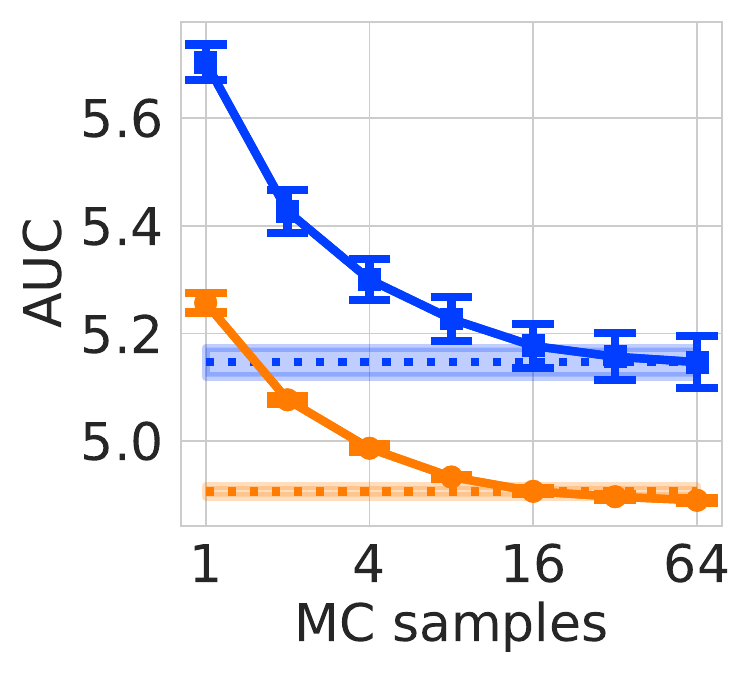}
  \caption{\textbf{Sel. Prediction} ($\downarrow$)}
  \end{subfigure}
  \begin{subfigure}{0.24\textwidth}
    \includegraphics[width=\textwidth]{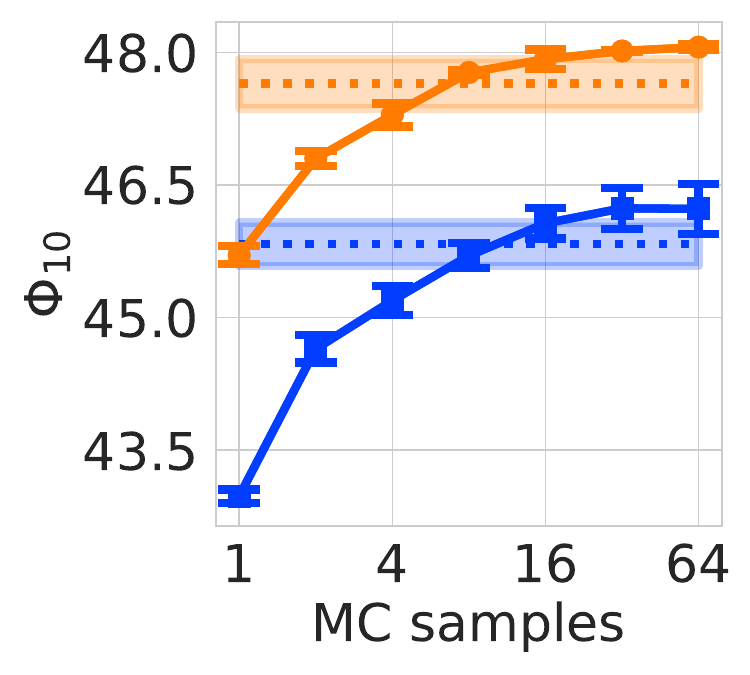}
    \caption{\textbf{Sel. Prediction}}
  \end{subfigure}
  \begin{subfigure}{0.24\textwidth}
    \includegraphics[width=\textwidth]{figures/b3l_vqa/phi100_ablation_b3l.pdf}
    \caption{\textbf{Sel. Prediction}}
  \end{subfigure}
  \begin{subfigure}{0.24\textwidth}
    \includegraphics[width=\textwidth]{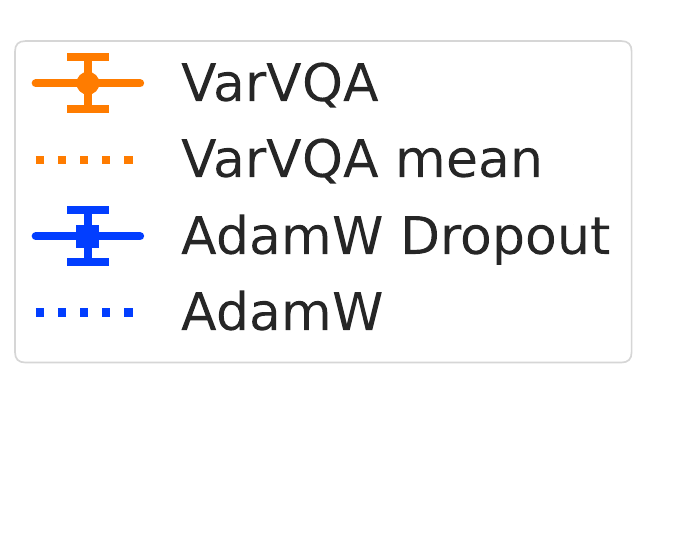}
  \end{subfigure}
  \caption{Sample ablation and comparison to MC dropout for BEiT-3 large on VQAv2. Lower is better for ECE and AUC. Standard error across three training runs with different seeds is shown for all methods.}
  \label{fig:mc_ablation_b3l_vqa}
\end{figure}

\begin{figure}[thbp]
  \centering
  \begin{subfigure}{0.24\textwidth}
    \includegraphics[width=\textwidth]{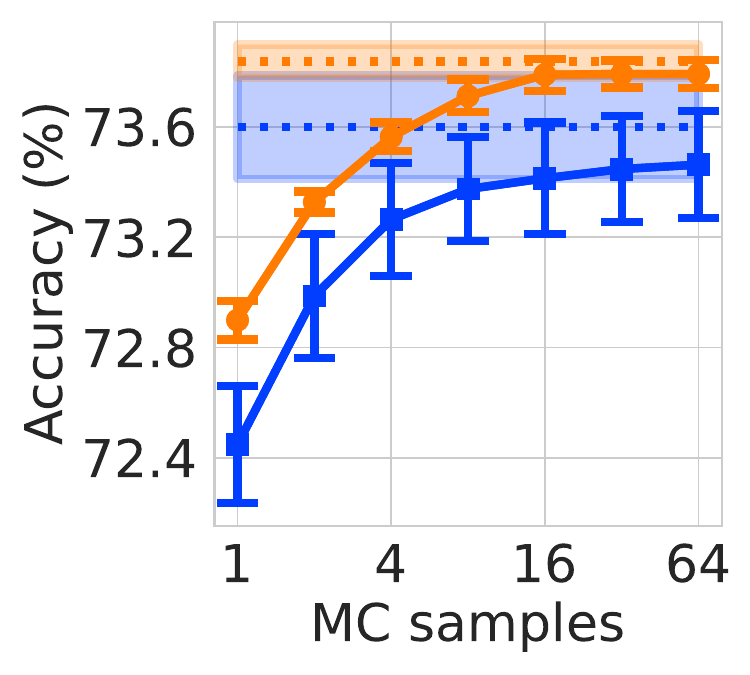}
    \caption{\textbf{Accuracy}}
  \end{subfigure}
  \begin{subfigure}{0.24\textwidth}
    \includegraphics[width=\textwidth]{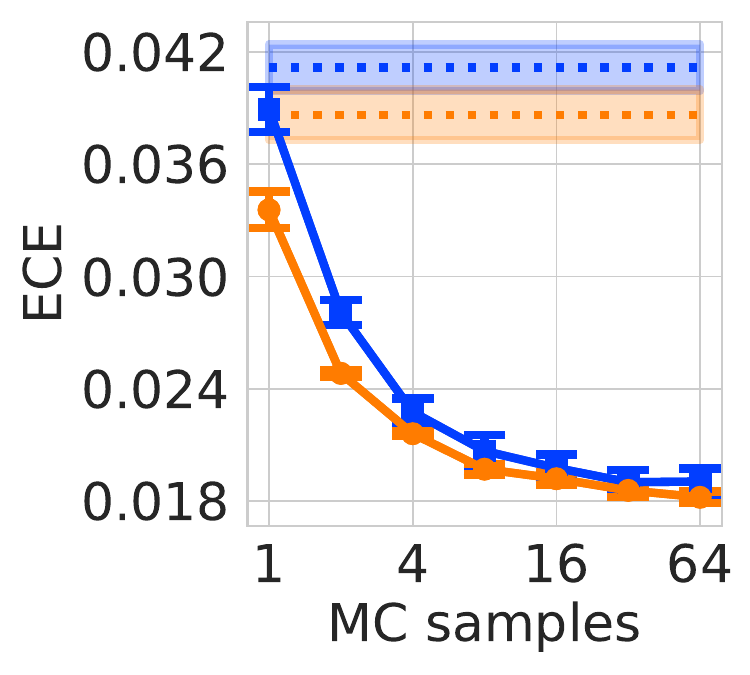}
    \caption{\textbf{Calibration} ($\downarrow$)}
  \end{subfigure}
  \begin{subfigure}{0.24\textwidth}
    \includegraphics[width=\textwidth]{figures/b3b_vqa/cov_0.01_ablation_b3b.pdf}
    \caption{\textbf{Sel. Prediction}}
  \end{subfigure}
  \begin{subfigure}{0.24\textwidth}
    \includegraphics[width=\textwidth]{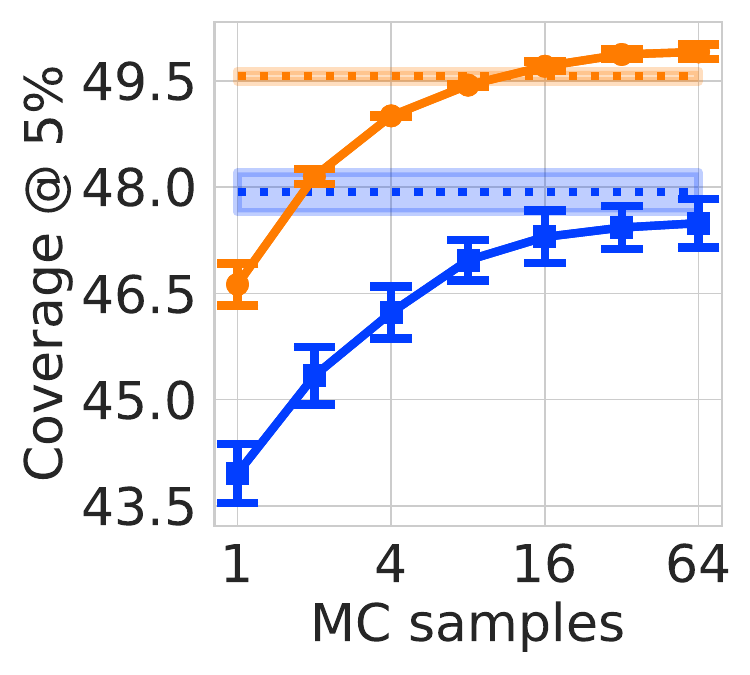}
    \caption{\textbf{Sel. Prediction}}
  \end{subfigure}
  \begin{subfigure}{0.24\textwidth}
    \includegraphics[width=\textwidth]{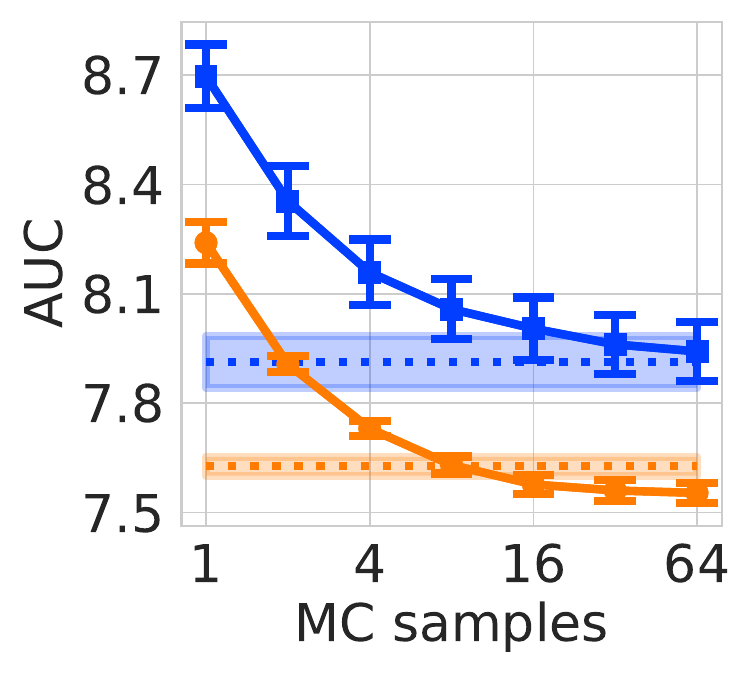}
    \caption{\textbf{Sel. Prediction} ($\downarrow$)}
  \end{subfigure}
  \begin{subfigure}{0.24\textwidth}
    \includegraphics[width=\textwidth]{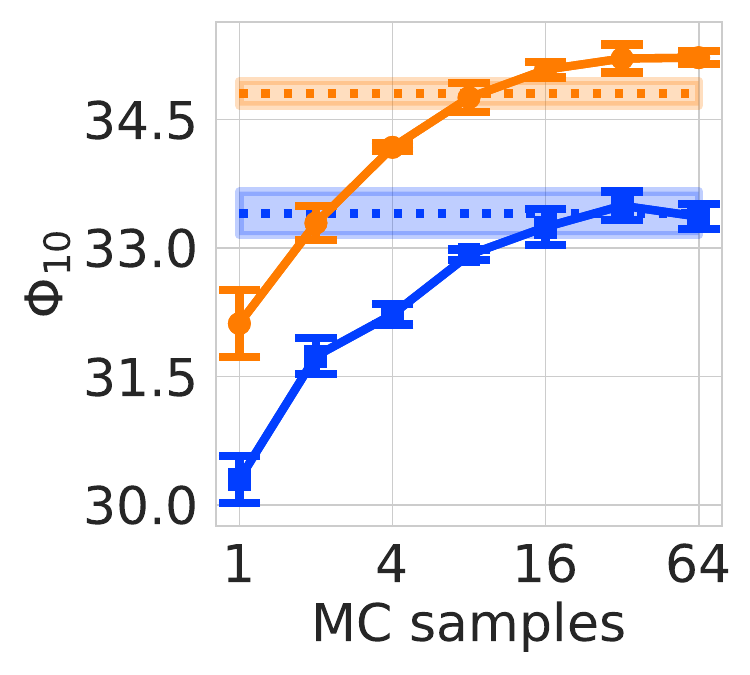}
    \caption{\textbf{Sel. Prediction}}
  \end{subfigure}
  \begin{subfigure}{0.24\textwidth}
    \includegraphics[width=\textwidth]{figures/b3b_vqa/phi100_ablation_b3b.pdf}
    \caption{\textbf{Sel. Prediction}}
  \end{subfigure}
  \begin{subfigure}{0.24\textwidth}
    \includegraphics[width=\textwidth]{figures/legends/4_vertical.pdf}
  \end{subfigure}
  \caption{Sample ablation and comparison to MC dropout for BEiT-3 base on VQAv2. Lower is better for ECE and AUC. Standard error across three training runs with different seeds is shown for all methods.}
  \label{fig:mc_ablation_b3b_vqa}
\end{figure}

\begin{figure}[thbp]
  \centering
  \begin{subfigure}{0.24\textwidth}
    \includegraphics[width=\textwidth]{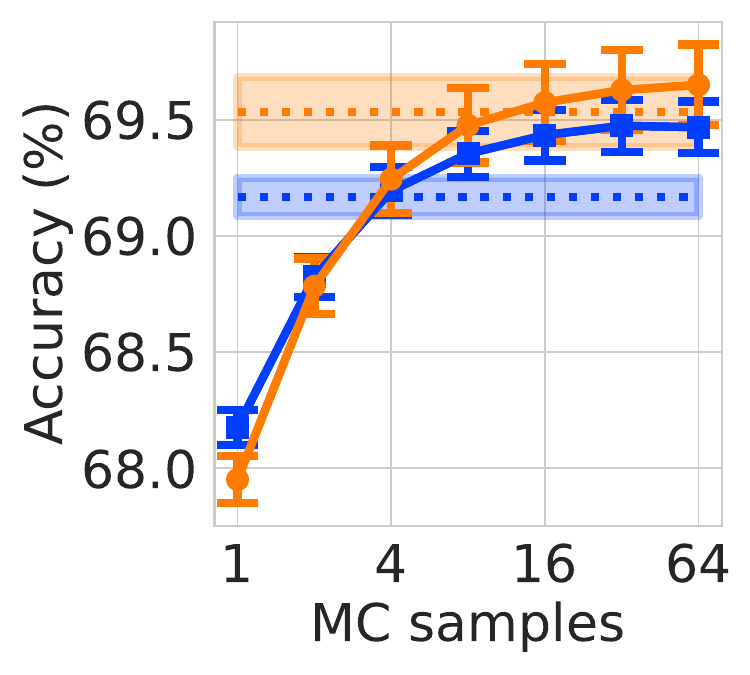}
    \caption{\textbf{Accuracy}}
  \end{subfigure}
  \begin{subfigure}{0.24\textwidth}
    \includegraphics[width=\textwidth]{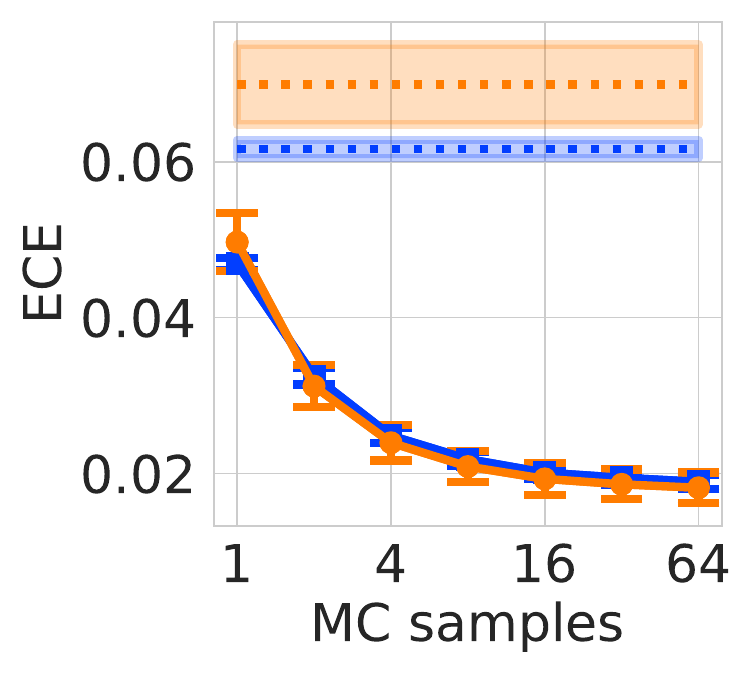}
    \caption{\textbf{Calibration} ($\downarrow$)}
  \end{subfigure}
  \begin{subfigure}{0.24\textwidth}
    \includegraphics[width=\textwidth]{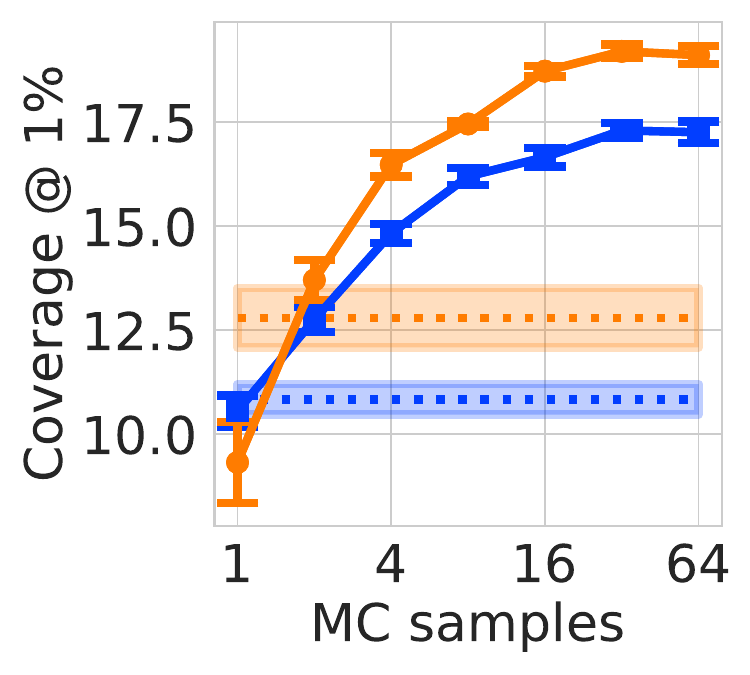}
    \caption{\textbf{Sel. Prediction}}
  \end{subfigure}
  \begin{subfigure}{0.24\textwidth}
    \includegraphics[width=\textwidth]{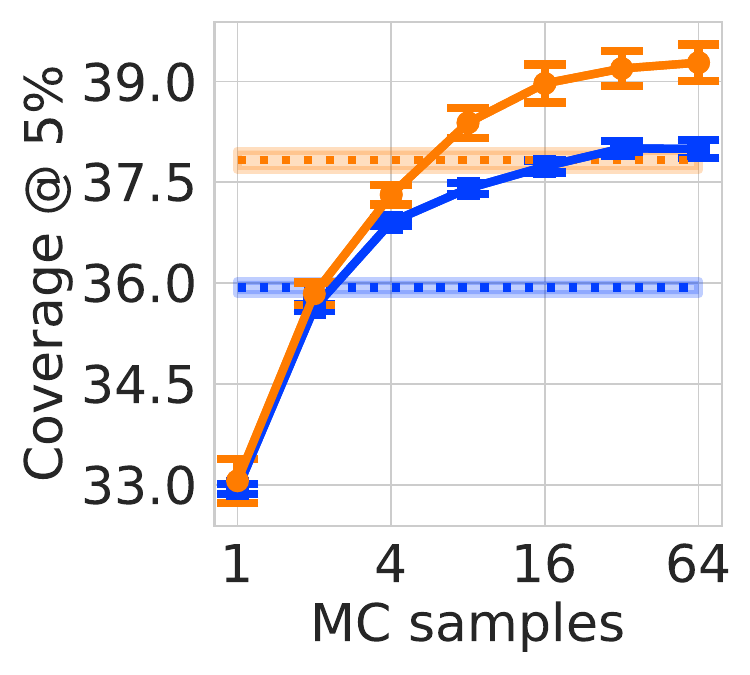}
    \caption{\textbf{Sel. Prediction}}
  \end{subfigure}
  \begin{subfigure}{0.24\textwidth}
    \includegraphics[width=\textwidth]{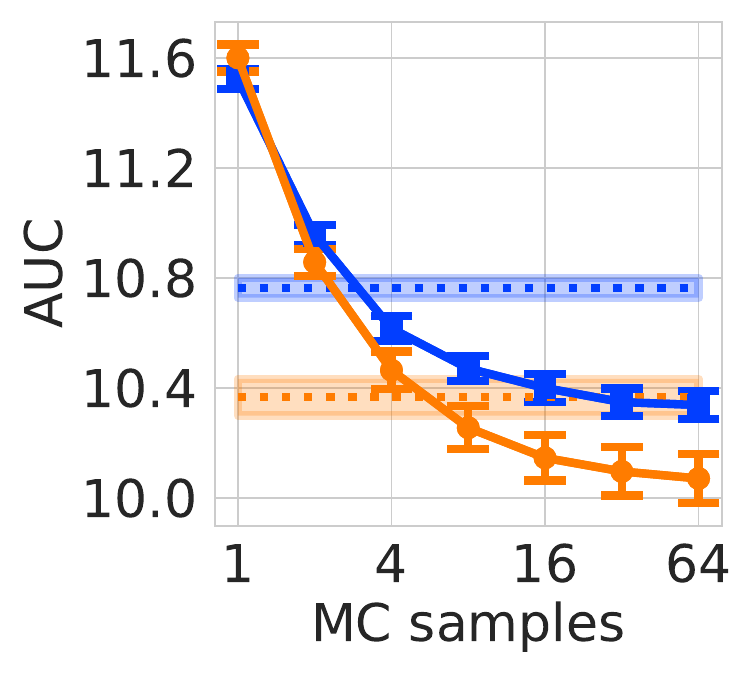}
    \caption{\textbf{Sel. Prediction} ($\downarrow$)}
  \end{subfigure}
  \begin{subfigure}{0.24\textwidth}
    \includegraphics[width=\textwidth]{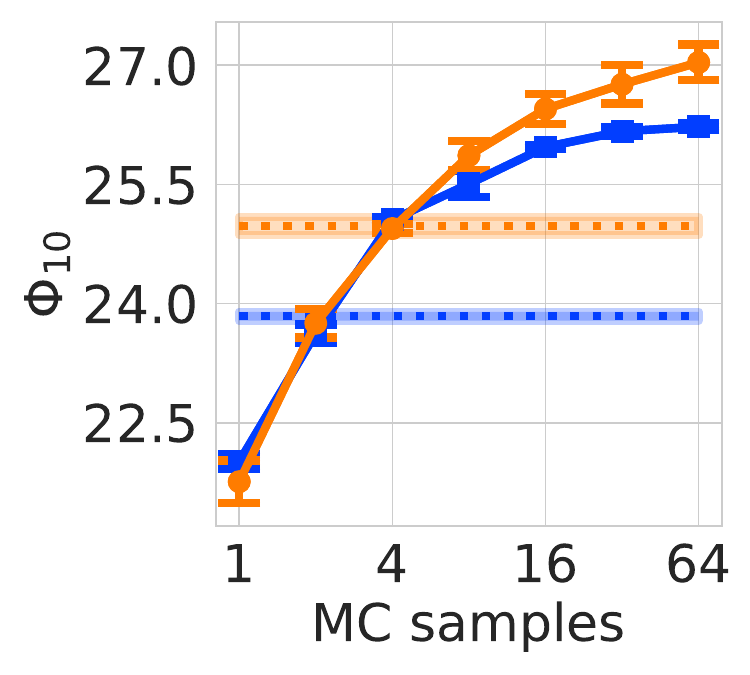}
    \caption{\textbf{Sel. Prediction}}
  \end{subfigure}
  \begin{subfigure}{0.24\textwidth}
    \includegraphics[width=\textwidth]{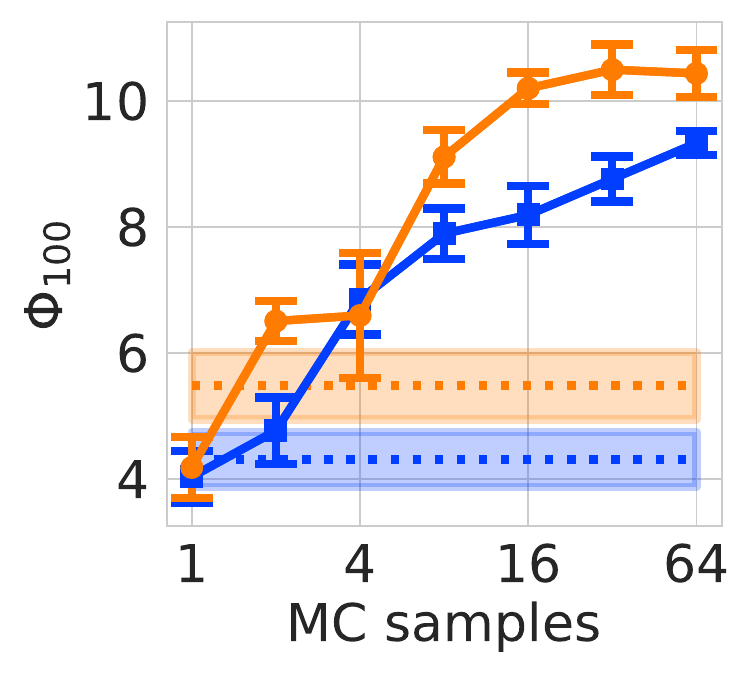}
    \caption{\textbf{Sel. Prediction}}
  \end{subfigure}
  \begin{subfigure}{0.24\textwidth}
    \includegraphics[width=\textwidth]{figures/legends/4_vertical.pdf}
  \end{subfigure}
  \caption{Sample ablation and comparison to MC dropout for ViLT on VQAv2. Lower is better for ECE and AUC. Standard error across three training runs with different seeds is shown for all methods.}
  \label{fig:mc_ablation_vilt_vqa}
\end{figure}

\begin{figure}[thbp]
  \centering
  \begin{subfigure}{0.24\textwidth}
    \includegraphics[width=\textwidth]{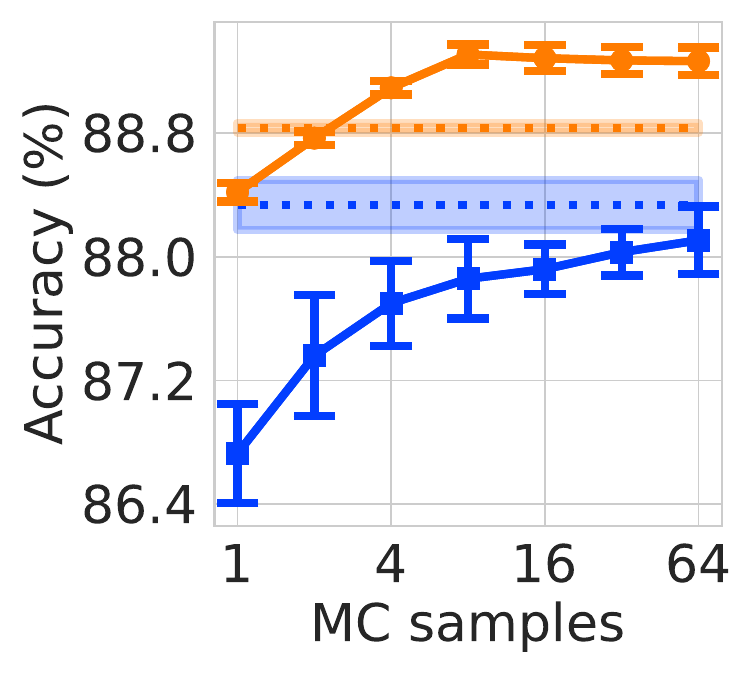}
    \caption{\textbf{Accuracy}}
  \end{subfigure}
  \begin{subfigure}{0.24\textwidth}
    \includegraphics[width=\textwidth]{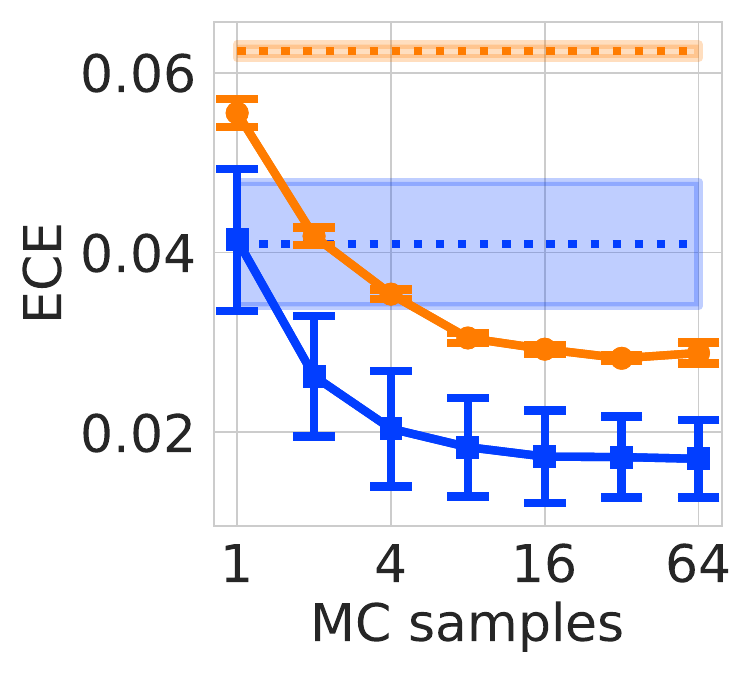}
    \caption{\textbf{Calibration} ($\downarrow$)}
  \end{subfigure}
  \begin{subfigure}{0.24\textwidth}
    \includegraphics[width=\textwidth]{figures/b3l_nlvr2/cov_0.01_ablation_b3l.pdf}
    \caption{\textbf{Sel. Prediction}}
  \end{subfigure}
  \begin{subfigure}{0.24\textwidth}
    \includegraphics[width=\textwidth]{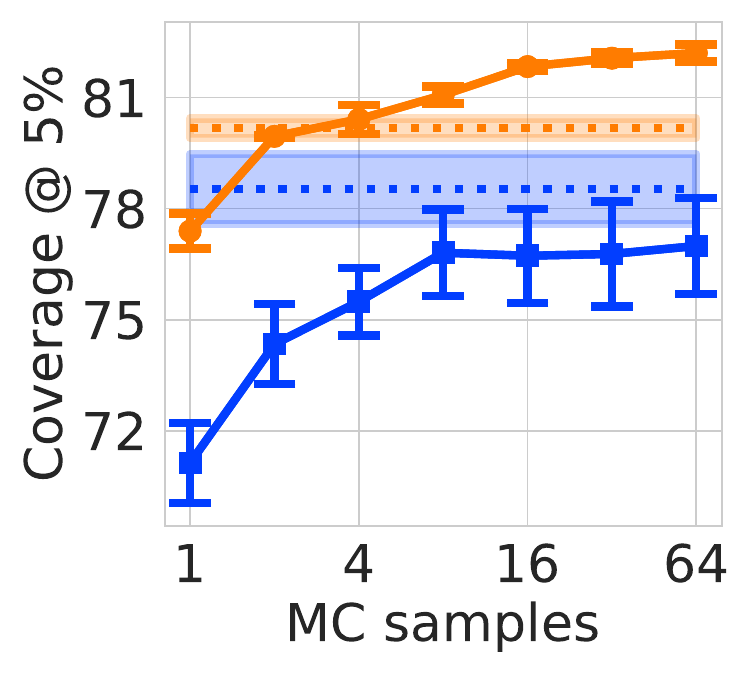}
    \caption{\textbf{Sel. Prediction}}
  \end{subfigure}
  \begin{subfigure}{0.24\textwidth}
    \includegraphics[width=\textwidth]{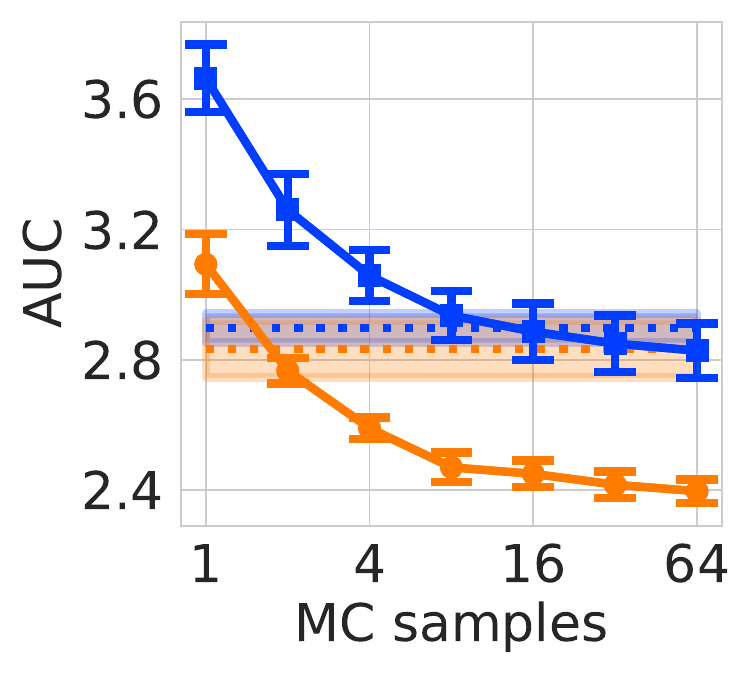}
    \caption{\textbf{Sel. Prediction} ($\downarrow$)}
  \end{subfigure}
  \begin{subfigure}{0.24\textwidth}
    \includegraphics[width=\textwidth]{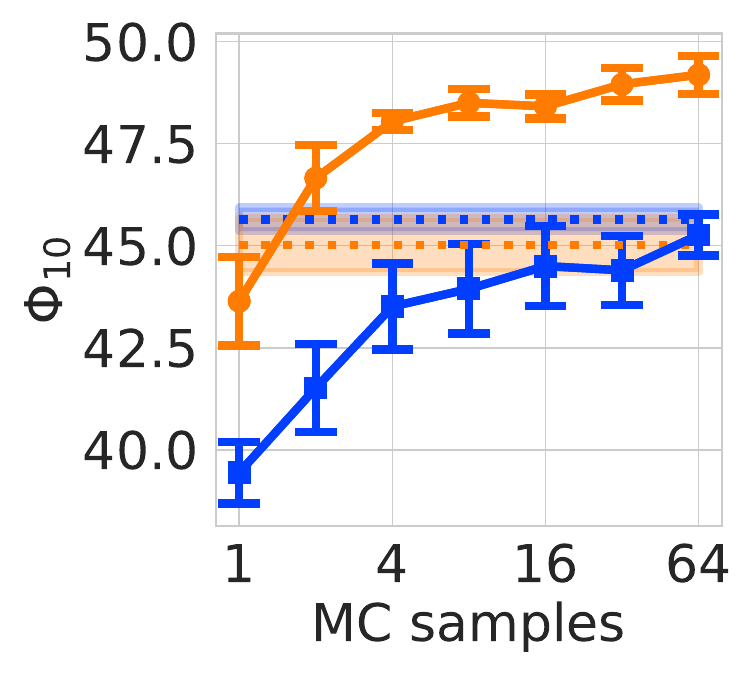}
    \caption{\textbf{Sel. Prediction}}
  \end{subfigure}
  \begin{subfigure}{0.24\textwidth}
    \includegraphics[width=\textwidth]{figures/b3l_nlvr2/phi50_ablation_b3l.pdf}
    \caption{\textbf{Sel. Prediction}}
  \end{subfigure}
  \begin{subfigure}{0.24\textwidth}
    \includegraphics[width=\textwidth]{figures/legends/4_vertical.pdf}
  \end{subfigure}
  \caption{Sample ablation and comparison to MC Dropout for BEiT-3 large on NLVR2. Lower is better for ECE and AUC. Standard error across three training runs with different seeds is shown for all methods.}
  \label{fig:mc_ablation_b3l_nlvr}
\end{figure}

\begin{figure}[thbp]
  \centering
  \begin{subfigure}{0.24\textwidth}
    \includegraphics[width=\textwidth]{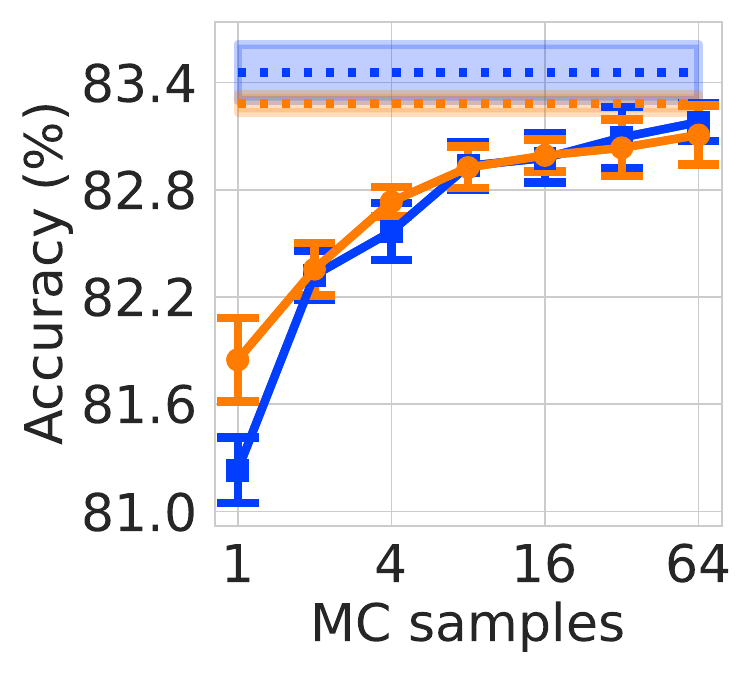}
    \caption{\textbf{Accuracy}}
  \end{subfigure}
  \begin{subfigure}{0.24\textwidth}
    \includegraphics[width=\textwidth]{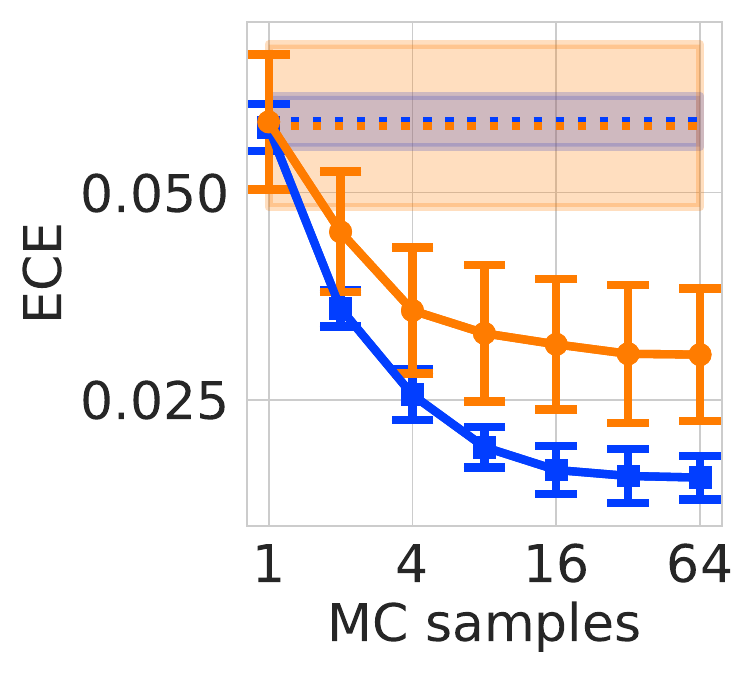}
    \caption{\textbf{Calibration} ($\downarrow$)}
  \end{subfigure}
  \begin{subfigure}{0.24\textwidth}
    \includegraphics[width=\textwidth]{figures/b3b_nlvr2/cov_0.01_ablation_b3b.pdf}
    \caption{\textbf{Sel. Prediction}}
  \end{subfigure}
  \begin{subfigure}{0.24\textwidth}
    \includegraphics[width=\textwidth]{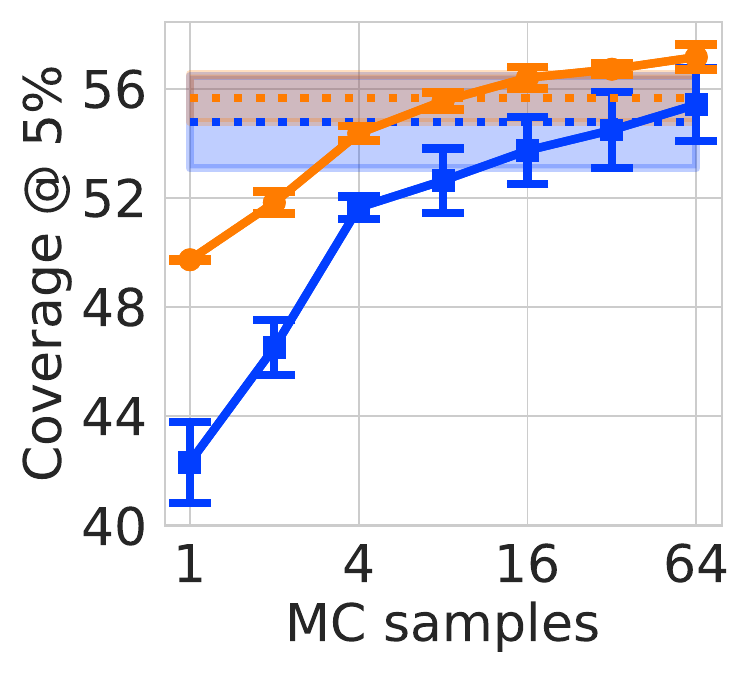}
    \caption{\textbf{Sel. Prediction}}
  \end{subfigure}
  \begin{subfigure}{0.24\textwidth}
    \includegraphics[width=\textwidth]{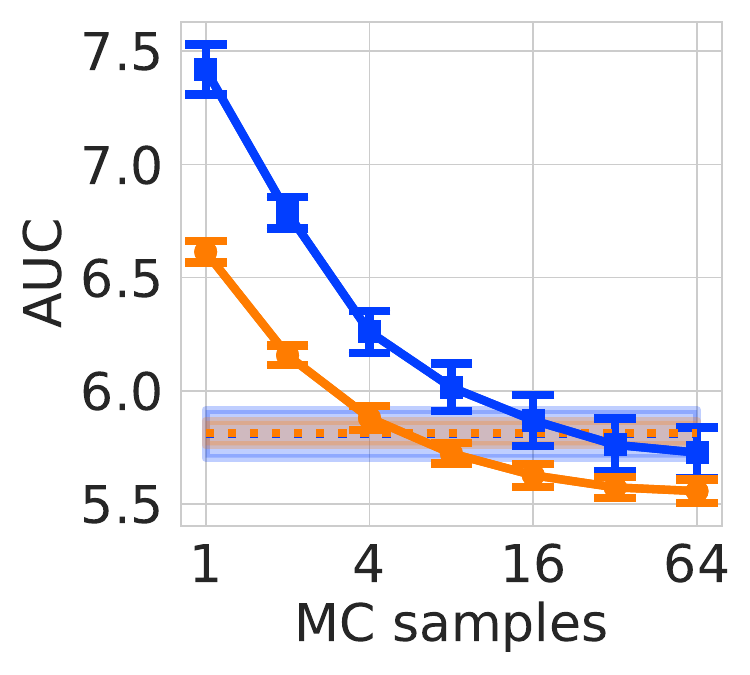}
    \caption{\textbf{Sel. Prediction} ($\downarrow$)}
  \end{subfigure}
  \begin{subfigure}{0.24\textwidth}
    \includegraphics[width=\textwidth]{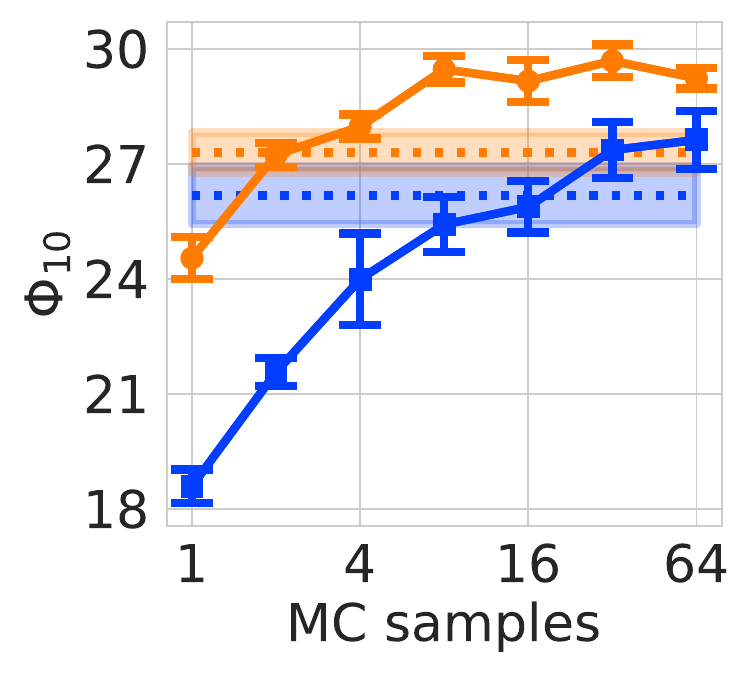}
    \caption{\textbf{Sel. Prediction}}
  \end{subfigure}
  \begin{subfigure}{0.24\textwidth}
    \includegraphics[width=\textwidth]{figures/b3b_nlvr2/phi50_ablation_b3b.pdf}
    \caption{\textbf{Sel. Prediction}}
  \end{subfigure}
  \begin{subfigure}{0.24\textwidth}
    \includegraphics[width=\textwidth]{figures/legends/4_vertical.pdf}
  \end{subfigure}
  \caption{Sample ablation and comparison to MC dropout for BEiT-3 base on NLVR2. Lower is better for ECE and AUC. Standard error across three training runs with different seeds is shown for all methods.}
  \label{fig:mc_ablation_b3b_nlvr}
\end{figure}

\begin{figure}[thbp]
  \centering
  \begin{subfigure}{0.24\textwidth}
    \includegraphics[width=\textwidth]{figures/ood_b3l/accuracy_comparison.pdf}
    \caption{\textbf{Accuracy}}
  \end{subfigure}
  \begin{subfigure}{0.24\textwidth}
    \includegraphics[width=\textwidth]{figures/ood_b3l/ece_comparison.pdf}
    \caption{\textbf{Calibration} ($\downarrow$)}
  \end{subfigure}
  \begin{subfigure}{0.24\textwidth}
    \includegraphics[width=\textwidth]{figures/ood_b3l/cov_0.01_comparison.pdf}
    \caption{\textbf{Sel. Prediction}}
  \end{subfigure}
  \begin{subfigure}{0.24\textwidth}
    \includegraphics[width=\textwidth]{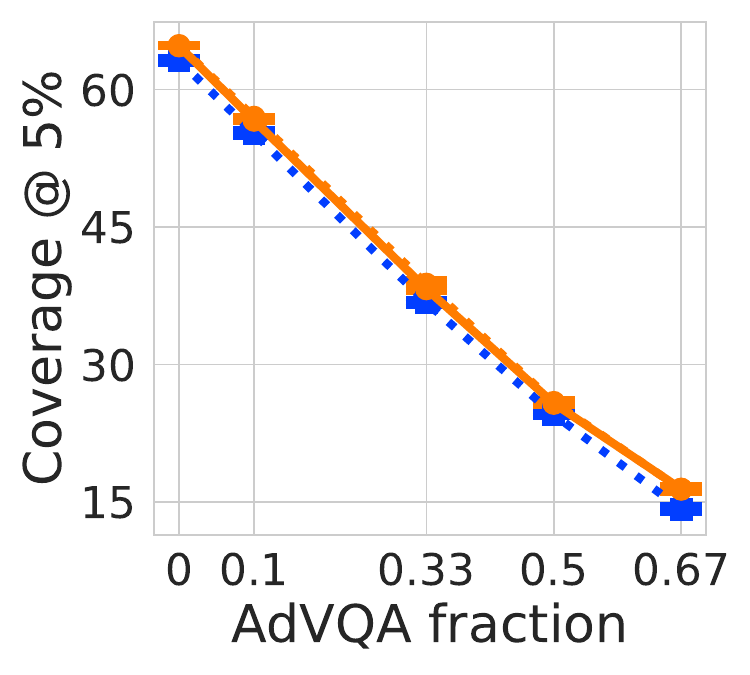}
    \caption{\textbf{Sel. Prediction}}
  \end{subfigure}
  \begin{subfigure}{0.24\textwidth}
    \includegraphics[width=\textwidth]{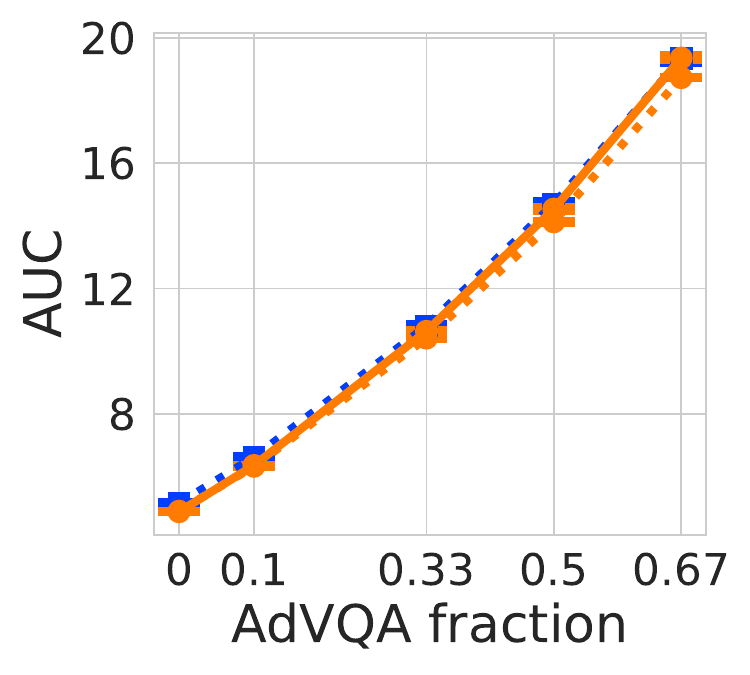}
    \caption{\textbf{Sel. Prediction} ($\downarrow$)}
  \end{subfigure}
  \begin{subfigure}{0.24\textwidth}
    \includegraphics[width=\textwidth]{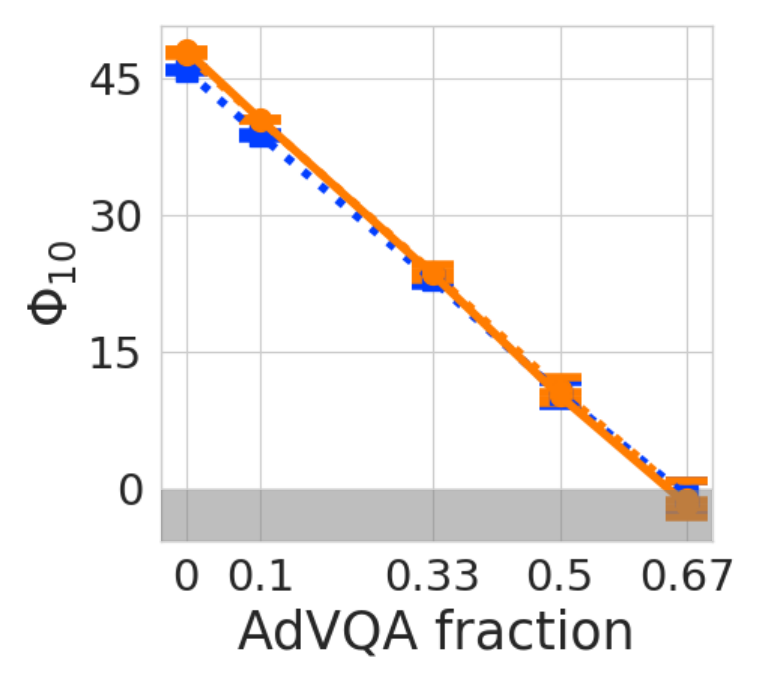}
    \caption{\textbf{Sel. Prediction}}
  \end{subfigure}
  \begin{subfigure}{0.24\textwidth}
    \includegraphics[width=\textwidth]{figures/ood_b3l/phi100_comparison.pdf}
    \caption{\textbf{Sel. Prediction}}
  \end{subfigure}
  \begin{subfigure}{0.24\textwidth}
    \includegraphics[width=\textwidth]{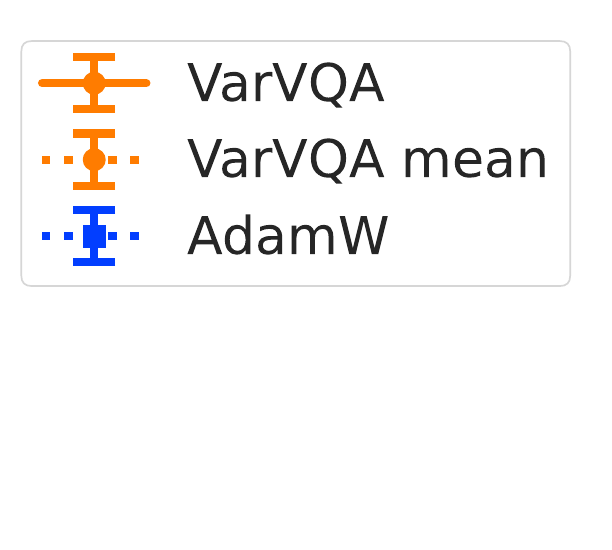}
  \end{subfigure}
  \caption{Performance on different ID/OOD (VQAv2/AdVQA) fractions for BEiT-3 large. In \textbf{(f), (g)}, every model in the gray area is performing worse than a model that abstains on every input.}
  \label{fig:id_ood_full_b3l}
\end{figure}

\begin{figure}[thbp]
  \centering
  \begin{subfigure}{0.24\textwidth}
    \includegraphics[width=\textwidth]{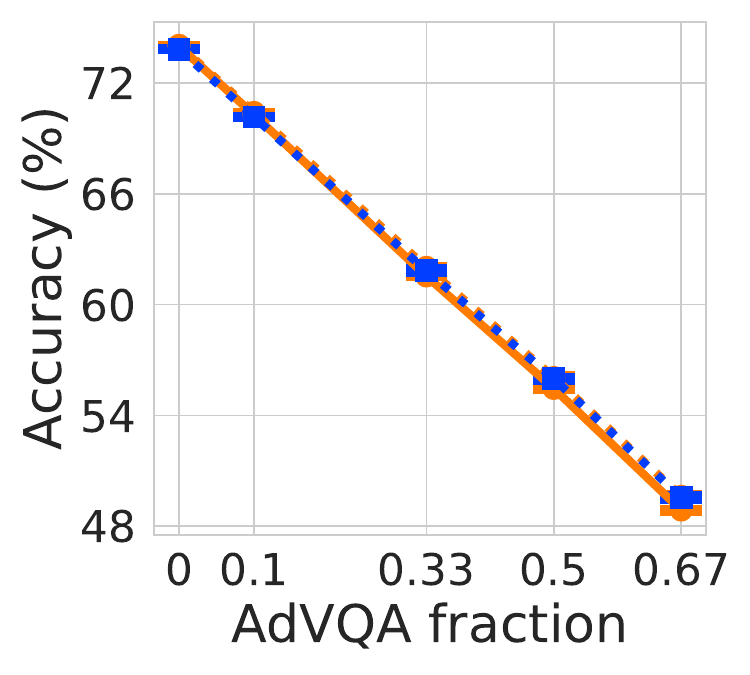}
    \caption{\textbf{Accuracy}}
  \end{subfigure}
  \begin{subfigure}{0.24\textwidth}
    \includegraphics[width=\textwidth]{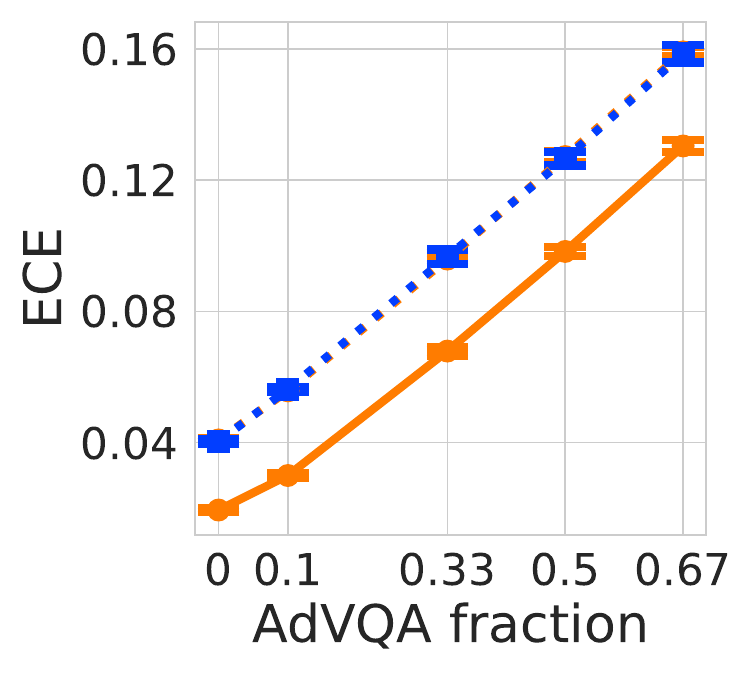}
    \caption{\textbf{Calibration} ($\downarrow$)}
  \end{subfigure}
  \begin{subfigure}{0.24\textwidth}
    \includegraphics[width=\textwidth]{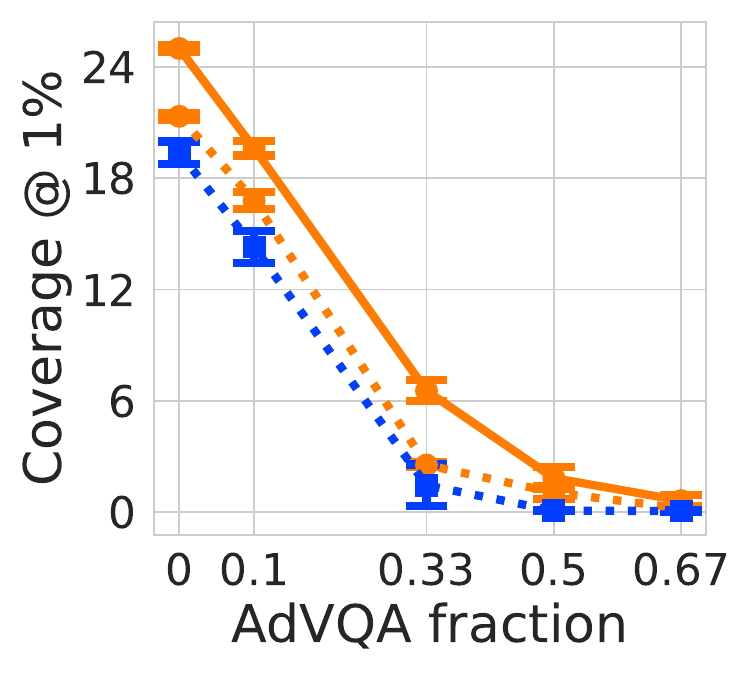}
    \caption{\textbf{Sel. Prediction}}
  \end{subfigure}
  \begin{subfigure}{0.24\textwidth}
    \includegraphics[width=\textwidth]{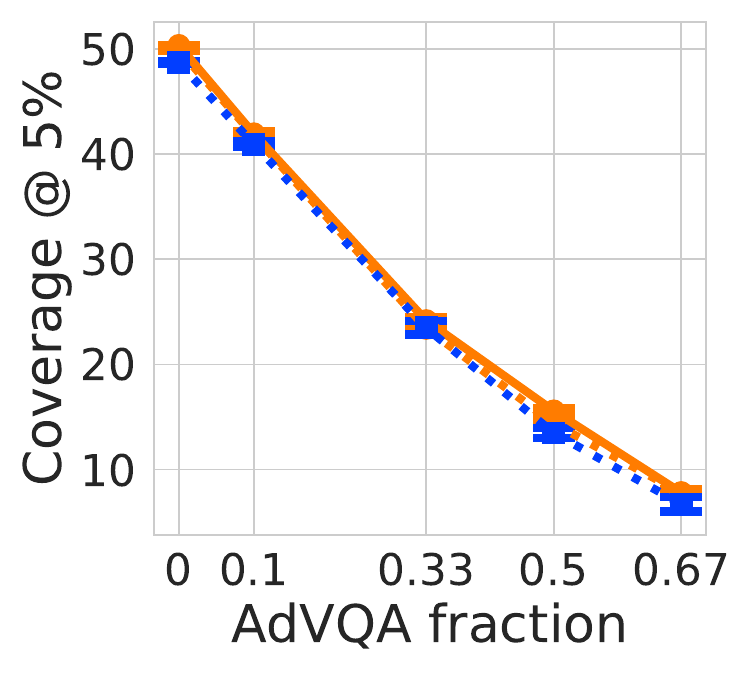}
    \caption{\textbf{Sel. Prediction}}
  \end{subfigure}
  \begin{subfigure}{0.24\textwidth}
    \includegraphics[width=\textwidth]{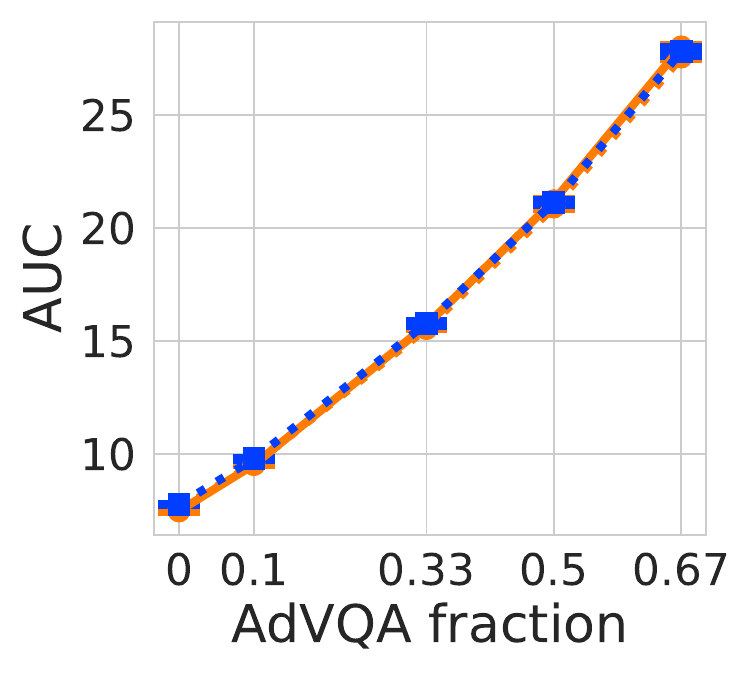}
    \caption{\textbf{Sel. Prediction} ($\downarrow$)}
  \end{subfigure}
  \begin{subfigure}{0.24\textwidth}
    \includegraphics[width=\textwidth]{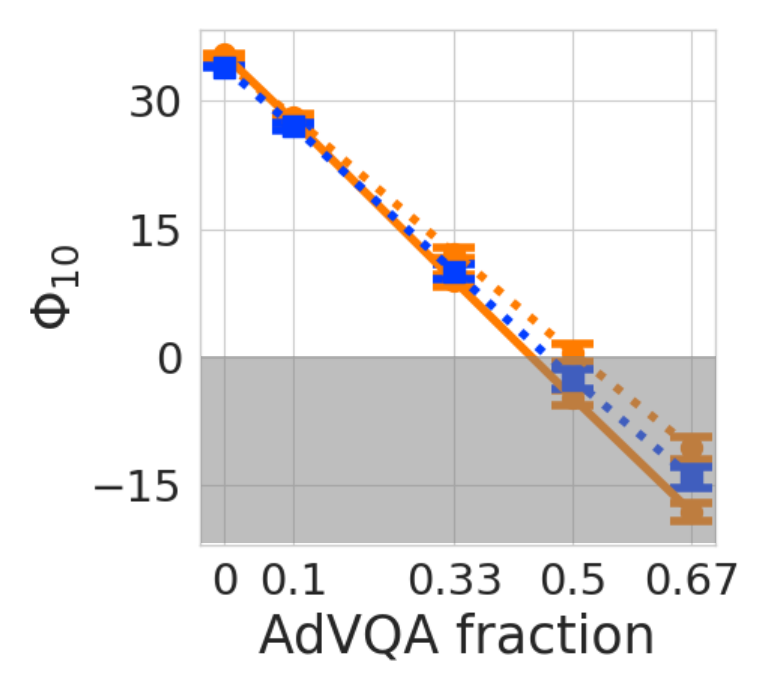}
    \caption{\textbf{Sel. Prediction}}
  \end{subfigure}
  \begin{subfigure}{0.24\textwidth}
    \includegraphics[width=\textwidth]{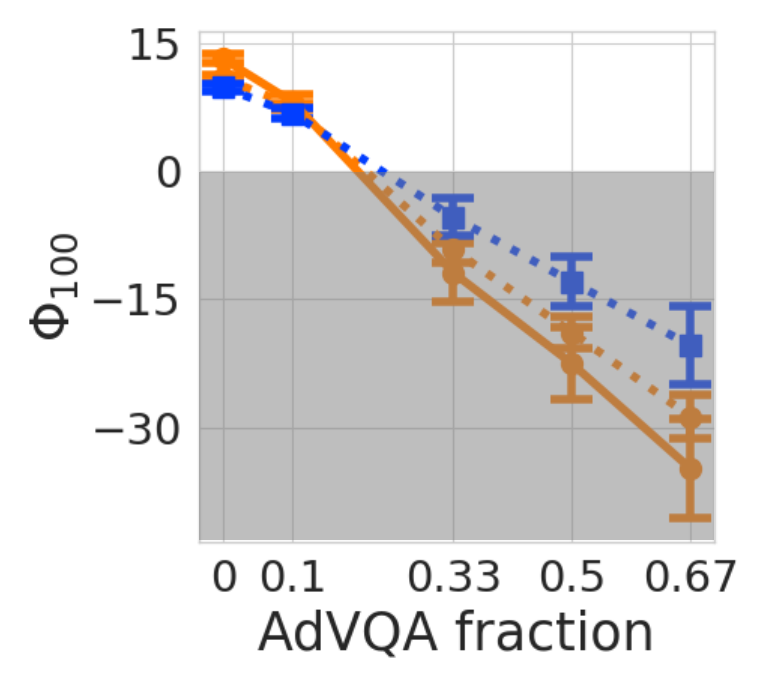}
    \caption{\textbf{Sel. Prediction}}
  \end{subfigure}
  \begin{subfigure}{0.24\textwidth}
    \includegraphics[width=\textwidth]{figures/legends/3_vertical.pdf}
  \end{subfigure}
  \caption{Performance on different ID/OOD (VQAv2/AdVQA) fractions for BEiT-3 base. In \textbf{(f), (g)}, every model in the gray area is performing worse than a model that abstains on every input.}
  \label{fig:id_ood_full_b3b}
\end{figure}

\begin{figure}[thbp]
  \centering
  \begin{subfigure}{0.24\textwidth}
    \includegraphics[width=\textwidth]{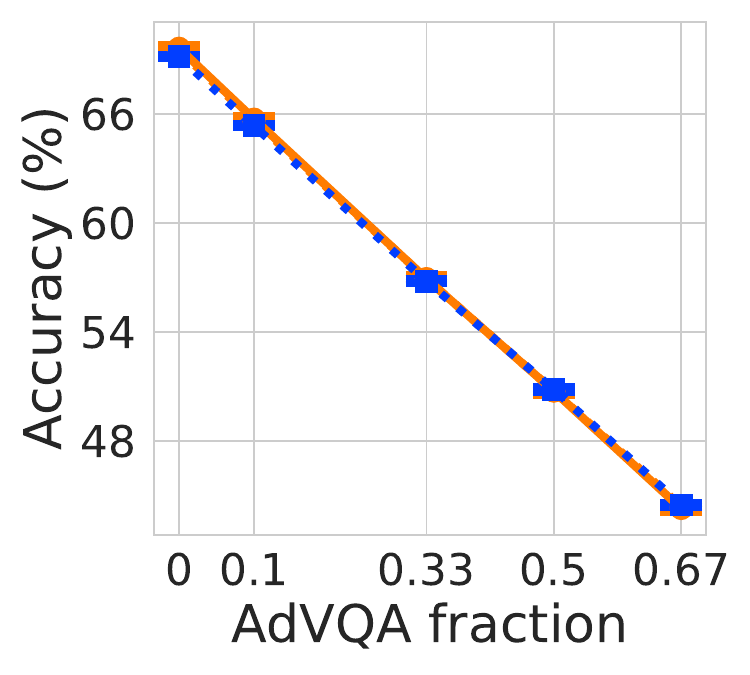}
    \caption{\textbf{Accuracy}}
  \end{subfigure}
  \begin{subfigure}{0.24\textwidth}
    \includegraphics[width=\textwidth]{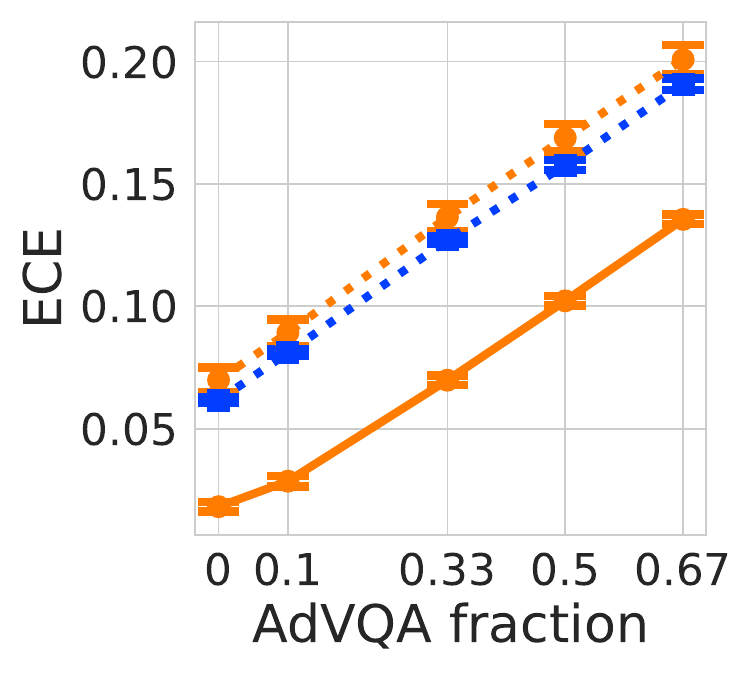}
    \caption{\textbf{Calibration} ($\downarrow$)}
  \end{subfigure}
  \begin{subfigure}{0.24\textwidth}
    \includegraphics[width=\textwidth]{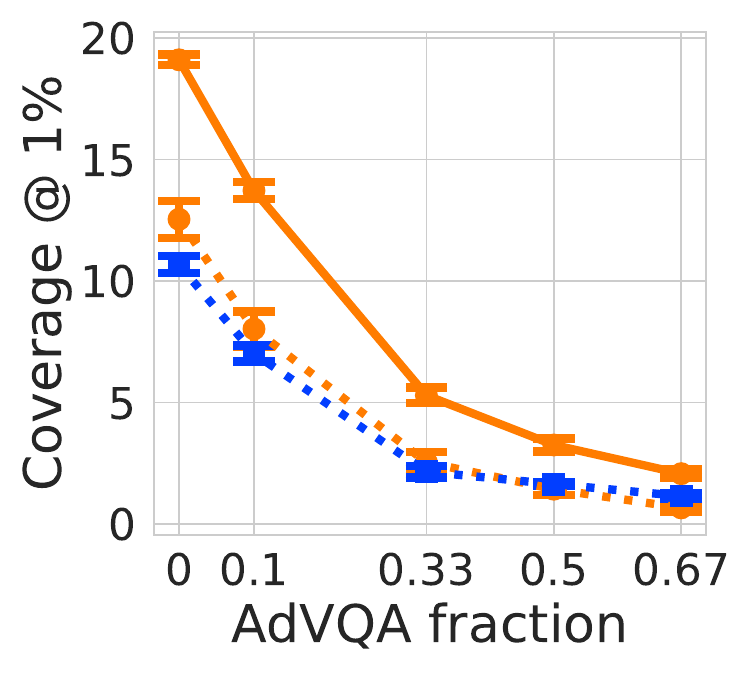}
    \caption{\textbf{Sel. Prediction}}
  \end{subfigure}
  \begin{subfigure}{0.24\textwidth}
    \includegraphics[width=\textwidth]{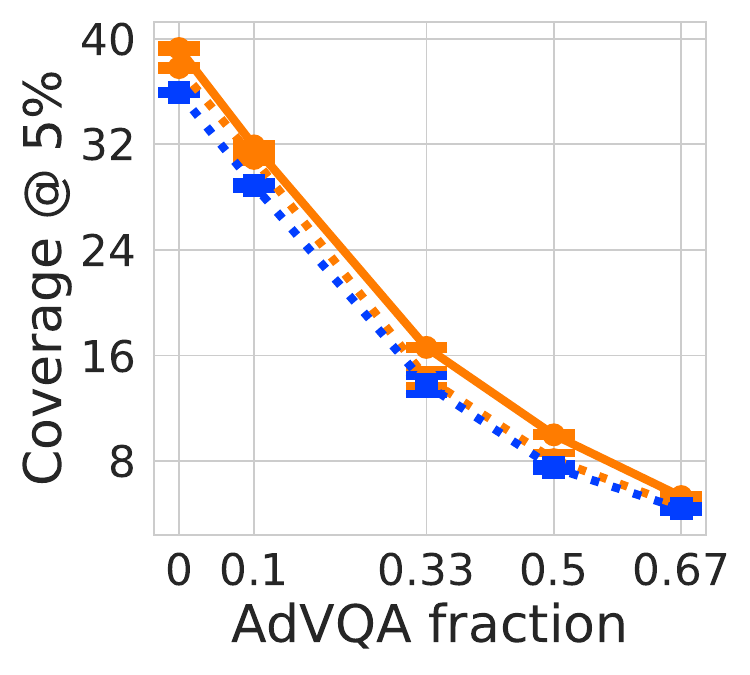}
    \caption{\textbf{Sel. Prediction}}
  \end{subfigure}
  \begin{subfigure}{0.24\textwidth}
    \includegraphics[width=\textwidth]{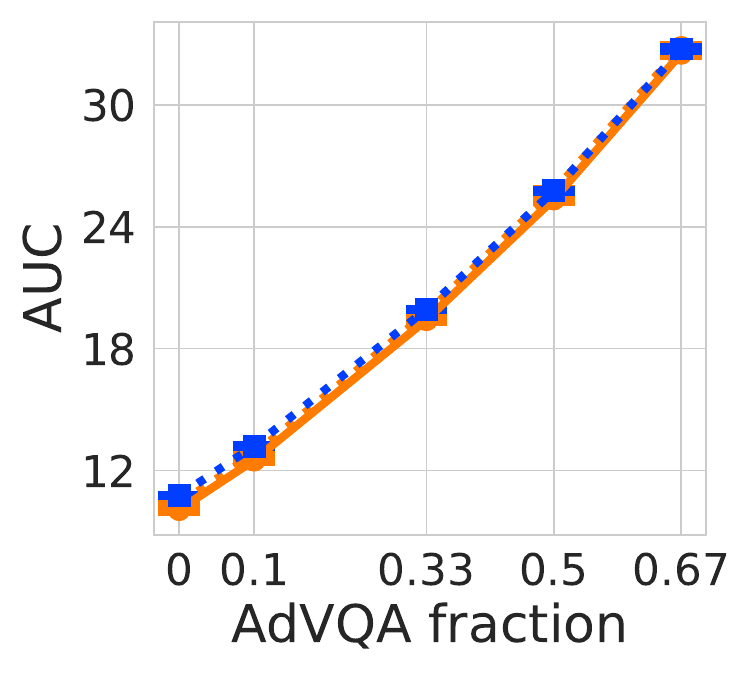}
    \caption{\textbf{Sel. Prediction} ($\downarrow$)}
  \end{subfigure}
  \begin{subfigure}{0.24\textwidth}
    \includegraphics[width=\textwidth]{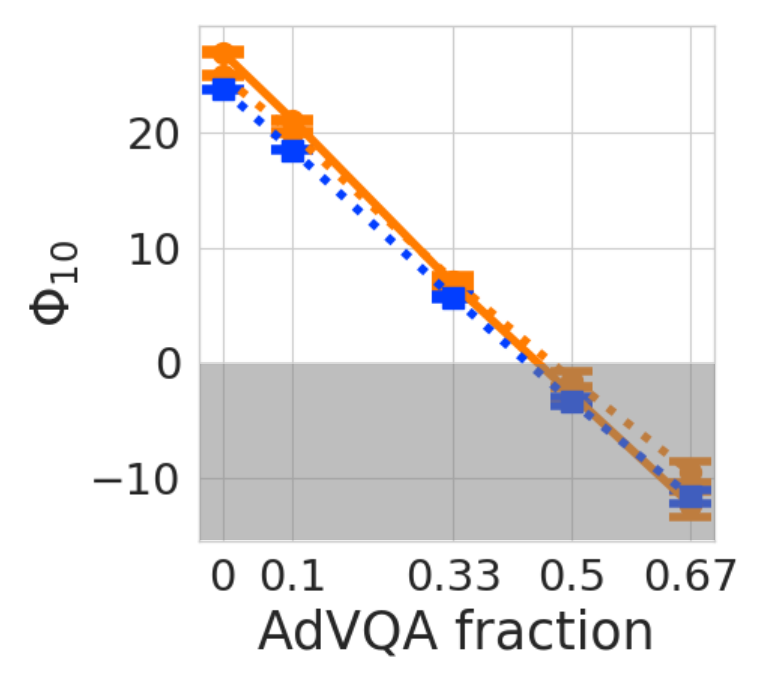}
    \caption{\textbf{Sel. Prediction}}
  \end{subfigure}
  \begin{subfigure}{0.24\textwidth}
    \includegraphics[width=\textwidth]{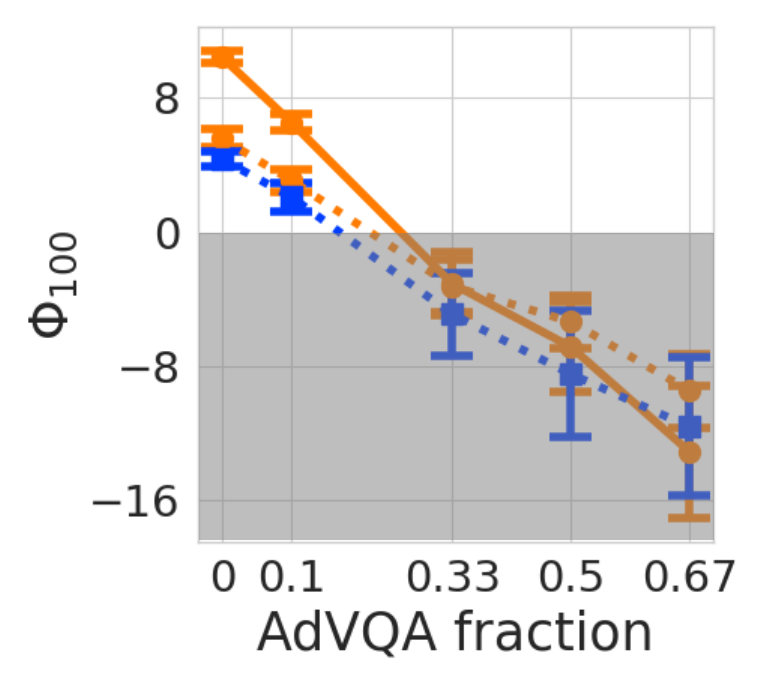}
    \caption{\textbf{Sel. Prediction}}
  \end{subfigure}
  \begin{subfigure}{0.24\textwidth}
    \includegraphics[width=\textwidth]{figures/legends/3_vertical.pdf}
  \end{subfigure}
  \caption{Performance on different ID/OOD (VQAv2/AdVQA) fractions for ViLT. In \textbf{(f), (g)}, every model in the gray area is performing worse than a model that abstains on every input.}
  \label{fig:id_ood_full_vilt}
\end{figure}

\begin{table}[thbp]
    \caption{Coverage on the three different VQA question types achieved by BEiT-3 large when the overall error tolerance is $1\%$. The fraction of each question type is shown in brackets. AdVQA has fewer `Binary', slight fewer `Other' and more `Number' Questions compared to VQAv2.}
    \centering
    \renewcommand{\arraystretch}{1.15}
    \begin{tabular}{l|l|rrrr} \toprule
        \textbf{ID/OOD} ($\%$) & \multirow{2}{*}{Method} & \multirow{2}{*}{All} & \multirow{2}{*}{Binary} &\multirow{2}{*}{Number} &\multirow{2}{*}{Other} \\ 
        (VQAv2/AdVQA) & & & & & \\
        \midrule
        \multirow{3}{*}{100/0} & & (100$\%$) & (38$\%$) & (13$\%$) & (49$\%$) \\
        & AdamW & 32.2 & 56.1 & 6.5 & 20.0 \\
        & VarVQA (ours)& \textbf{37.3} & \textbf{58.9} & \textbf{13.2} & \textbf{26.7} \\ 
        \midrule
        \multirow{3}{*}{90/10} & & (100$\%$) & (36$\%$) & (15$\%$) & (49$\%$) \\
        & AdamW & 25.4 & 48.5 & 2.6 & 14.5 \\
        & VarVQA (ours)& \textbf{30.2} & \textbf{51.4} & \textbf{7.6} & \textbf{20.7} 
        \\ \midrule
        \multirow{3}{*}{67/33} & & (100$\%$) & (33$\%$) & (19$\%$) & (48$\%$) \\
        & AdamW & 11.0 & 26.7 & 0.0 & 4.0 \\
        & VarVQA (ours) & \textbf{15.8} & \textbf{32.8} & \textbf{1.1} & \textbf{9.3} \\
        \midrule
        \multirow{3}{*}{50/50} & & (100$\%$) & (31$\%$) & (22$\%$) & (46$\%$) \\
        & AdamW & 6.0 & 16.3 & 0.0 & 1.5 \\
        & VarVQA (ours) & \textbf{10.5} & \textbf{24.7} & \textbf{0.2} & \textbf{5.3} \\
        \midrule
        \multirow{3}{*}{33/67} & & (100$\%$) & (29$\%$) & (26$\%$) & (45$\%$) \\
        & AdamW & 2.6 & 7.8 & 0.0 & 0.6 \\
        & VarVQA (ours) & \textbf{6.5} & \textbf{17.5} & 0.0 & \textbf{2.8} \\
        \bottomrule
    \end{tabular}
    
    \label{tab:qual_categories_vqa}
    \vspace{1em}
\end{table}

\newpage

\begin{figure}[thbp]
    \centering
    \includegraphics[width=\textwidth]{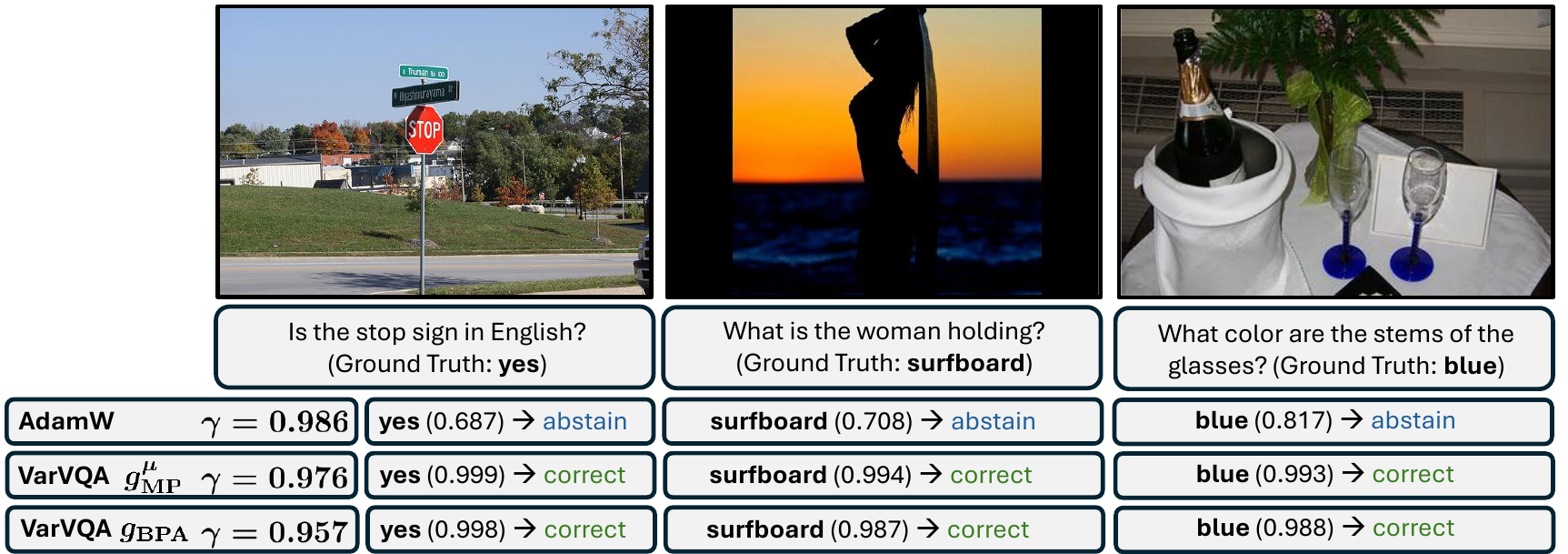}
    \caption{Qualitative examples on VQAv2 with BEiT-3 large where AdamW abstains while VarVQA is correct. The abstention thresholds $\gamma$ were determined by optimizing $\Phi_{100}$ on VQAv2 validation data. Model answers are displayed in \textbf{bold}, the corresponding answer confidences are provided in brackets.}
    \label{fig:qual_adam_abstain_ivon_correct_vqa}
    \vspace{1em}
\end{figure}

\begin{figure}[thbp]
    \centering
    \includegraphics[width=\textwidth]{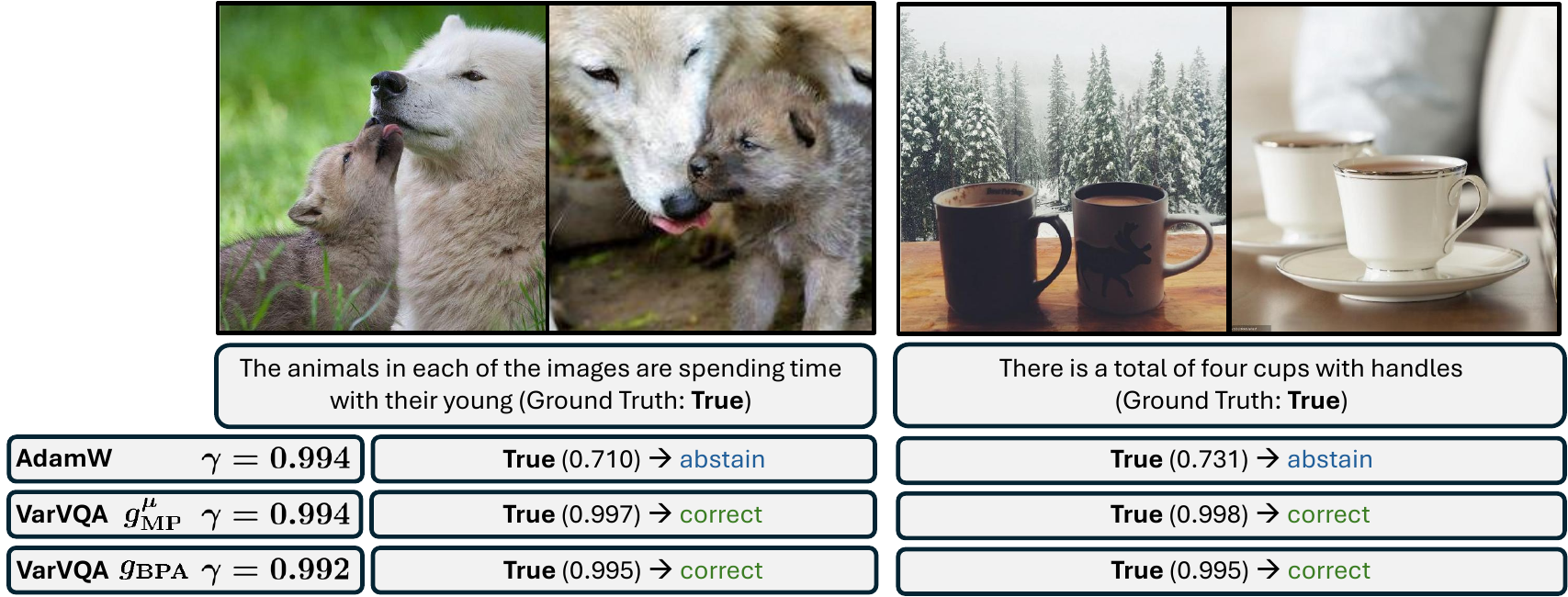}
    \caption{Qualitative examples on NLVR2 with BEiT-3 large where AdamW abstains while VarVQA is correct. The abstention thresholds $\gamma$ were determined by optimizing $\Phi_{100}$ on NLVR2 validation data. Model answers are displayed in \textbf{bold}, the corresponding answer confidences are provided in brackets.}
    \label{fig:qual_adam_abstain_ivon_correct_nlvr}
    \vspace{1em}
\end{figure}

\begin{figure}[thbp]
    \centering
    \includegraphics[width=\textwidth]{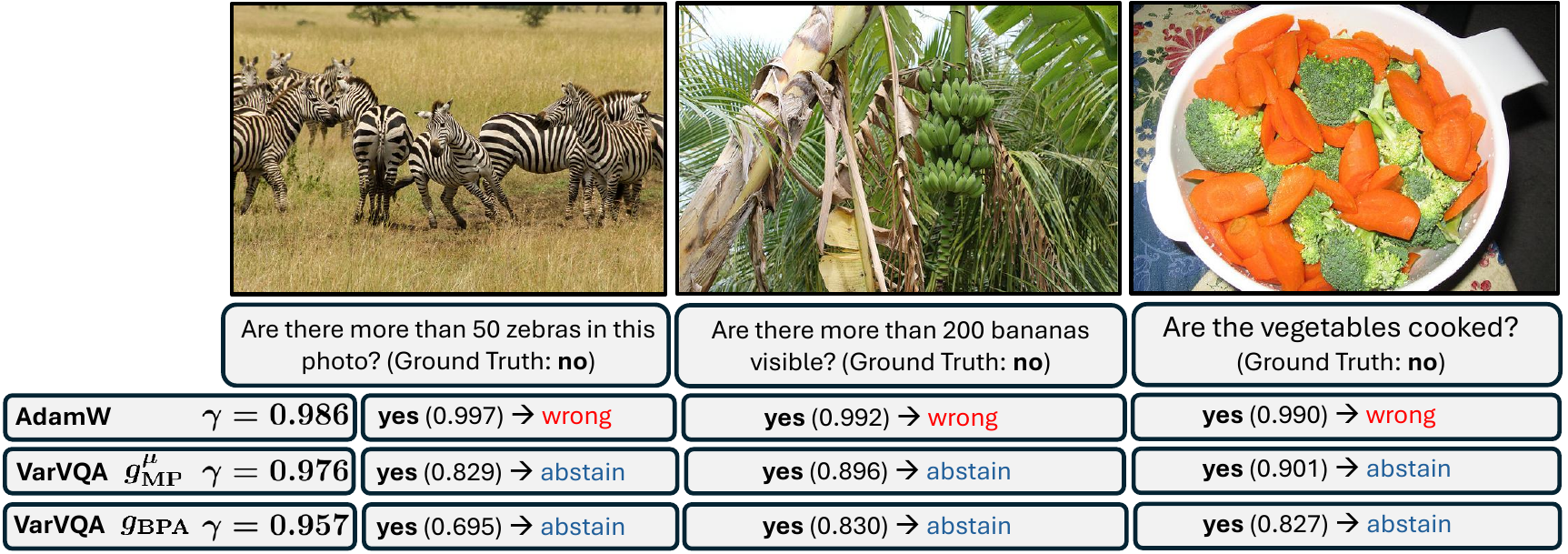}
    \caption{Qualitative examples on AdVQA with BEiT-3 large where AdamW is wrong while VarVQA abstains. The abstention thresholds $\gamma$ were determined by optimizing $\Phi_{100}$ on VQAv2 validation data. Model answers are displayed in \textbf{bold}, the corresponding answer confidences are provided in brackets.}
    \label{fig:qual_adam_wrong_ivon_abstain_advqa}
    \vspace{1em}
\end{figure}

\begin{figure}[thbp]
    \centering
    \includegraphics[width=\textwidth]{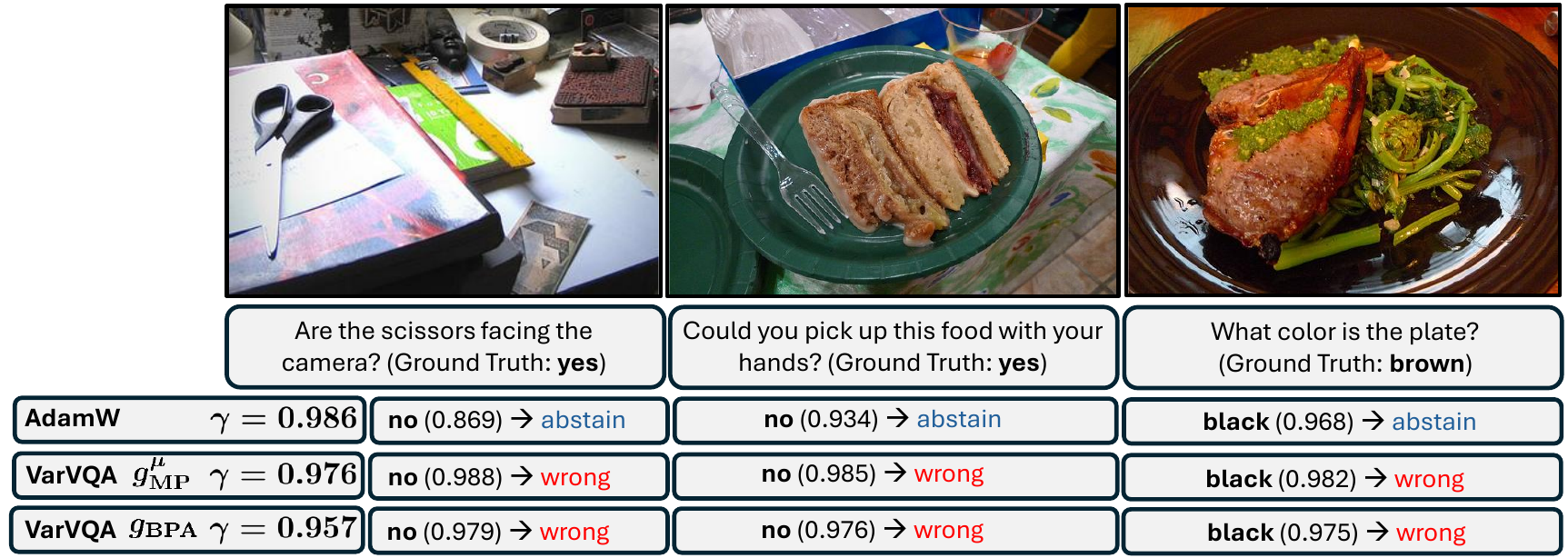}
    \caption{Failure cases on VQAv2 with BEiT-3 large where AdamW abstains while VarVQA is wrong. The abstention thresholds $\gamma$ were determined by optimizing $\Phi_{100}$ on VQAv2 validation data. Model answers are displayed in \textbf{bold}, the corresponding answer confidences are provided in brackets.}
    \label{fig:qual_adam_abstain_ivon_wrong_vqa}
    \vspace{1em}
\end{figure}

\begin{figure}[thbp]
    \centering
    \includegraphics[width=\textwidth]{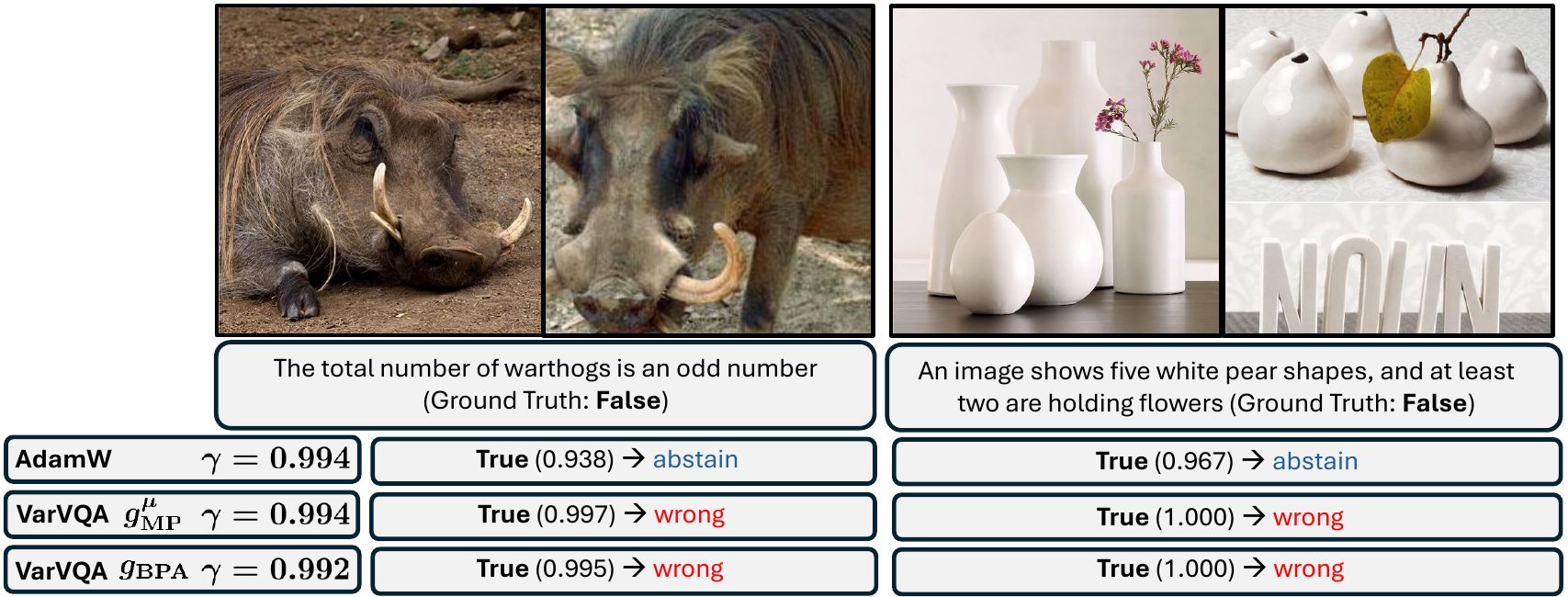}
    \caption{Failure cases on NLVR2 with BEiT-3 large where AdamW abstains while VarVQA is wrong. The abstention thresholds $\gamma$ were determined by optimizing $\Phi_{100}$ on NLVR2 validation data. Model answers are displayed in \textbf{bold}, the corresponding answer confidences are provided in brackets.}
    \label{fig:qual_adam_abstain_ivon_wrong_nlvr}
\end{figure}

%% file: sec_tmlr/G_ensembles.tex
\section{Comparison to Ensembling}\label{sec:supp_ensembling}

We evaluate the impact of applying the Deep Ensembles method \citep{deep_ensembles} on top of all other previously explored confidence estimation methods, as training multiple models constitutes an orthogonal direction of investigation. The results for VQAv2 and NLVR2 are shown in \Cref{tab:metrics_ensembling_id_vqav2} and \Cref{tab:metrics_ensembling_id_nlvr2}, respectively. On VQAv2, which is the much larger and thus much more robust dataset in terms of the sensitive selective prediction metrics, VarVQA combined with Deep Ensembles is often the best method, although the general performance gap is much smaller than without the Deep Ensembles (\cf \Cref{tab:metrics_id_vqav2}). On NLVR2, the results are more mixed. The small dataset size might be a factor, causing high sensitivity of the high-stakes selective prediction metrics to individual samples, although further study is needed to confirm this.

\begin{table}[!tbhp]
    \small
    \renewcommand{\arraystretch}{1.2}
    \centering
    \caption{Reliability evaluation on VQAv2 for fine-tuned models with an additional step of ensembling over three models (\citet{deep_ensembles}). See \cref{tab:metrics_id_vqav2} for the comparison of the uncalibrated models. The variable $N$ denotes the number of forward passes. Best results per model are \textbf{bold}.}
    \begin{tabular}{ll|c|c|c|cccc|cc} \toprule
        \multirow{3}{*}{Model} & \multirow{3}{*}{Method} & \multirow{3}{*}{$N$} & \multirow{3}{*}{\gray{Acc.}} & \multirow{2}{*}{Calibration} & \multicolumn{4}{c|}{Selective Prediction} & \multicolumn{2}{c}{\gray{Sel. Prediction}} \\
        & & & & & \multicolumn{4}{c|}{\textit{high-stakes}} & \multicolumn{2}{c}{\textit{\gray{low-stakes}}} \\
        & & & & ECE~($\downarrow$) & $C@\frac{1}{2}\%$ & $C@1\%$ & $\Phi_{50}$ & $\Phi_{100}$ & \gray{$C@5\%$} & \gray{$\Phi_{10}$}\\ \midrule
        
        \multirow{4}{*}{\centering \;\,ViLT} & AdamW & 1 & \gray{69.69} & 0.049 & 6.97 & 12.30 & 9.97 & 3.67 & \gray{37.86} & \gray{25.14} \\ 
        & VarVQA mean & 1 & \gray{70.03} & 0.062 & 8.26 & 15.33 & 11.16 & 6.15 & \gray{39.50} & \gray{26.04} \\
        \cdashline{2-11}
        & AdamW Dropout & 64 & \gray{69.97} & \textbf{0.015} & 11.68 & 17.10 & 13.33 & 8.52 & \gray{38.97} & \gray{26.70} \\
        & VarVQA & 64 & \gray{70.08} & 0.018 & \textbf{14.39} & \textbf{20.09} & \textbf{14.88} & \textbf{10.65} & \gray{\textbf{40.51}} & \gray{\textbf{27.57}} \\ \midrule
        
        \multirow{4}{1.1cm}{\centering BEiT-3 base} & AdamW & 1 & \gray{74.70} & 0.018 & 15.84 & 23.66  & 19.19 & 11.82 & \gray{\textbf{51.33}} & \gray{\textbf{36.65}} \\
        & VarVQA mean & 1 & \gray{74.29} & 0.029 & 16.86 & 24.51 & 19.12 & 11.77 & \gray{51.12} & \gray{35.82} \\ \cdashline{2-11}
        & AdamW Dropout & 64 & \gray{74.45} & \textbf{0.009} & 16.77 & 24.08 & 19.23 & \textbf{13.91} & \gray{50.61} & \gray{35.65} \\
        & VarVQA & 64 & \gray{74.18} & 0.015 & \textbf{18.34} & \textbf{25.70} & \textbf{19.27} & 10.98 & \gray{51.16} & \gray{36.01} \\ \midrule
        
        \multirow{4}{1.1cm}{\centering BEiT-3 large} & AdamW & 1 & \gray{79.45} & 0.020 & 26.50 & 36.95 & \textbf{30.51} & 19.78 & \textbf{\gray{66.25}} & \textbf{\gray{48.84}} \\ 
        & VarVQA mean & 1 & \gray{79.19} & 0.029 & 27.07 & 36.84 & 28.75 & 23.08 & \gray{65.69} & \gray{48.74} \\
        \cdashline{2-11}
        & AdamW Dropout & 64 & \gray{79.20} & \textbf{0.011} & 28.20 & \textbf{38.12} & 30.15 & 22.81 & \gray{65.44} & \gray{48.60} \\
        & VarVQA & 64 & \gray{79.14} & 0.015 & \textbf{28.68} & 37.97 & 30.40 & \textbf{23.91} & \gray{65.56} & \gray{48.52} \\ \bottomrule
    \end{tabular}
    \label{tab:metrics_ensembling_id_vqav2}
\end{table}

\begin{table}[!tbhp]
    \small
    \renewcommand{\arraystretch}{1.2}
    \centering
    \caption{Reliability evaluation on NLVR2 for fine-tuned models with an additional step of ensembling over three models (\citet{deep_ensembles}). See \cref{tab:metrics_id_nlvr2} for the comparison of the uncalibrated models. The variable $N$ denotes the number of forward passes. Best results per model are \textbf{bold}.}
    \begin{tabular}{ll|c|c|c|cccc|cc} \toprule
        \multirow{3}{*}{Model} & \multirow{3}{*}{Method} & \multirow{3}{*}{$N$} & \multirow{3}{*}{\gray{Acc.}} & \multirow{2}{*}{Calibration} & \multicolumn{4}{c|}{Selective Prediction} & \multicolumn{2}{c}{\gray{Sel. Prediction}} \\
        & & & & & \multicolumn{4}{c|}{\textit{high-stakes}} & \multicolumn{2}{c}{\textit{\gray{low-stakes}}} \\
        & & & & ECE~($\downarrow$) & $C@\frac{1}{2}\%$ & $C@1\%$ & $\Phi_{50}$ & $\Phi_{100}$
        & \gray{$C@5\%$} & \gray{$\Phi_{10}$}\\ \midrule
        
        \multirow{4}{1.1cm}{\centering BEiT-3 base} & AdamW & 1 & \gray{84.90} & 0.020 & 21.36 & 26.07 & 14.67 & 4.68  & \textbf{\gray{63.26}} & \gray{32.09} \\
        & VarVQA mean & 1 & \gray{83.82} & 0.043 & 9.56 & 26.61 & 14.14 & 3.07 & \gray{58.63} & \gray{30.66} \\
        \cdashline{2-11}
        & AdamW Dropout & 64 & \gray{84.58} & 0.017 & 16.81 & \textbf{27.89} & 12.90 & 6.44 & \gray{62.09} & \textbf{\gray{32.34}} \\
        & VarVQA & 64 & \gray{83.59} & \textbf{0.016} & \textbf{21.69} & 25.94 & \textbf{15.33} & \textbf{7.87} & \gray{59.77} & \gray{31.72} \\ \midrule
        
        \multirow{4}{1.1cm}{\centering BEiT-3 large} & AdamW & 1 & \gray{89.61} & \textbf{0.009} & 41.52 & \textbf{53.21} & 29.48 & 16.54 & \textbf{\gray{83.75}} & \textbf{\gray{52.72}} \\ 
        & VarVQA mean & 1 & \gray{89.16} & 0.052 & 26.02 & 43.12 & 19.15 & 9.85 & \gray{81.11} & \gray{47.62} \\
        \cdashline{2-11}
        & AdamW Dropout & 64 & \gray{89.35} & 0.016 & \textbf{43.30} & 53.18 & \textbf{32.01} & \textbf{25.64} & \gray{83.01} & \gray{51.05} \\
        & VarVQA & 64 & \gray{89.55} & 0.023 & 33.53 & 52.15 & 29.48 & 15.72 & \gray{82.81} & \gray{49.49} \\ \bottomrule
    \end{tabular}
    \label{tab:metrics_ensembling_id_nlvr2}
\end{table}

%% file: sec_tmlr/H_threshold_generalization.tex
\section{Threshold Generalization}\label{sec:thresh_gen}

We evaluate how well thresholds for selective prediction that were chosen on a validation split generalize to the test set. \Cref{tab:threshold_generalization_vqav2} and \Cref{tab:threshold_generalization_nlvr2} show the test risk generalization for all methods from the main paper on VQAv2 and NLVR2, respectively. We do not know show test coverage as it is generally strongly correlated with test risk, \ie a higher test risk will lead to a higher test coverage and vice versa. Generally, deviations are similar across methods, and relative deviations are higher for smaller risk levels.

\begin{table}[tbhp]
    \small
    \renewcommand{\arraystretch}{1.2}
    \centering
    \caption{Threshold generalization on VQAv2 for different methods. Shown is the risk on the test set when using the abstention threshold determined on the validation set. The variable $N$ denotes the number of forward passes.}
    \begin{tabular}{ll|c|cc|cc} \toprule
        \multirow{3}{*}{Model} & \multirow{3}{*}{Method} & \multirow{3}{*}{$N$} & \multicolumn{2}{c|}{\textit{high-stakes Sel. Prediction}} & \multicolumn{2}{c}{\textit{low-stakes Sel. Prediction}} \\
        & & & \multirow{2}{2cm}{\centering Test Risk ($R_{\textrm{val}}=0.5\%$)} & \multirow{2}{2cm}{\centering Test Risk ($R_{\textrm{val}}=1\%$)} & \multirow{2}{2cm}{\centering Test Risk ($R_{\textrm{val}}=5\%$)} & \multirow{2}{2cm}{\centering Test Risk ($R_{\textrm{val}}=10\%$)} \\
        & & & & & & \\
        \midrule
        
        \multirow{4}{*}{\centering \;\,ViLT} & AdamW & 1 & 0.43\% & 0.83\% & 5.13\% & 10.03\% \\
         & VarVQA mean & 1 & 0.40\% & 0.77\% & 5.30\% & 10.17\% \\ \cdashline{2-7}
        & AdamW Dropout & 64 & 0.37\% & 0.83\% & 5.10\% & 10.23\% \\
        & VarVQA & 64 & 0.40\% & 0.87\% & 5.13\% & 10.00\% \\ \midrule
        \multirow{4}{1.1cm}{\centering BEiT-3 base} & AdamW & 1 & 0.47\% & 1.23\% & 5.30\% & 10.37\% \\
        & VarVQA mean & 1 & 0.47\% & 1.03\% & 5.20\% & 10.20\% \\ \cdashline{2-7}
        & AdamW Dropout & 64 & 0.67\% & 1.20\% & 5.30\% & 10.43\% \\
        & VarVQA & 64 & 0.47\% & 1.10\% & 5.30\% & 10.37\% \\ \midrule
        
        \multirow{4}{1.1cm}{\centering BEiT-3 large} & AdamW & 1 & 0.57\% & 1.23\% & 5.23\% & 10.10\% \\
        & VarVQA mean & 1 & 0.57\% & 1.17\% & 5.20\% & 10.03\% \\ \cdashline{2-7}
        & AdamW Dropout & 64 & 0.60\% & 1.17\% & 5.23\% & 10.20\% \\
        & VarVQA & 64 & 0.60\% & 1.20\% & 5.23\% & 10.20\% \\ \bottomrule
        
    \end{tabular}
    \label{tab:threshold_generalization_vqav2}
\end{table}

\begin{table}[tbhp]
    \small
    \renewcommand{\arraystretch}{1.2}
    \centering
    \caption{Threshold generalization on NLVR2 for different methods. Shown is the risk on the test set when using the abstention threshold determined on the validation set. The variable $N$ denotes the number of forward passes.}
    \begin{tabular}{ll|c|cc|cc} \toprule
        \multirow{3}{*}{Model} & \multirow{3}{*}{Method} & \multirow{3}{*}{$N$} & \multicolumn{2}{c|}{\textit{high-stakes Sel. Prediction}} & \multicolumn{2}{c}{\textit{low-stakes Sel. Prediction}} \\
        & & & \multirow{2}{2cm}{\centering Test Risk ($R_{\textrm{val}}=0.5\%$)} & \multirow{2}{2cm}{\centering Test Risk ($R_{\textrm{val}}=1\%$)} & \multirow{2}{2cm}{\centering Test Risk ($R_{\textrm{val}}=5\%$)} & \multirow{2}{2cm}{\centering Test Risk ($R_{\textrm{val}}=10\%$)} \\
        & & & & & & \\
        \midrule
        
        \multirow{4}{1.1cm}{\centering BEiT-3 base} & AdamW & 1 & 0.33\% & 1.17\% & 4.73\% & 9.30\% \\
        & VarVQA mean & 1 & 0.90\% & 0.77\% & 4.93\% & 9.87\% \\ \cdashline{2-7}
        & AdamW Dropout & 64 & 0.60\% & 0.97\% & 4.43\% & 9.27\% \\
        & VarVQA & 64 & 0.50\% & 0.97\% & 4.83\% & 9.53\% \\ \midrule
        
        \multirow{4}{1.1cm}{\centering BEiT-3 large} & AdamW & 1 & 0.43\% & 1.10\% & 4.80\% & 9.60\% \\
        & VarVQA mean & 1 & 0.53\% & 1.13\% & 4.77\% & 9.83\% \\ \cdashline{2-7}
        & AdamW Dropout & 64 & 0.43\% & 0.83\% & 4.83\% & 9.40\% \\
        & VarVQA & 64 & 0.73\% & 1.10\% & 4.83\% & 9.20\% \\ \bottomrule
        
    \end{tabular}
    \label{tab:threshold_generalization_nlvr2}
\end{table}

%% file: sec_tmlr/I_selector_comparison.tex
\section{Comparison to the Selector}\label{sec:selec_comp}

We compare VarVQA to the Selector \cite{reliable_vqa} on VQAv2, which requires the overhead a task-specific additional training phase. Unlike the Selector, VarVQA does not need this phase, but still performs competitively, yielding well-calibrated uncertainty estimates out-of-the-box.

\begin{table}[!tbhp]
    \small
    \renewcommand{\arraystretch}{1.2}
    \centering
    \caption{Comparison of VarVQA to AdamW and the AdamW-Selector \citep{reliable_vqa} on VQAv2. Best results per model are \textbf{bold}.}
    \begin{tabular}{ll|c|c|cccc|cc} \toprule
        \multirow{3}{*}{Model} & \multirow{3}{*}{Method} & \multirow{3}{*}{\gray{Acc.}} & \multirow{2}{*}{Calibration} & \multicolumn{4}{c|}{Selective Prediction} & \multicolumn{2}{c}{\gray{Sel. Prediction}} \\
        & & & & \multicolumn{4}{c|}{\textit{high-stakes}} & \multicolumn{2}{c}{\textit{\gray{low-stakes}}} \\
        & & & ECE~($\downarrow$) & $C@\frac{1}{2}\%$ & $C@1\%$ & $\Phi_{50}$ & $\Phi_{100}$ & \gray{$C@5\%$} & \gray{$\Phi_{10}$}\\ \midrule
        
        \multirow{3}{1.1cm}{\centering BEiT-3 base}  & AdamW & \gray{73.60} & 0.041 & 10.35 & 18.55  & 15.59 & 8.65 & \gray{47.93} & \gray{33.40} \\
        & VarVQA  & \gray{73.79} & 0.018 & 18.10 & 24.66 & 19.26 & 13.90 & \gray{49.76} & \gray{35.22} \\
        & AdamW Selector & \gray{74.26} & \textbf{0.010} & \textbf{18.75} & \textbf{26.53} & \textbf{19.96} & \textbf{14.67} & \textbf{\gray{51.31}} & \textbf{\gray{36.12}} \\ \midrule
        
        \multirow{3}{1.1cm}{\centering BEiT-3 large} 
        & AdamW & \gray{78.59} & 0.039 & 21.63 & 32.15 & 26.31 & 17.80 & \gray{63.19} & \gray{45.83} \\
        & VarVQA  & \gray{78.89} & 0.018 & 28.13 & 37.05 & 29.56 & 23.21 & \gray{64.68} & \textbf{\gray{48.06}} \\ 
        & AdamW Selector & \gray{79.45} & \textbf{0.011} & \textbf{31.67} & \textbf{39.53} & \textbf{32.58} & \textbf{26.15} & \textbf{\gray{65.31}} & \gray{47.57} \\ \bottomrule
        
    \end{tabular}
    \label{tab:selector_comparison_vqav2}
\end{table}